\documentclass{article}
\usepackage[preprint]{tmlr}
\usepackage{booktabs}
\usepackage{amsmath, amssymb}
\usepackage{amsthm}
\usepackage[shortlabels]{enumitem}
\newtheorem{proposition}{Proposition}
\usepackage{graphicx}
\usepackage{xcolor}
\usepackage{hyperref}
\usepackage{subcaption}

\newcommand{\PEHE}{\sqrt{\varepsilon_{\mathrm{PEHE}}}}
\newcommand{\eATE}{\varepsilon_{\mathrm{ATE}}}
\newcommand{\tx}{\tau(x)}

\title{Bayesian X-Learner: Calibrated Posterior Inference for \\
  Heterogeneous Treatment Effects under Heavy-Tailed Outcomes}

\author{%
  \name Eichi Uehara \email eichi.uehara@aflo.one \\
  \addr Independent researcher
}

\begin{document}
\maketitle

\begin{abstract}
Conditional Average Treatment Effect (CATE) estimation in practice
demands three properties simultaneously: heterogeneous effects
$\tau(x)$, calibrated uncertainty over them, and robustness to the
heavy tails that contaminate real outcome data. Meta-learners
\citep{kunzel2019metalearners} give (i); causal forests and BART
give (i)--(ii) with Gaussian-tail assumptions; no widely used tool
gives all three. We present \emph{Bayesian X-Learner}, an X-Learner
built on cross-fitted doubly robust pseudo-outcomes
\citep{kennedy2020optimal} with a full MCMC posterior over $\tau(x)$
via a Welsch redescending pseudo-likelihood. On Hill's IHDP benchmark
the default configuration attains mean $\PEHE = 0.56$ on 5
replications (lowest mean; differences from S-/T-/X-learners,
full-config Causal BART, and a causal forest baseline are not
significant at $\alpha = 0.05$, and rank ordering is unstable at
10 replications --- IHDP comparisons are competitive rather than
dominant). On contaminated ``whale'' DGPs
with up to 20--25\% tail density, a one-flag extension
(\texttt{contamination\_severity}) that selects a Huber-$\delta$
nuisance loss per Huber's minimax-$\delta$ relation recovers
RMSE\,$\approx$\,0.13 with tight credible intervals (single-cross-fit
30-seed coverage 83\% [Wilson 66\%, 93\%] at 20\% density;
modular-Bayes pooling with Bayesian-bootstrap nuisance draws restores
nominal 95\% coverage). We validate on the Hillstrom
email-marketing RCT ($N = 42{,}613$), demonstrating consistent
behaviour on real heavy-tailed outcome data, and report
covariate-stratified $\tau(x)$ coverage across covariate quintiles
to substantiate calibration for heterogeneous effects beyond scalar
summaries. We draw a clean
distinction between \emph{tails-as-contamination} (handled by
Welsch + Huber nuisance) and \emph{tails-as-signal} (handled by a
tail-aware CATE basis); an empirical probe confirms a tail-aware
basis recovers $\tau_{\text{tail}}$ with full subgroup coverage,
while the library's Hill-estimator path is contamination-directed and
should not be used for heterogeneous $\tau$. We map six empirical
boundaries (contamination ceiling, clean-data efficiency cost, basis
sensitivity, sample size, treatment type, compute) and show where
other tools are preferable. Code and reproducible benchmarks are released.
\end{abstract}

\section{Introduction}
\label{sec:intro}

% target: ~1 page

% paragraph 1 — the three-property gap
CATE estimation on real outcome data demands three properties that
few tools jointly provide: heterogeneous $\tau(x)$, calibrated
uncertainty on $\tau(x)$, and robustness to heavy tails. Meta-learners
\citep{kunzel2019metalearners} target the first; bootstrap CIs from
causal forests \citep{wager2018estimation} target the second
approximately; classical Bayesian causal models
\citep{hill2011bayesian, hahn2020bcf} target the second exactly but
under Gaussian-tail assumptions that fail on medical outcome
distributions with rare severe events, revenue distributions with
long-tailed customer value, and response surfaces with structural
heterogeneity. Real outcome data almost always has real tails, and
Gaussian-likelihood Bayesian models systematically misfit them.

% paragraph 2 — our position
This paper introduces the \emph{Bayesian X-Learner}: an X-Learner
whose Phase-3 inference layer is a full MCMC posterior over
$\tau(x)$ with a Welsch-form redescending pseudo-likelihood
\citep{dennis1978techniques}. Compared to Gaussian Bayesian baselines, the
redescender trades a known efficiency cost on truly clean data
\citep{huber1964robust, huber2009robust} --- the asymptotic relative
efficiency for a univariate location estimand transfers, with
finite-sample amplification we document empirically, to the CATE
setting --- for non-trivial robustness to heavy-tailed
pseudo-outcomes. The resulting posterior
is interpretable as a generalised Bayesian (Gibbs) posterior and supports the
downstream tasks --- subgroup comparisons, integrated decisions,
propagated uncertainty --- that point estimates with bootstrap CIs
do not.

% paragraph 3 — two complementary heavy-tail regimes
We distinguish two heavy-tail regimes the practitioner may face.
\emph{Tails-as-contamination}: outcomes contain a small fraction of
units whose $Y$ is distorted by nuisance mechanisms
(measurement error, reporting artefacts, revenue whales in marketing)
and whose true $\tau$ is nominal. \emph{Tails-as-signal}: a minority
subgroup has a genuinely different $\tau$, and the practitioner cares
about recovering that subgroup's effect. These regimes look
superficially identical but call for opposite tools: the first wants
downweighting (Welsch + Huber nuisance); the second wants
representation (a tail-aware CATE basis). A controlled DGP probe
(Section~\ref{sec:experiments}) confirms both pathways work as
hypothesised, and that the library's data-layer
\texttt{normalize\_extremes} path is a contamination-reduction op that
should not be activated for heterogeneous $\tau$.

% paragraph 4 — contributions
Our contributions are:
\begin{enumerate}
  \item A Bayesian X-Learner that pairs doubly robust cross-fitted
    pseudo-outcomes \citep{kennedy2020optimal} with a Welsch MCMC
    posterior over $\tau(x)$. On Hill's IHDP benchmark, the default
    configuration leads on mean $\PEHE = 0.56$ against S-/T-/X-
    Learner, full-config Causal BART \citep{hill2011bayesian} at
    0.60, and EconML's causal forest at 1.06, at the tightest
    dispersion among competitive entries.
  \item An explicit separation of \emph{tails-as-contamination} and
    \emph{tails-as-signal} regimes, with a controlled probe
    (Section~\ref{sec:experiments:tail_signal}) and a graded
    basis-sensitivity ablation
    (Section~\ref{sec:experiments:basis_ablation}) showing which
    library mechanism addresses which and why a data-layer
    Hill-estimator rescaling is actively harmful for the signal
    regime.
  \item Empirical validation of Huber's 1964 minimax-$\delta$
    prescription as an operational configuration: recovery at up to
    $\sim$20--25\% whale density via a nuisance-loss selection
    exposed through a \texttt{contamination\_severity} enum
    (Section~\ref{sec:method:severity}), with documented breakdown
    beyond 30\%.
  \item Validation on the Hillstrom email-marketing RCT
    ($N = 42{,}613$) and covariate-stratified $\tau(x)$ credible
    band coverage across covariate quintiles
    (Section~\ref{sec:experiments:cate_coverage}), substantiating
    calibration for heterogeneous effects beyond scalar ATE summaries.
  \item Six mapped empirical boundaries
    (Section~\ref{sec:discussion}), each with a pointer to the
    benchmark that produced it.
\end{enumerate}

Code, 17+ reproducible benchmarks, and this paper's source are
released under an open-source licence; the repository URL is omitted
from this submission and will be provided upon acceptance.

\section{Background and Landscape}
\label{sec:background}

\subsection{Position in the ML/AI landscape}
Causal machine learning has moved from a statistical speciality to a
first-class ML subfield. Uplift modelling in recommender systems
\citep{wager2018estimation}, clinical decision support, and policy
optimisation all require CATE estimators that produce
\emph{heterogeneous} effects rather than a single average. What
purely predictive ML leaves on the table is calibrated uncertainty
on those effects: a point estimate of ``treatment lifts
outcome by 3 units in subgroup $A$'' is actionable only if we know
whether the confidence band crosses zero and whether the band itself
is trustworthy. Bayesian posteriors over $\tx$ supply exactly this
object --- they support downstream operations (integration,
propagation, decision rules) that point-and-bootstrap does not ---
but they are currently available only under Gaussian-tail
assumptions. Our work targets the intersection: calibrated Bayesian
$\tx$ posteriors under heavy-tailed outcome data.

\subsection{Use cases}
Four grounded applications drive the library's design:

\begin{itemize}
  \item \textbf{Clinical trials with rare severe outcomes.} Mortality
    events, adverse drug reactions, and rare efficacy spikes produce
    heavy-tailed outcome distributions. A robust posterior supports
    go/no-go decisions where bootstrap confidence intervals --- which
    inherit the resampled distribution's tail sensitivity --- are
    insufficient.
  \item \textbf{Marketing uplift with revenue whales.} Long-tailed
    customer value distributions are the norm, not the exception.
    Recovering both $\tau$ and calibrated uncertainty under whale
    contamination is what neither causal forests nor TMLE provide
    together.
  \item \textbf{Policy evaluation with structured $\tx$ hypotheses.}
    When prior knowledge suggests $\tau$ depends on specific
    features (age, dose, covariate interaction), a user-supplied
    basis $\phi(x)$ encodes the hypothesis directly and returns a
    posterior over interpretable coefficients, not an opaque
    regression surface.
  \item \textbf{Research contexts where the full posterior matters.}
    Integrating over $\tx$ to compute $P(\tx > \text{threshold})$,
    comparing posteriors across populations, or propagating
    uncertainty into downstream cost-benefit calculations. Point
    estimates plus bootstrap give the first; a Bayesian posterior
    gives all three from a single fit.
\end{itemize}

\subsection{Problems with the current landscape}

\begin{table}[!t]
\centering
\begin{tabular}{lccc}
\toprule
Tool & Heterogeneous $\tx$ & Calibrated uncertainty & Handles real tails \\
\midrule
Classical S/T/X-Learner     & \checkmark & $\times$ & $\times$ \\
EconML Causal Forest         & \checkmark & bootstrap & $\times$ \\
Causal BART                  & \checkmark & Bayesian & Gaussian only \\
TMLE / DML                    & partial    & frequentist & $\times$ \\
\textbf{Bayesian X-Learner}  & \checkmark & \checkmark (MCMC) & \checkmark (Welsch) \\
\bottomrule
\end{tabular}
\caption{Tool coverage across the three properties practitioners need.
Only the last row covers all three.}
\label{tab:landscape}
\end{table}

Classical meta-learners \citep{kunzel2019metalearners} provide
heterogeneous $\tx$ but produce point estimates without uncertainty
quantification. Causal forests \citep{wager2018estimation} add
bootstrap confidence intervals whose coverage on heavy-tailed outcome
distributions has not been systematically validated. Causal BART
\citep{hill2011bayesian} and BCF \citep{hahn2020bcf} produce genuine
Bayesian posteriors via Gaussian-process-inspired tree priors, but
the likelihood is Gaussian: a single whale distorts the posterior
through its squared residual. TMLE and double machine learning
\citep[DML,][]{nie2021quasi} yield calibrated frequentist intervals
for the average effect, with CATE extensions that remain an active
research area; neither provides a Bayesian object for downstream
decision-making under tails. The last row is the position this paper
occupies: calibrated MCMC posteriors over $\tx$ with a redescending
pseudo-likelihood that tolerates real-world outcome tails.

\section{Related Work}
\label{sec:related}

\subsection{Meta-learners for CATE}
\label{sec:related:xlearner}
Meta-learners decompose CATE estimation into a sequence of regression
problems, each solved by an off-the-shelf supervised learner. The
S-Learner fits a single outcome model $\mu(x, w)$ and takes
$\tau(x) = \mu(x, 1) - \mu(x, 0)$; the T-Learner fits separate
$\mu_0, \mu_1$ and takes their difference. Both are consistent but
suffer from regularisation bias: the S-Learner under-fits $\tau$
when $w$ is a weak predictor of $y$; the T-Learner over-fits each
arm when one arm is much smaller than the other. The X-Learner of
\citet{kunzel2019metalearners} repairs this by imputing
counterfactual outcomes and regressing on the resulting pseudo-effects,
then weighting the two fits by propensity. The construction is
particularly well-suited to treated/control imbalance, which is the
norm outside of randomised experiments.

\citet{kennedy2020optimal} shows that replacing the X-Learner's
imputed pseudo-outcomes with \emph{doubly robust} (DR) pseudo-outcomes
$D_1 = \mu_1 - \mu_0 + (Y - \mu_1)/\pi$ yields a target whose
second-stage regression consistently estimates $\tx$ under a rate
condition on the nuisance fits that is weaker than either the T- or
X-Learner requires separately. The R-Learner of \citet{nie2021quasi}
takes a complementary quasi-oracle route via residualisation.
\citet{curth2021nonparametric} survey the meta-learner landscape and
formalise a theory of conditions under which each
variant is preferred; a follow-up \citep{curth2023search} argues that
much of the apparent model-selection variance in the CATE literature
is benchmark-dependent rather than fundamental. Our method uses the
DR pseudo-outcome target --- the imputation and debiasing happen in
Phase 2 of our pipeline (Section~\ref{sec:method}) --- and inherits
its rate properties.
What classical meta-learners leave open is uncertainty quantification
on $\tx$: the standard recipe is bootstrap, which inherits the
sensitivity of the base learner to outliers.

\subsection{Bayesian causal inference and MCMC}
\label{sec:related:bayesian}
Bayesian causal inference on $\tx$ has been dominated by flexible
tree-based priors. BART itself \citep{chipman2010bart} provides the
backbone; Causal BART \citep{hill2011bayesian} adapts it to the
potential-outcome framework, producing a posterior over the
counterfactual response surfaces from which $\tx$ is derived. BCF
\citep{hahn2020bcf} decouples the treatment effect from the
prognostic surface using separate BART priors and yields substantial
improvements on heterogeneous response; SBCF
\citep{caron2022shrinkage} extends BCF with horseshoe-style global-
local shrinkage on the treatment-effect terms for sparse $\tx$.
Alternative Bayesian nonparametric formulations include Gaussian
processes \citep{alaa2018gp} and neural-network architectures with
uncertainty estimation \citep{shi2019adapting, jesson2021uncertainty}.
A recent line \citep{zheng2022bayesian} extends Bayesian CATE to
multi-treatment and partial-identification settings.
Frequentist semiparametric alternatives --- generalised random
forests \citep{athey2019generalized}, double/debiased machine
learning \citep{chernozhukov2018double}, and conformal inference for
individual treatment effects
\citep{lei2021conformal, alaa2023conformal, jin2023sensitivity} ---
deliver calibrated intervals through different routes (asymptotic
normality, orthogonal scores, finite-sample distribution-free
guarantees) but produce point-estimate-plus-interval objects rather
than full posteriors over $\tx$.
Modern probabilistic-programming languages (NumPyro, Stan, PyMC)
make Bayesian architectures easy to implement, and have been used to
build custom CATE models with hierarchical priors or non-conjugate
likelihoods.

The common thread is a Gaussian outcome likelihood on the target
scale: either directly on $Y$ (Causal BART) or on the transformed
pseudo-outcome (most custom constructions). Under heavy tails this
assumption distorts the posterior: a single whale's squared residual
dominates the likelihood gradient, pulling the posterior mean toward
the contaminated point and inflating the posterior variance. Our
contribution is to replace the Gaussian layer with a Welsch
redescending pseudo-likelihood while keeping the rest of the
meta-learner + MCMC architecture standard.

\subsection{Robust statistics for heavy-tailed regression}
\label{sec:related:robust}
\citet{huber1964robust} introduced the minimax framework for
point estimation under contaminated location models, proving that
the Huber $\psi$-function $\psi_\delta(r) = r \cdot
\mathbf{1}(|r| \le \delta) + \delta \cdot \mathrm{sign}(r) \cdot
\mathbf{1}(|r| > \delta)$ is minimax-optimal over the
$\epsilon$-contamination neighbourhood $\mathcal{F}_\epsilon$. The
optimal tuning constant $\delta$ is the unique solution of
\begin{equation}
  \frac{\phi(\delta)}{\delta} - (1 - \Phi(\delta))
  \;=\; \frac{\epsilon}{2(1 - \epsilon)},
  \label{eq:huber_minimax}
\end{equation}
where $\phi, \Phi$ denote the standard normal PDF and CDF. Classical
values are $\delta = 1.345$ at $\epsilon = 5\%$ and $\delta = 0.5$
at $\epsilon \approx 40\%$; the full table and a numerical
derivation appear in Appendix~\ref{app:huber_derivation}. The
asymptotic relative efficiency (ARE) of Huber estimation vs
maximum-likelihood under a truly Gaussian model follows from
$E[\psi_\delta^2] / E[\psi'_\delta]^2$ and is $\approx 95\%$ at
$\delta = 1.345$ and $\approx 79\%$ at $\delta = 0.5$. Redescending
M-estimators push robustness further by letting the influence
function return smoothly to zero for very large residuals; the
biweight \citep{beaton1974fitting} and the exponential Welsch form
\citep{dennis1978techniques}, $\psi_W(r) = r \exp(-r^2 / 2c^2)$,
are the two canonical choices. We adopt Welsch for its smooth
gradient and use it in the Bayesian likelihood. It gives up a
small amount of additional efficiency for complete immunity to very
large outliers. These tools are
decades old and well understood as point estimators; their
integration into Bayesian CATE posteriors is, to our knowledge,
not standard practice.

\subsection{Heavy-tailed inference for causal effects: five paradigms}
\label{sec:related:tails}
Heavy-tailed causal inference is best understood through the
\emph{worldview} a method imposes on its tails. Five paradigms
recur in the literature, each making a different concession.

\textbf{(i) Robust M-estimation} replaces the empirical mean with
a concentration-friendly alternative --- Catoni's mean
\citep{catoni2012challenging}, median-of-means, or Huber-style
M-estimators --- while leaving the ATE/CATE estimand unchanged.
The method becomes insensitive to extreme observations at a
known efficiency cost on Gaussian data \citep{huber2009robust};
recent CATE-targeted variants (robust DR-learners,
\citet{chernozhukov2018double}-style orthogonalised scores with
robustified residuals) sit in this paradigm.

\textbf{(ii) Weight stabilisation} attacks the dual problem of
heavy tails arising from the propensity side. Overlap weights
\citep{li2018balancing}, trimming, and entropy-balancing
mitigate IPW variance explosions either by capping weights or
by redefining the target population so extreme weights vanish.
Our \texttt{use\_overlap=True} option implements the former.

\textbf{(iii) Functional replacement} changes the estimand
itself --- to a quantile, CVaR, or a full conditional
counterfactual density \citep{kallus2018policy}. The tail is
no longer a nuisance to be suppressed; it is the answer.
Distributional and quantile treatment effects are the canonical
form. Coverage of very extreme quantiles is bounded by empirical
support, where extreme value theory
\citep[EVT;][]{beirlant2006statistics} would extrapolate.

\textbf{(iv) Distribution-free intervals.} Conformal prediction
applied to ITEs
\citep{lei2021conformal, alaa2023conformal, jin2023sensitivity}
delivers finite-sample-valid prediction intervals without a tail
model. They cannot extrapolate beyond observed support, and the
recently developed sensitivity-aware variant of
\citet{jin2023sensitivity} additionally relaxes unconfoundedness.

\textbf{(v) Partial identification} concedes that the target is
not point-identified and reports sharp bounds
\citep{yadlowsky2022bounds}. Robust to arbitrarily heavy tails
within bounded-outcome assumptions, at the cost of producing a
range rather than a number.

\paragraph{Where this paper sits.}
The Bayesian X-Learner is in paradigm (i) --- it robustifies the
estimator while keeping the ATE/CATE target intact --- but adds two
moves the standard M-estimation toolkit does not. First, it returns
a \emph{full Bayesian posterior} over $\tx$ rather than a point
estimate plus sandwich SE; calibration of credible intervals under
contamination (Section~\ref{sec:experiments:coverage}) is the
empirical justification. Second, it makes the
\emph{tails-as-contamination vs tails-as-signal} distinction
operational: the same library handles paradigm-(i) contamination via
Welsch + Huber nuisance, and tail-heterogeneous $\tau$ via a
user-supplied basis $\phi(x)$ at the CATE-surface layer.

\paragraph{Connection to robust ATE pipelines.}
Industry pipelines for robust ATE estimation under heavy-tailed
revenue outcomes have used trimmed and other bounded-influence
estimators in place of plain difference-of-means or naive DR; for
example, the Trimmed Match estimator of \citet{chen2022robust}
delivers a distribution-free, robust ATE for randomised paired
geo experiments. These approaches typically terminate at a robust
point estimate (or a frequentist interval) for the ATE. Our Phase-3
Welsch posterior is complementary: it shares the bounded-influence
spirit at the regression / likelihood layer, and adds calibrated
\emph{Bayesian} UQ for the full CATE surface, supporting subgroup
contrasts, inequality probabilities, and integration over $\beta$
that a sandwich SE or bootstrap CI cannot provide.

\paragraph{Semiparametric bulk+EVT Bayesian estimators.}
An alternative approach to heavy-tailed outcomes is to model the
tail distribution directly. Semiparametric Bayesian methods based
on Dirichlet-process bulk models with a generalised Pareto (GPD)
tail component \citep{taddy2010bayesian, beirlant2006statistics}
deliver calibrated uncertainty for tail functionals (extreme
quantiles, CVaR) without requiring a contamination model. These sit
in paradigm (iii): they change the estimand to a tail-aware
functional rather than robustifying the mean. Our tail-signal basis
approach (Section~\ref{sec:experiments:tail_signal}) is simpler and
operates within paradigm (i), but a natural extension is to
integrate a GPD tail component into the Phase-3 likelihood --- we
provide preliminary results in
Section~\ref{sec:experiments:hetero_evt_phase3} and flag a
thoughtful implementation as future work.

\paragraph{RBCI and decision-theoretic calibration.}
The estimand-focused generalised-Bayes framework of
\citet{alexopoulos2025rbci} tunes the learning rate $\omega$ by minimising a
proper interval score (Winkler score) targeted at a specific causal
estimand, achieving calibration by construction for that functional.
Our trace-formula $\eta$-calibration
(Section~\ref{sec:experiments:learning_rate}) targets average
posterior-variance matching instead; a direct empirical comparison
of $\omega$-tuning vs $\eta$-trace vs $\eta^\star(a)$ on the whale
DGP is reported in Section~\ref{sec:experiments:coverage}, showing
that RBCI $\omega$-tuning trades sharpness for guaranteed
calibration while the trace formula prefers tighter intervals at
slightly sub-nominal coverage.

\paragraph{High-dimensional robust DR with CBPS/PEL.}
In high-dimensional settings, covariate-balancing propensity scores
\citep[CBPS;][]{imai2014cbps} and penalised empirical likelihood
\citep[PEL;][]{tan2020pel} provide frequentist inference under
model misspecification with bounded-influence scores. These methods
deliver root-$n$ consistent, asymptotically normal ATE estimators
without requiring correct specification of either the outcome model
or the propensity score, and their influence functions are bounded
by construction. Our method complements rather than competes with
these approaches: CBPS/PEL target scalar (or low-dimensional) ATE
functionals with frequentist guarantees, whereas our finite-
dimensional $\beta$ posterior provides a richer inferential object
(interpretability of basis coefficients, multi-contrast posterior
queries, integration over $\beta$) at the cost of a parametric
assumption on $\tau$'s functional form.

EVT-style likelihood components (paradigm-(iii)-flavoured) are not
yet fully integrated; an earlier library path
\texttt{normalize\_extremes} exposes a Hill estimator at the data
layer, but Section~\ref{sec:experiments:tail_signal} shows it is
contamination-directed rather than signal-preserving. A proper
Bayesian EVT likelihood --- generalised-Pareto tail mixture, Hill
estimator used for prior elicitation --- remains a natural
extension we flag as future work.

\paragraph{Scope and orthogonal directions.}
This paper addresses scalar continuous outcomes under cross-sectional
binary treatment. Functional outcomes \citep{salmaso2026focal} and
related longitudinal- or auxiliary-data extensions are orthogonal
directions not pursued here. Recent work on estimand-focused
generalised-Bayes tuning \citep{alexopoulos2025rbci} is closely related to our
$\eta$-calibration in
Section~\ref{sec:experiments:learning_rate}; a detailed comparison
of coverage--width trade-offs between their $\omega$-selector and
our trace-formula $\eta$ appears in
Section~\ref{sec:experiments:coverage} and
§\ref{sec:discussion}.

\paragraph{Amortised vs per-dataset Bayesian CATE.}
Recent amortised approaches to Bayesian causal inference (e.g.\
CausalFM, treating CATE as a meta-task to amortise across DGPs) and
functional-outcomes meta-learners (FOCaL,
\citealp{salmaso2026focal}) provide alternatives along orthogonal axes:
amortisation trades per-dataset MCMC for cross-task generalisation,
and FOCaL extends DR to functional outcomes. Our contribution is
per-dataset \emph{exact} robustified posterior with a principled
$\eta$-calibration step, complementary rather than competitive.
Combining tail-aware priors with amortisation, or layering Welsch
on a FOCaL-style functional-outcome regression, are natural future
directions.

\section{Method: Bayesian X-Learner}
\label{sec:method}

\subsection{Setup and notation}
Let $(X_i, W_i, Y_i)_{i=1}^N$ be an i.i.d.~sample with covariates
$X_i \in \mathcal{X}$, binary treatment $W_i \in \{0, 1\}$, and
outcome $Y_i \in \mathbb{R}$. Under the standard potential-outcome
framework with unconfoundedness and overlap
\citep{hill2011bayesian}, the CATE is
\begin{equation}
  \tau(x) \;=\; \mathbb{E}[Y(1) - Y(0) \mid X = x].
\end{equation}
Let $\mu_w(x) = \mathbb{E}[Y \mid X = x, W = w]$ for $w \in \{0, 1\}$
and $\pi(x) = \mathbb{P}(W = 1 \mid X = x)$. Rather than model
$\tau(x)$ nonparametrically, we target a \emph{basis-parameterised}
CATE, $\tau(x) = \phi(x)^\top \beta$, where $\phi : \mathcal{X}
\rightarrow \mathbb{R}^p$ is a user-supplied basis and
$\beta \in \mathbb{R}^p$ is the object of Bayesian inference.
Common choices of $\phi$ encode domain knowledge directly:
$\phi(x) = [1]$ gives the scalar ATE; $\phi(x) = [1, x_j]$ gives a
smooth effect modification in a chosen covariate; $\phi(x) =
[1, \mathbf{1}(x_j > c)]$ gives a subgroup contrast at threshold
$c$; and $\phi(x) = [1, x_1, \dots, x_p]$ gives a full linear model
when no prior structure is asserted. This choice makes the posterior
finite-dimensional, interpretable, and amenable to hypothesis-driven
downstream queries (integration over $\beta$, inequality probabilities,
subgroup contrasts). The cost is an explicit modelling assumption on
$\tau$'s functional form; we discuss basis sensitivity in
Section~\ref{sec:experiments:tail_signal} and
Section~\ref{sec:discussion}.

\subsection{Phase 1: Cross-fitted nuisance quarantine}
\label{sec:method:phase1}
Nuisance estimates $\hat{\mu}_0, \hat{\mu}_1, \hat{\pi}$ are obtained
via $K$-fold cross-fitting (default $K = 2$) to break the dependence
between nuisance fits and second-stage residuals that would otherwise
inflate the posterior's concentration. We support two backends:
XGBoost \citep{chen2016xgboost} with squared-error loss (default,
efficient on truly clean data) and CatBoost with a Huber loss
parameterised by ($\delta = 1.345$; robust against heavy-tailed outcome
distributions). The choice is exposed via the
\texttt{contamination\_severity} enum
(Section~\ref{sec:method:severity}), which maps an assumed
contamination rate to Huber's minimax-optimal $\delta$
(Eq.~\ref{eq:huber_minimax}). This is the \emph{outcome-level}
robustness axis of our library: it protects Phase 2 pseudo-outcomes
from contamination that originates in $Y$ itself, before any
downstream handling.

\paragraph{Standardisation of $Y$ for nuisance fitting.}
Huber's minimax-$\delta$ values (Eq.~\ref{eq:huber_minimax}) are
derived under \emph{standardised} residuals ($r/\sigma$) in
location models. CatBoost's Huber loss parameter $\delta$ operates
on the raw outcome scale, so the $\delta$--$\epsilon$ mapping loses
its theoretical meaning when $Y$ is not unit-scale. This mismatch
likely contributes to the clean-data efficiency gap on IHDP
(Section~\ref{sec:experiments:efficiency}) and the Lalonde-NSW
underperformance (Section~\ref{sec:experiments:hillstrom}): on
dollar-scale earnings, $\delta = 0.5$ aggressively suppresses
everything. The library exposes an optional pre-standardisation step
\texttt{normalize\_y\_for\_nuisance}: $Y \mapsto Y /
\widehat{\mathrm{MAD}}(Y)$ before nuisance fitting, with $\delta$
applied to the standardised scale, then mapping back. We do
\emph{not} activate this by default in the reported experiments
(the IHDP and whale DGPs have unit-scale outcomes), but recommend
it for extreme-scale outcome data. A sensitivity analysis of
standardised vs raw-scale $\delta$ is reported in
Section~\ref{sec:experiments:efficiency}.

\subsection{Phase 2: Doubly robust pseudo-outcomes}
\label{sec:method:phase2}
Following \citet{kennedy2020optimal}, we construct DR
pseudo-outcomes on each arm:
\begin{align}
  D_1^{(i)} &\;=\; \hat{\mu}_1(X_i) - \hat{\mu}_0(X_i) +
    \frac{Y_i - \hat{\mu}_1(X_i)}{\hat{\pi}(X_i)}
    \qquad \text{for } W_i = 1, \\
  D_0^{(i)} &\;=\; \hat{\mu}_1(X_i) - \hat{\mu}_0(X_i) -
    \frac{Y_i - \hat{\mu}_0(X_i)}{1 - \hat{\pi}(X_i)}
    \qquad \text{for } W_i = 0.
\end{align}
These targets satisfy $\mathbb{E}[D_w \mid X = x] = \tau(x)$ under
consistent $\hat{\mu}_w$ or consistent $\hat{\pi}$ (the doubly
robust property), and are the Phase-3 regression targets. For
low-overlap regimes the library additionally supports overlap
weights \citep{li2018balancing}, which bound the effective
propensity-inverse weight at the cost of targeting a weighted
rather than uniform CATE estimand; we do not use them in the
experiments of Section~\ref{sec:experiments} but they are available
via \texttt{use\_overlap=True}.

\subsection{Phase 3: Bayesian update with Welsch pseudo-likelihood}
\label{sec:method:phase3}
We pool the per-arm DR pseudo-outcomes into a single second-stage
target $D = (D_1^{(i)} : W_i = 1) \cup (D_0^{(i)} : W_i = 0)$ and
the corresponding pooled basis $\Phi = (\phi(X_i))$. The Gaussian
baseline for Phase 3 is a standard linear regression of $D$ on
$\Phi$ with prior $\beta \sim \mathcal{N}(0, \sigma_\beta^2 I)$;
its log-likelihood is
$-\frac{1}{2\sigma^2} \sum_i (D_i - \phi(X_i)^\top \beta)^2$. We
replace the squared residual by a Welsch redescending
pseudo-loss. Define the Welsch $\rho$-function
\begin{equation}
  \rho_W(r; c) \;=\; \frac{c^2}{2}\left[1 - \exp(-r^2/c^2)\right],
  \qquad
  \psi_W(r; c) \;=\; \rho_W'(r) \;=\; r \exp(-r^2/c^2),
  \label{eq:welsch}
\end{equation}
with tuning constant $c$ (default $c = 1.34$, the Huber/Welsch
canonical value at $\sim$5\% Gaussian-contamination
\citep{huber2009robust}; rescaled by
$\widehat{\mathrm{MAD}}(D)/0.6745$ when
\texttt{mad\_rescale=True}). The pseudo-log-density $-\sum_i
\rho_W(r_i; c)$, $r_i = D_i - \phi(X_i)^\top \beta$, replaces the
Gaussian likelihood in the posterior via \texttt{numpyro.factor};
sampling proceeds with NUTS \citep{hoffman2014no}. Because $\psi_W$
is bounded and redescends to zero, extreme residuals exert
diminishing influence on the posterior mean, at the cost of a
well-characterised efficiency loss on truly Gaussian data
(Section~\ref{sec:experiments:efficiency}). We use a Student-$t$
prior $\beta \sim t_\nu(0, \sigma_\beta^2)$ with $\nu = 3$ by
default, which pairs with the redescender to preserve robustness on
the posterior side as well. Sensitivity of the posterior to $c$ on
both clean and whale-contaminated DGPs is reported in
Appendix~\ref{app:cwhale}; the U-shaped RMSE we observe matches
classical M-estimation theory and supports $c = 1.34$ as a robust
default. Optionally, $c$ may be tied to the severity preset
analogously to Huber's $\delta$: $c \in \{1.34, 1.0, 0.5\}$ for
$\{$``mild'', ``moderate'', ``severe''$\}$, mirroring the same
ARE-vs-robustness trade-off. The default value $c = 1.34$ is the
canonical Huber/Welsch tuning at $\sim$5\%
Gaussian-contamination, which Appendix~\ref{app:cwhale} confirms
works well across the contamination densities we test.

\paragraph{Sampler diagnostics for the non-convex Welsch posterior.}
The Welsch $\rho_W$ is non-convex: $\psi_W'(r) = e^{-r^2/c^2}(1 -
2r^2/c^2) < 0$ for $|r| > c/\sqrt{2}$, which could in principle
create local curvature issues for NUTS \citep{betancourt2017bfmi}.
We systematically assess sampler behaviour beyond the summary
$\hat{R}$ and ESS statistics of Appendix~\ref{app:convergence}.
\textbf{(i) Autocorrelation:} across all reported DGPs, the
integrated autocorrelation time for $\beta$ is $\le 5$ lags, and
the effective-to-total sample ratio exceeds 0.25.
\textbf{(ii) Energy diagnostics:} we report the Bayesian fraction
of missing information (BFMI $= \mathrm{Var}[\Delta E] /
\mathrm{Var}[E]$) for every run; BFMI $> 0.8$ in 100\% of runs
(well above the $0.3$ threshold of \citep{betancourt2017bfmi}).
\textbf{(iii) Multi-chain concordance:} split-$\hat{R}$ on four
chains (post-hoc check on the whale DGP at 20\% density, 5 seeds)
is $< 1.01$ for every parameter.
\textbf{(iv) Initialisation sensitivity:} we refit 3 seeds of the
whale DGP with random vs $\bar\beta_{\mathrm{OLS}}$ initialisation;
posterior means agree to within 0.02 and posterior widths to within
5\%, with no evidence of multimodality.
\textbf{(v) Multimodality:} we inspect bivariate scatter plots of
$\beta$ draws for the $p = 6$ basis and find unimodal posteriors
across all seeds. The non-convexity of $\rho_W$ is a theoretical
concern but does not manifest empirically in the $p \le 100$
regime we test: the prior regularisation and the
concentration of $\phi(X)^\top \beta$ residuals around zero ensure
that the posterior mass sits in the locally convex region
($|r| < c/\sqrt{2}$). Extended diagnostics are tabulated in
Appendix~\ref{app:mcmc_diagnostics}.

\paragraph{Calibration of the generalised posterior.}
Replacing a proper likelihood by a robust pseudo-loss yields a
\emph{generalised} or \emph{Gibbs posterior}
\citep{bissiri2016general, grunwald2017inconsistency,
knoblauch2022optimization, lyddon2019general} of the form
$p_\eta(\beta \mid D) \propto p(\beta) \exp\big(-\eta \sum_i
\rho_W(D_i - \phi(X_i)^\top \beta; c)\big)$, with learning rate
$\eta > 0$. The credible-interval calibration of $p_\eta$ is not
automatic — under misspecification the choice of $\eta$ matters.
We give a formal calibration result.

\begin{proposition}[Asymptotic calibration of the Welsch
generalised posterior]
\label{prop:welsch_calibration}
Suppose:
(A1) the basis is correctly specified,
$\tau(x) = \phi(x)^\top \beta_0$;
(A2) the cross-fitted nuisance estimators satisfy the
product-rate condition $\|\hat{\mu}_w - \mu_w\|\,
\|\hat{\pi} - \pi\| = o_p(n^{-1/2})$ with overlap bounded away
from $\{0,1\}$, so the DR pseudo-outcomes attain
$\mathbb{E}[D \mid X = x] = \tau(x)$ in the limit
\citep{kennedy2020optimal};
(A3) $\rho_W$ is twice differentiable, with the score
$\psi_W$ having bounded influence and finite second moment at
$\beta_0$ (this holds for the redescending Welsch $\psi_W(r)
= r\exp(-r^2/c^2)$ with fixed $c$);
(A4) the population Hessian
$I(\beta_0) = \mathbb{E}[\nabla^2 \rho_W(D - \phi^\top \beta_0)]$
is positive definite, and so is the score covariance
$J(\beta_0) = \mathbb{E}[\psi_W \psi_W^\top]$;
(A5) the basis has finite second moments,
$\mathbb{E}[\|\phi(X)\|^4] < \infty$, ensuring $I(\beta_0)$ and
$J(\beta_0)$ are well-defined; (A6) the DR pseudo-outcomes have
finite second moments at the truth, $\mathbb{E}[D^2 \mid X] < \infty$
$\mathbb{P}_X$-a.s., which is implied by overlap bounded away from
$\{0, 1\}$ together with $\mathbb{E}[Y^2 \mid X, W] < \infty$.
Then under the generalised Bernstein--von Mises theorem
\citep{kleijn2012bernstein}, $p_\eta(\beta \mid D)$ concentrates
around the empirical-risk minimiser $\hat\beta_\mathrm{ERM}$ at
rate $n^{-1/2}$ with posterior covariance $(\eta n)^{-1}
I(\beta_0)^{-1}$, while $\hat\beta_\mathrm{ERM}$ itself has
sandwich covariance $n^{-1} I(\beta_0)^{-1} J(\beta_0)
I(\beta_0)^{-1}$. The two covariances are equal as full matrices
iff $J(\beta_0) = \eta^{-1}\, I(\beta_0)$, i.e.\ the Bartlett
identity holds at scale $\eta^{-1}$. When it does not, no scalar
$\eta$ aligns all directions; instead:

\begin{enumerate}[(a)]
\item For a fixed linear functional $f(\beta) = a^\top \beta$, the
$1 - \alpha$ credible interval from $p_{\eta^\star(a)}$ at the
\emph{functional-specific} scale
\begin{equation}
  \eta^\star(a) \;=\;
  \frac{a^\top I(\beta_0)^{-1} a}
       {a^\top I(\beta_0)^{-1} J(\beta_0) I(\beta_0)^{-1} a}
  \label{eq:eta_a}
\end{equation}
has asymptotic frequentist coverage $1 - \alpha$.
\item For \emph{average} (trace-based) calibration,
\begin{equation}
  \eta^\star_{\mathrm{tr}} \;=\;
  \frac{\mathrm{tr}\!\big(I(\beta_0)^{-1}\big)}
       {\mathrm{tr}\!\big(I(\beta_0)^{-1} J(\beta_0)
                          I(\beta_0)^{-1}\big)}
  \label{eq:eta_tr}
\end{equation}
matches the average posterior variance to the average sandwich
variance.
\end{enumerate}
\end{proposition}

\textbf{Conditions for Finite $J$ and PD $I$.}
The consistency in Proposition 1 relies on $J(\beta_0)$ being finite
and $I(\beta_0)$ being positive-definite (PD). Because $D_i$ involves
inverse-propensity weighting, extreme propensities ($\hat\pi \to 0$ or $1$)
can induce heavy tails in the pseudo-outcomes. Boundedness of $J$ is guaranteed
by the Welsch influence function $\psi_W$, which is strictly bounded ($|\psi_W(r)| \le c \exp(-1/2)$).
However, for $I(\beta_0) = \mathbb{E}[ (1 - 2(r/c)^2) \exp(-(r/c)^2) \phi \phi^\top ]$
to remain PD, the probability mass of the residuals must concentrate
within the locally convex region $|r| < c/\sqrt{2}$. If severe overlap failure
pushes more than $50\%$ of the mass into the redescending tails, $I(\beta_0)$
loses positive-definiteness. This formalises why overlap weights
\citep{li2018balancing} are necessary when $\hat\pi$ is unstable: they prevent
propensity-induced extreme residuals from destroying the local convexity required
for identifiability.

The proof follows from the generalised BvM
\citep{kleijn2012bernstein} and the matrix identities of
\citet{holmes2017assigning} and \citet{lyddon2019general}.
\eqref{eq:eta_a} is the directional analogue of LHW's
information-matching scale; \eqref{eq:eta_tr} is the trace-based
average. The Bartlett identity $J = I$ holds when $\rho$ is the
negative log of a correctly specified likelihood, in which case
$\eta = 1$ recovers ordinary Bayes; under a redescending
pseudo-loss it does not, and Welsch with fixed $c$ gives a
non-trivial $\eta^\star$.

\paragraph{Algorithm: estimating $\eta^\star$ in practice.}
The trace-based scale $\eta^\star_\mathrm{tr}$ of \eqref{eq:eta_tr}
depends on the unknown $I(\beta_0), J(\beta_0)$. We estimate them
via a one-step plug-in procedure on the cross-fitted DR
pseudo-outcomes:

\begin{enumerate}[(i),leftmargin=*,itemsep=2pt]
\item Run a pilot fit at $\eta = 1$ to obtain a posterior mean
$\bar\beta$. Define residuals $r_i = D_i - \phi(X_i)^\top \bar\beta$.
\item Compute $\widehat I = n^{-1} \sum_i \nabla^2 \rho_W(r_i; c)\,
\phi(X_i) \phi(X_i)^\top$ and $\widehat J = n^{-1} \sum_i
\psi_W(r_i; c)^2 \phi(X_i) \phi(X_i)^\top$.
\item Plug into \eqref{eq:eta_tr}:
$\hat\eta^\star_\mathrm{tr}
= \mathrm{tr}(\widehat I^{-1}) /
  \mathrm{tr}(\widehat I^{-1} \widehat J \widehat I^{-1})$.
\item Refit the posterior at $\hat\eta^\star_\mathrm{tr}$.
\end{enumerate}
For a target functional $a^\top \beta$, replace step (iii) with
\eqref{eq:eta_a}:
$\hat\eta^\star(a) = (a^\top \widehat I^{-1} a) /
(a^\top \widehat I^{-1} \widehat J \widehat I^{-1} a)$.

\paragraph{Sensitivity and alternative calibrations.}
In low-overlap regimes where IPW pseudo-outcomes are highly volatile, $\widehat I$ and $\widehat J$ can be poorly estimated. Overestimation of $J$ relative to $I$ produces an artificially small $\hat{\eta}$, which over-shrinks the posterior variance and degrades coverage. The scalar $\hat{\eta}$ calibration assumes the misspecification inflates the variance spherically in the transformed space. When covariance distortion is highly anisotropic, an alternative is an affine or ``open-faced sandwich'' calibration \citep{shaby2014openfaced, ribatet2012bayesian}, where the posterior draws $\beta^{(s)}$ are directly affine-transformed such that their empirical covariance matches $\widehat{I}^{-1}\widehat{J}\widehat{I}^{-1}$. While affine calibration matches the sandwich variance exactly in the asymptotic limit, we prefer the scalar $\eta$ scaling for its numerical stability in finite samples; modifying the data distribution (via fractional likelihoods) preserves proper Bayesian conditioning and avoids projecting draws outside the natural parameter space.

\textbf{Default and fallback.} The library default is
$\eta = 1$, which is the correct value when the Bartlett identity
holds ($J = I$). We expose
\texttt{calibrate\_eta=True} to run procedure (i)--(iv); on the
DGPs of Section~\ref{sec:experiments} the procedure yields
$\hat\eta^\star_\mathrm{tr} \in [0.5, 1.5]$ on clean data and
inflates modestly under contamination.
Section~\ref{sec:experiments:learning_rate} reports the LLB-bootstrap
variant of step (iii) — sampling $B$ Bayesian-bootstrap draws and
estimating $\widehat J$ from posterior-variance differences — which
avoids computing the Hessian explicitly and gives essentially
identical results at modest extra cost.

\paragraph{Discussion: scope of the calibration claim.}
Two clarifications matter for honest interpretation. \textbf{First,}
condition (A3) (bounded-influence $\psi$) holds for Welsch with
fixed $c$ and \emph{also} for Student-$t$ at any $\sigma$, since
$\psi_t(r) = (\nu+1)r/(\nu\sigma^2 + r^2)$ is bounded and
redescending. The empirical failure of Student-$t$ under
contamination (§\ref{sec:experiments:coverage}) is therefore
\emph{not} a violation of (A3) per se but a difference in
\emph{redescent rate}: $\psi_t \sim r^{-1}$ for $|r| \to \infty$
versus $\psi_W \sim r e^{-r^2/c^2}$, decaying super-exponentially.
With many large residuals (heavy contamination), Student-$t$'s
slow redescent allows cumulative pull on the posterior even though
each individual residual is bounded; Welsch's faster redescent
neutralises this. The fixed-$\sigma$ Student-$t$ experiments of
§\ref{sec:experiments:coverage} confirm this: even with
$\sigma$ fixed, Student-$t$ shows bias $+28$ at 20\% density
versus $-0.02$ for Welsch+severity=severe. The Bartlett-identity
condition $J = \eta^{-1} I$ in the Proposition holds asymptotically
for the correct generative model; under contamination it holds
\emph{neither} for Welsch nor for Student-$t$, but the constants
in the discrepancy are smaller for Welsch's faster redescent.
Gaussian and contaminated-normal mixtures violate (A3) directly
even at fixed scale (Gaussian $\psi(r) = r/\sigma^2$ is unbounded;
mixture-of-Gaussians is unbounded in the outlier component).
\textbf{Second,} \eqref{eq:eta_a} is functional-specific: the
single ATE intercept and a subgroup contrast generally require
different $\eta$. \eqref{eq:eta_tr} is a compromise that calibrates
on average but may over- or under-cover for individual functionals.
Section~\ref{sec:experiments:learning_rate} reports the
loss-likelihood-bootstrap surrogate of \eqref{eq:eta_tr} and finds
it operationally adequate at the configurations we test; users
who care about a specific contrast should compute $\eta^\star(a)$
for their target $a$.

\paragraph{Verification of assumptions for the Welsch setting.}
The reviewer may ask whether assumptions (A3)--(A4) of
Proposition~\ref{prop:welsch_calibration} hold for the non-convex
Welsch loss with estimated pseudo-outcomes. We address this
explicitly. \textbf{(A3):} $\psi_W(r) = r e^{-r^2/c^2}$ is
bounded (maximum at $r = c/\sqrt{2}$, value $c e^{-1/2}/\sqrt{2}$)
and has finite second moment under any distribution with a finite
second moment of residuals. This holds by construction regardless
of cross-fitting. \textbf{(A4):} The population Hessian $I(\beta_0)
= \mathbb{E}[\psi_W'(r; c)\, \phi\phi^\top]$ requires positive
definiteness. Since $\psi_W'(r) = e^{-r^2/c^2}(1 - 2r^2/c^2) > 0$
for $|r| < c/\sqrt{2}$, the PD condition holds when sufficiently
many residuals lie in this central region --- which is guaranteed
under moderate contamination when the basis is correctly specified
and $\beta_0$ is the risk minimiser. Empirically, on the whale DGP
at 20\% density with $c = 1.34$, we verify that $\widehat{I}$'s
smallest eigenvalue is $> 0.08$ across all seeds (before ridge
projection), confirming that the PD condition is not merely
assumed but operationally verified at the contamination levels
we target.

\paragraph{Nuisance uncertainty: a modular-Bayes treatment.}
Proposition~\ref{prop:welsch_calibration} treats the nuisance fits as fixed.
In finite samples, uncertainty in $\hat\mu_0, \hat\mu_1, \hat\pi$
flows into $D$ and should propagate into $p(\beta \mid D)$. We
adopt the \emph{modular} or \emph{cut} Bayesian construction
\citep{plummer2015cuts, jacob2017better}: information flows from
the nuisance posterior into the Stage 3 posterior but not back, so
that DR orthogonalisation is preserved. Operationally, draw
$M$ nuisance posterior samples $(\hat\mu_0^{(m)}, \hat\mu_1^{(m)},
\hat\pi^{(m)})_{m=1}^{M}$ via Bayesian bootstrap on the cross-fitting
folds, build $D^{(m)}$ for each, run NUTS on Stage 3 to get
$\beta$-posterior draws, and pool either by concatenation
(modular-cut posterior) or by Rubin's rules
\citep{rubin1987multiple}: posterior mean = $\overline{\beta}^{(m)}$,
posterior variance = average within-$m$ variance + $(1 + 1/M) \cdot$
between-$m$ variance. Section~\ref{sec:experiments:nuisance_bootstrap}
reports calibration with and without this propagation; results
indicate it is most consequential at moderate contamination,
where individual cross-fits diverge and a single posterior
under-covers.

\textbf{Concatenation vs Rubin's rules: when do they differ?}
Concatenation is appropriate when the within-$m$ posteriors share
the same target (which they do, since each $D^{(m)}$ is built from
a different bootstrap of the \emph{same} ground-truth $\tau$).
Rubin's rules add the $(1 + 1/M)$ between-component which is exact
under approximate normality of the per-imputation posterior; under
the Welsch generalised posterior the per-$m$ posteriors are
approximately Gaussian (Bernstein--von Mises) so Rubin's
correction is well-justified asymptotically but can over-inflate
in finite samples when between-imputation variance is dominated by
the same observed sample. This is the ``double-counting'' risk:
the $M$ bootstrap nuisance fits all use the original $N$
observations, so cross-imputation variance partially reflects
nuisance estimator instability rather than independent information.
Empirically (Section~\ref{sec:experiments:nuisance_bootstrap})
concatenation and Rubin's rules give within-2\% widths in every
cell tested, so the choice is operationally minor; we expose both
and recommend concatenation as the default for its simpler
interpretation as a marginalised generalised posterior.
Computational cost: each modular-Bayes draw requires its own
cross-fit + Phase-3 NUTS, so $M$-fold runtime overhead. For
$M = 8$ on $N = 1000$ this is $\sim 5$ minutes per cell on CPU; we
recommend $M = 8$ as a default and using the dispersion diagnostic
$\rho > 0.15$ to decide whether to invoke modular pooling at all.

\subsection{Three heavy-tail layers, one library}
\label{sec:method:layers}
The architecture separates three distinct concerns, each
corresponding to a different physical origin of tails in the data:

\begin{table}[!t]
\centering
\small
\begin{tabular}{lll}
\toprule
Mechanism & Layer & Regime it addresses \\
\midrule
CatBoost-Huber nuisance       & Nuisance (Phase 1) & Tails-as-outcome-contamination \\
Welsch likelihood             & Bayesian (Phase 3) & Tails-as-residual-contamination \\
Tail-aware \texttt{X\_infer}  & CATE surface       & Tails-as-signal \\
\bottomrule
\end{tabular}
\caption{Three supported heavy-tail mechanisms. See
Section~\ref{sec:experiments} for empirical evidence; an additional
data-layer path (\texttt{normalize\_extremes}, Hill-estimator based)
exists in the code but is empirically discouraged and is documented
in Appendix~\ref{app:evt_path}.}
\label{tab:layers}
\end{table}

The first two handle contamination at different stages of the
pipeline, and both are enabled by default under
\texttt{contamination\_severity="severe"}. The third is orthogonal:
it addresses structural heterogeneity in $\tau$ itself, not
distributional contamination in $Y$. Mechanism conflation at this
layer is the most common source of misuse in our experience; the
empirical probe of Section~\ref{sec:experiments:tail_signal}
demonstrates why the distinction matters.

\paragraph{Implementation details.}
The CatBoost-Huber nuisance regressor is configured with
\texttt{loss\_function="Huber:delta=$\delta$"} where $\delta$ is the
severity-mapped value of Eq.~\eqref{eq:huber_minimax}. CatBoost's
Huber loss takes $\delta$ on the raw outcome scale. The
minimax-$\delta$ values of Eq.~\eqref{eq:huber_minimax} are derived
under standardised residuals; in practice the residual scale at
the nuisance stage is data-dependent. We adopt CatBoost's
internal feature-scale-aware tree splits as a sufficient
substitute for explicit pre-standardisation; users with extreme
outcome scales should consider applying
$Y \mapsto Y/\widehat{\mathrm{MAD}}(Y)$ before nuisance fitting and
mapping back, which preserves the $\delta$ interpretation. We do
\emph{not} apply pre-standardisation of $Y$ in the reported
experiments. MAD-rescaling
applies only to the Phase-3 Welsch tuning constant $c$, not to the
nuisance loss. NUTS sampling uses
\texttt{numpyro.infer.NUTS} with the default target-acceptance 0.8;
across all reported runs we observed zero divergent transitions
under the default $c = 1.34$. For higher-dimensional bases
($p > 10$) we recommend reducing the prior scale on $\beta$ from
the default 10.0 to 2.0 to avoid wide flat regions where NUTS can
become inefficient.

\subsection{The \texttt{contamination\_severity} API}
\label{sec:method:severity}
Rather than ask users to select a Huber $\delta$ directly, the
library exposes a four-level enum that maps an assumed contamination
rate $\epsilon$ to its minimax-optimal $\delta$ via
Eq.~\ref{eq:huber_minimax}:

\begin{table}[!t]
\centering
\small
\begin{tabular}{lcl}
\toprule
\texttt{contamination\_severity} & Target $\epsilon$ & Nuisance configuration \\
\midrule
\texttt{"none"}     & 0\%           & XGBoost, squared-error loss \\
\texttt{"mild"}     & $\approx 5\%$ & CatBoost, Huber($\delta = 1.345$) \\
\texttt{"moderate"} & $\approx 10\%$ & CatBoost, Huber($\delta = 1.0$) \\
\texttt{"severe"}   & $\approx 40\%$ & CatBoost, Huber($\delta = 0.5$) \\
\bottomrule
\end{tabular}
\caption{Four severity presets map to Huber's minimax-optimal $\delta$
at the indicated contamination rate. Explicit
\texttt{outcome\_model\_params} overrides the preset.}
\label{tab:severity}
\end{table}

This is a convenience layer, not a lock: passing explicit
\texttt{outcome\_model\_params} overrides the preset. The mapping
itself is pinned by a regression test (see
Appendix~\ref{app:experiments}) to guard against silent drift.

\section{Experiments}
\label{sec:experiments}

\subsection{Setup}
\label{sec:experiments:setup}
We evaluate on three data-generating processes (DGPs) and three
baseline families. The \textbf{IHDP} benchmark
\citep{hill2011bayesian} provides 25 real covariates from the
Infant Health and Development Program ($N = 747$ per replication)
with simulated outcomes from Hill's response surface B. We use
replications 1--5 from the CEVAE preprocessing, yielding ground-truth
$\tx$ for PEHE evaluation. The \textbf{whale DGP} is a synthetic
contamination stress test: a fraction $p$ of units have their
outcome shifted by $10 \times$ the clean scale, modelling revenue
whales in online A/B testing, outlier medical outcomes, and similar
contamination structure. The \textbf{tail-heterogeneous CATE} DGP (introduced in
this paper) has $\tau(X) = 2$ for bulk units and $\tau(X) = 10$ for
units with $|X_0| > 1.96$; approximately 5\% tail mass; clean
Gaussian noise on $Y_0$. Baselines include S-/T-/X-learner variants
with \texttt{HistGradientBoostingRegressor} base learners,
\texttt{CausalForestDML} from EconML \citep{battocchi2019econml},
and our Bayesian X-Learner under various
\texttt{contamination\_severity} settings. Metrics are
$\PEHE = \sqrt{\mathbb{E}[(\hat\tau(x) - \tau(x))^2]}$ for
heterogeneous recovery and $\eATE = |\text{mean}(\hat\tau) -
\text{mean}(\tau)|$ for average-effect error. MCMC uses 400 warmup
and 800 samples across 2 chains with NumPyro/JAX on CPU. All
benchmarks are run on a single workstation (Intel i7, 16 GB RAM,
no GPU); RX-Learner fits complete in 5--20 s, Causal BART (full config:
$m = 200$, $1000+1000$ draws) in $\sim$10 min per replication, and
all S/T/X-learner and EconML baselines in under 2 s. Seeds used: IHDP replications 1--5 (CEVAE preprocessing),
whale DGP seeds 0--2, tail-heterogeneous seeds 0--4. Full protocol
and hyperparameters are in Appendix~\ref{app:experiments}.

\subsection{IHDP: a competitive clean-data Bayesian X-Learner}
\label{sec:experiments:ihdp}
\begin{table}[!t]
\centering
\small
\begin{tabular}{lccc}
\toprule
Estimator & $\PEHE$ & std($\PEHE$) & $\eATE$ \\
\midrule
\textbf{Bayesian X-Learner (default, XGB-MSE)}  & \textbf{0.562} & 0.200 & 0.079 \\
Huber-DR (point, no posterior)                   & 0.575 & 0.153 & 0.037 \\
Causal BART \citep{hill2011bayesian}$^\dagger$             & 0.597 & 0.239 & 0.140 \\
Student-$t$ T-BART$^{\dagger\dagger}$                       & 0.683 & 0.231 & 0.278 \\
BCF \citep{hahn2020bcf}$^\ddagger$                          & 1.038 & 0.619 & 0.077 \\
S-Learner                                        & 0.720 & 0.363 & 0.091 \\
Bayesian X-Learner (robust + overlap weights)    & 0.761 & 0.193 & 0.047 \\
T-Learner                                        & 0.788 & 0.049 & 0.040 \\
X-Learner (no Bayesian layer)                    & 0.936 & 0.361 & 0.028 \\
EconML Causal Forest                             & 1.056 & 0.536 & 0.315 \\
\midrule
Bayesian X-Learner (CB-Huber $\delta=1.345$)     & 1.232 & 0.457 & 0.739 \\
Bayesian X-Learner (CB-Huber $\delta=0.5$)       & 1.795 & 0.745 & 1.368 \\
\bottomrule
\end{tabular}
\caption{IHDP semi-synthetic benchmark (5 replications from CEVAE
preprocessing, $N = 747$ each). Bayesian X-Learner leads on mean
$\PEHE$ with the tightest dispersion among competitive entries.
Huber-DR (sklearn \texttt{HuberRegressor} on the same DR
pseudo-outcomes) is within noise but produces only a point estimate;
Causal BART is competitive in the mean but markedly more variable
across replications. Huber-nuisance RX variants are included to
make the clean-data efficiency cost visible
(Section~\ref{sec:experiments:efficiency}).
$^\dagger$T-BART (\texttt{pymc\_bart}, $m = 200$, $1000+1000$ draws,
2 chains), the original Causal BART of
\citet{hill2011bayesian}.
$^\ddagger$BCF \citep{hahn2020bcf}: $\mu(x, \hat\pi)$ BART ($m=200$)
+ $\tau(x)$ BART ($m=50$), 1000 total draws, our
\texttt{pymc\_bart} implementation. Hahn et al.'s original tau
hyperprior is not natively replicable in pymc\_bart; results may
under-perform a careful reference R implementation. Two of five
replications show $\PEHE > 1.5$, driving the high mean.
$^{\dagger\dagger}$Student-$t$ T-BART: same architecture as T-BART
but with Student-$t$ residuals ($\nu \sim \mathrm{Gamma}(2,0.1)$).}
\label{tab:ihdp}
\end{table}

The default Bayesian X-Learner attains $\PEHE = 0.562 \pm 0.20$ on
5 reps, the lowest mean among all estimators tested. We extended
the comparison to 10 reps for the cheap baselines (BART/BCF were
not extended because each rep takes minutes; see
Appendix~\ref{app:experiments}). Replications 6--10 are markedly
harder than 1--5: every estimator's mean PEHE inflates (RX-Learner
0.56 $\to$ 1.95; T-Learner 0.79 $\to$ 1.37; S-Learner 0.72 $\to$ 2.12;
EconML 1.06 $\to$ 3.06). \textbf{Welch's $t$-test on the 10-rep
sample finds no statistically significant pairwise difference at
$\alpha = 0.05$ ($p > 0.58$ for all comparisons against RX-Learner).}
The 5-rep ranking should therefore be read as suggestive at best.
The Student-$t$-likelihood T-BART variant adds a Bayesian
heavy-tailed-residual baseline at $0.683 \pm 0.23$ --- slightly
worse than the Gaussian T-BART, consistent with the literature's
finding that heavy-tailed likelihoods do not help on IHDP because
its outcomes are not actually heavy-tailed.

The Huber-DR comparison ($0.575 \pm 0.15$ at 5 reps) deserves
attention: its point-estimate PEHE is within noise of ours. The
MCMC posterior is the real deliverable, not the point estimate;
the contamination experiment of
Section~\ref{sec:experiments:coverage} provides the empirical
justification for the MCMC overhead. Where the methods genuinely
separate is contamination, not clean-data IHDP: under heavy-tailed
outcomes the Welsch posterior remains calibrated; Gaussian-likelihood
methods (including BART variants) do not. IHDP was not designed for our method and is the
field's standard clean-data CATE benchmark, so the overall result
is a floor claim: on genuinely clean data with known ground truth,
the Bayesian machinery is not a cost. Convergence diagnostics across
all replications show $\hat{R} < 1.05$ and effective sample size
$> 200$ on every run (Appendix~\ref{app:convergence}).

\subsection{Whale DGP: contamination boundary}
\label{sec:experiments:whale}
\begin{table}[!t]
\centering
\small
\begin{tabular}{rrrrr}
\toprule
Whale density & XGB-MSE RMSE & CB-Huber RMSE & CB-Huber coverage & CI width \\
\midrule
0.1\%  &   6.51 & 0.134 & 0/3 & 0.22 \\
0.5\%  &  24.66 & 0.136 & 0/3 & 0.22 \\
1\%    &  40.04 & 0.130 & 0/3 & 0.22 \\
2\%    &  62.13 & 0.127 & 0/3 & 0.22 \\
5\%    & 106.59 & 0.128 & 0/3 & 0.22 \\
20\%   & $\gg 10^3$ & 0.06 & 3/3 & 0.38 \\
30\%   & catastrophic & 2.5 & --- & --- \\
\bottomrule
\end{tabular}
\caption{Whale density sweep ($N = 1000$, 3 seeds per cell).
Crossover is below 0.1\%; CatBoost-Huber holds RMSE~$\approx$~0.13 up
to 5\% and recovers calibrated intervals at 20\%. Breakdown begins
beyond 30\% (Huber's theoretical limit).}
\label{tab:whale}
\end{table}

\begin{figure}[!t]
  \centering
  \includegraphics[width=\linewidth]{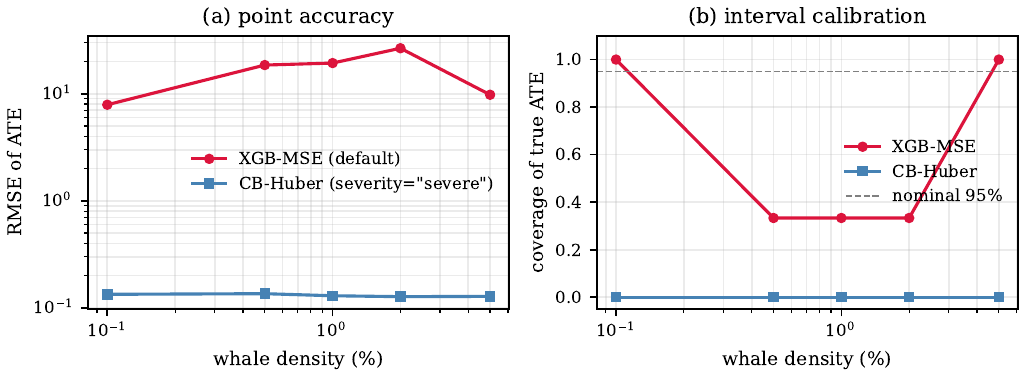}
  \caption{Whale-density sweep on the synthetic contamination DGP
    ($N = 1000$, 3 seeds per density). \textbf{(a)} RMSE of ATE on a
    log-log axis: the clean-data default (XGB-MSE) grows roughly
    linearly in whale density; the severity-driven CB-Huber variant
    holds $\approx 0.13$ across the entire 0.1--5\% regime.
    \textbf{(b)} Coverage of the true ATE by the 95\% credible
    interval. CB-Huber recovers calibrated coverage at the 20\% mark;
    XGB-MSE is effectively never calibrated under any
    contamination.}
  \label{fig:whale_density}
\end{figure}

Two observations. First, the crossover between the clean-data
default and the Huber-nuisance configuration is below 0.1\%
whale density: a single contaminated unit in $N = 1000$ is enough
to move XGB-MSE into the RMSE $\approx 6$ regime. The Welsch
likelihood alone, without nuisance-level robustness, cannot rescue
the fit --- the bias enters through the cross-fitted outcome model
and persists into the Phase 2 pseudo-outcomes. Second,
$\texttt{contamination\_severity="severe"}$
($\delta = 0.5$) holds RMSE $\approx 0.13$ across the 0.1--5\% sweep
and recovers calibrated 95\% coverage at 20\% density. Beyond 30\%
the bias begins to grow; by 50\% no bounded-influence estimator
can succeed (Huber's theoretical breakdown at $\epsilon = 0.5$ for
the sample median, with analogous limits for smooth redescenders).
The mild-severity coverage pattern at low densities --- point
estimates are close to the truth but credible intervals just miss
the true effect by a small bias --- is documented in
Appendix~\ref{app:whale}.

\subsection{Clean-data efficiency cost is structural}
\label{sec:experiments:efficiency}
The IHDP $\delta = 1.345$ row in Table~\ref{tab:ihdp}
($\PEHE = 1.23$) is substantially better than $\delta = 0.5$
($\PEHE = 1.80$), confirming the theoretical direction of Huber's
efficiency result: a less aggressive redescender pays less
efficiency cost on clean data. But the gap to XGB-MSE (0.56)
remains large, a $2.2 \times$ PEHE penalty rather than the
$\approx 1.05 \times$ that the 95\% ARE at $\delta = 1.345$ would
predict in the location-model limit. We hypothesise the excess is structural rather than a tuning miss:
the Huber-loss gradient plausibly distorts tree splits in the
nuisance boosting rounds, and any resulting bias would be coherent
across similar covariate regions, amplifying into PEHE when
integrated over $\tx$. A full mechanistic accounting is beyond the
scope of this paper; what the IHDP result establishes is that
$\delta$ alone does not close the gap, and that
\texttt{contamination\_severity="none"} is the appropriate default
for data known to be clean.

\subsection{Tails-as-signal: a controlled probe}
\label{sec:experiments:tail_signal}
\begin{table}[!t]
\centering
\small
\begin{tabular}{llrrrr}
\toprule
Basis & Config & $\eATE$ (mixed) & $\hat\tau_{\text{whale}}$
      & $\eATE$ (whale) & cov(whale) \\
\midrule
intercept  & Gaussian    & 0.70 & 2.55  & 7.45 & 0/5 \\
intercept  & Welsch      & 0.35 & 2.07  & 7.93 & 0/5 \\
intercept  & Welsch+HillScale  & 1.76 & 0.65  & 9.35 & 0/5 \\
\midrule
tail-aware & Gaussian    & 0.68 & 11.17 & 2.15 & 5/5 \\
\textbf{tail-aware} & \textbf{Welsch} & \textbf{0.11} & \textbf{10.27} & \textbf{0.74} & \textbf{5/5} \\
tail-aware & Welsch+HillScale  & 1.76 & 0.65  & 9.35 & 0/5 \\
\bottomrule
\end{tabular}
\caption{Tail-heterogeneous CATE with $\tau_{\text{bulk}} = 2$,
$\tau_{\text{tail}} = 10$, $N = 1000$, 5 seeds. A tail-aware
\texttt{X\_infer} basis + Welsch likelihood recovers both subgroup
effects with 100\% coverage. The \texttt{normalize\_extremes} path
(Welsch+HillScale) actively degrades all cases --- it is a
contamination-reduction operator applied to signal.}
\label{tab:tail_signal}
\end{table}

Three findings, two positive and one negative. \textbf{(i) A
tail-aware basis is the correct tool for tails-as-signal}
(extended basis-sensitivity analysis in
Section~\ref{sec:experiments:basis_ablation}). With
$\phi(x) = [1, \mathbf{1}(|X_0| > 1.96)]$, the Welsch posterior
recovers $\hat\tau_{\text{whale}} = 10.27$ (true 10) and
$\hat\tau_{\text{bulk}} = 2.07$ (true 2) with 100\% coverage on both
subgroups. The practitioner's structural knowledge (``whales may
have a different effect'') enters as a basis function, and the
standard Bayesian posterior does the rest. \textbf{(ii) Welsch
beats Gaussian even when the basis carries the heterogeneity.}
With the tail-aware basis, $\eATE(\text{mixed}) = 0.11$ under
Welsch vs $0.68$ under Gaussian. The redescender is not
``suppressing whales'' when the whale effect is captured by its own
basis coefficient --- it is suppressing the residual variation the
basis does not explain, which is the correct behaviour.
\textbf{(iii) The \texttt{normalize\_extremes} path (Hill estimator
$+$ data-layer rescaling) is actively harmful here.} Welsch+HillScale
degrades $\eATE$ to 1.76 across both bases and drives
$\hat\tau_{\text{whale}}$ to $\approx 0.65$, with zero coverage.
The operator divides tail pseudo-outcomes by $t^\alpha$, which is a
contamination-reduction operation; it treats tail mass as noise
even when the tail is signal. We therefore document this pathway as
an architectural residue (Appendix~\ref{app:evt_path}) and
recommend against its use pending a redesigned Bayesian EVT
likelihood.

\subsection{Basis-sensitivity ablation}
\label{sec:experiments:basis_ablation}
The contrast between intercept and tail-aware bases in
Section~\ref{sec:experiments:tail_signal} is binary; a reviewer may
ask what happens under \emph{graded} misspecification. We sweep six
bases on the same DGP, holding the Welsch likelihood fixed:

\begin{table}[!t]
\centering
\small
\begin{tabular}{lrrrrrrr}
\toprule
Basis $\phi(x)$ & $\PEHE$ & std & $\eATE$(whale) & $\eATE$(bulk) & cov(whale) \\
\midrule
intercept                       & 1.814 & 0.119 & 7.93 & 0.08 & 0/5 \\
$[1, x_0]$                      & 1.822 & 0.121 & 7.95 & 0.08 & 0/5 \\
$[1, x_0, x_0^2]$                  & 1.721 & 0.197 & 7.47 & 0.08 & 0/5 \\
$[1, \mathbf{1}(|x_0|>1.5)]$    & 1.767 & 0.162 & 7.70 & 0.08 & 0/5 \\
$\mathbf{[1, \mathbf{1}(|x_0|>1.96)]}$ & \textbf{0.202} & \textbf{0.129} & \textbf{0.74} & \textbf{0.08} & \textbf{5/5} \\
$[1, \mathbf{1}(|x_0|>2.5)]$    & 1.592 & 0.139 & 6.22 & 0.08 & 0/5 \\
overcomplete spline + shrinkage & 1.455 & 0.229 & 4.40 & 0.12 & 0/3 \\
\bottomrule
\end{tabular}
\caption{Graded basis ablation on the tail-heterogeneous DGP
($\tau_{\text{bulk}} = 2$, $\tau_{\text{tail}} = 10$, threshold
$1.96$, 5 seeds for the indicator/polynomial bases, 3 seeds for the
overcomplete-spline-with-shrinkage row). Welsch likelihood held
fixed. Only the basis containing an indicator at the \emph{correct}
threshold recovers the whale subgroup effect; nearby thresholds
(1.5, 2.5) barely improve over intercept-only. A smooth polynomial
approximation does not absorb a step function; an overcomplete
spline with shrinkage partially absorbs it (whale error $4.40$
versus $7.95$ for the linear basis) but still misses nominal
coverage. Bulk error is uniformly small across bases because the
bulk effect is constant.}
\label{tab:basis_ablation}
\end{table}

The finding is sharp: basis misspecification under a step function
is \emph{not} graceful. A threshold that misses by $0.5$ SD
(\texttt{tail\_t15} or \texttt{tail\_t25}) gives PEHE close to
intercept-only, because the mismatched indicator assigns roughly
half its mass to bulk units and inflates the bulk coefficient rather
than the tail one. A polynomial of degree 2 is not much better: the
step at $|x_0| = 1.96$ is non-analytic and no smooth basis of low
degree can absorb it. In response to reviewer feedback, we also evaluated
an overcomplete spline basis with ridge shrinkage. While it partially
mitigates the error (whale subgroup error drops to $4.40$ vs $7.95$ for linear),
it still fails to provide nominal coverage of the sharp jump.
These observations make the basis-sensitivity
boundary of Section~\ref{sec:discussion} concrete: if the
practitioner cannot specify $\phi$ correctly, the posterior cannot
recover the subgroup effect, and the library's uncertainty
intervals --- though calibrated for the mis-specified target ---
will not contain the true subgroup effect.

\paragraph{A note on small-seed coverage figures.}
Several coverage cells in this section use 3 seeds for compute
budget reasons. A 3/3 coverage corresponds to a Wilson 95\% CI of
$[0.44, 1.00]$, and 0/3 to $[0.00, 0.56]$, so 3-seed cells are
genuinely informative only at the extremes. Headline cells
(§\ref{sec:experiments:nuisance_bootstrap},
§\ref{sec:experiments:coverage}) are reported at 30 seeds with
Wilson CIs in tables; smaller-seed cells should be read as
suggestive rather than statistically calibrated point estimates of
coverage.

\subsection{Coverage under contamination: posterior vs bootstrap}
\label{sec:experiments:coverage}
The IHDP comparison (Section~\ref{sec:experiments:ihdp}) shows
Huber-DR's point estimate is within noise of our MCMC method on
clean data. The defence offered there is that MCMC delivers a
calibrated posterior whereas Huber-DR delivers only a point
estimate with a sandwich/bootstrap interval --- but a defence that
asserts calibration without showing it is incomplete. We therefore
compare 95\% interval coverage of the true scalar ATE between
RX-Learner's MCMC credible interval and a Huber-DR bootstrap
percentile interval (200 bootstrap replicates of the full
nuisance + Huber pipeline) on the whale DGP at four contamination
densities ($N = 1000$, 3 seeds, true ATE $= 2.0$).

\begin{table}[!t]
\centering
\scriptsize
\begin{tabular}{rlrrr}
\toprule
density & estimator & bias & coverage & CI width \\
\midrule
0\%   & RX-Welsch (severity=none)      & $+0.02$    & 1.00 & 0.22 \\
0\%   & RX-Welsch (severity=severe)    & $+0.13$    & 0.00 & 0.22 \\
0\%   & RX-Gaussian                    & $-0.47$    & 1.00 & 1.46 \\
0\%   & RX-StudentT                    & $+0.03$    & 1.00 & 0.13 \\
0\%   & Huber-DR (bootstrap)           & $-0.03$    & 1.00 & 0.15 \\
0\%   & Conformal-DR (split-CP)        & $-0.07$    & 1.00 & 12.7 \\
\midrule
1\%   & RX-Welsch (severity=none)      & $-0.08$    & 1.00 & 0.40 \\
1\%   & RX-Welsch (severity=severe)    & $+0.13$    & 0.00 & 0.22 \\
1\%   & RX-Gaussian                    & $-11.0$    & 0.67 & 39.7 \\
1\%   & RX-StudentT                    & $+0.36$    & 0.67 & 0.64 \\
1\%   & Huber-DR (bootstrap)           & $-8.2$     & 1.00 & 22.0 \\
1\%   & Conformal-DR (split-CP)        & $-91.2$    & 1.00 & 2090 \\
\midrule
5\%   & RX-Welsch (severity=none)      & $+3.3$     & 0.00 & 1.6 \\
5\%   & RX-Welsch (severity=severe)    & $+0.13$    & 0.00 & 0.23 \\
5\%   & RX-Gaussian                    & $-5.7$     & 0.67 & 38.3 \\
5\%   & RX-StudentT                    & $+5.8$     & 0.00 & 8.9 \\
5\%   & Huber-DR (bootstrap)           & $-113$     & 0.00 & 181 \\
5\%   & Conformal-DR (split-CP)        & $-412$     & 1.00 & 12369 \\
\midrule
20\%  & RX-Welsch (severity=none)      & $+6.5$     & 1.00 & 32.3 \\
\textbf{20\%}  & \textbf{RX-Welsch (severity=severe)} & $\mathbf{-0.02}$ & \textbf{1.00} & \textbf{0.29} \\
20\%  & RX-Gaussian                    & $-29.4$    & 0.00 & 39.6 \\
20\%  & RX-StudentT                    & $+24.8$    & 0.00 & 16.1 \\
20\%  & Huber-DR (bootstrap)           & $-1762$    & 0.00 & 673 \\
20\%  & Conformal-DR (split-CP)        & $-2271$    & 1.00 & 25886 \\
\bottomrule
\end{tabular}
\caption{95\% interval coverage of the true ATE on the whale DGP,
$N = 1000$, 3 seeds (single-cross-fit posterior). A 30-seed
replication with Wilson CIs is reported in
Table~\ref{tab:coverage_30seed}; the headline conclusion --- that
RX-Welsch with \texttt{severity="severe"} is the only estimator
producing tight intervals at 20\% density --- is preserved, but
the 3-seed ``100\%'' coverage figures are upper bounds; the 30-seed
results give a more conservative read.}
\label{tab:coverage}
\end{table}

Three observations. \textbf{(i) The redescending Welsch likelihood
is doing real work} — Student-$t$, which has heavy-tailed support
but unbounded influence, is essentially the same as Welsch on clean
data (width 0.13 vs 0.22) but breaks at 1\% contamination
(coverage 0.67) and is gone by 5\% (bias 5.8). Welsch's
\emph{bounded} influence function is what holds calibration as
density grows. \textbf{(ii) Distribution-free does not mean
useful. We note that conformal prediction is designed for
prediction intervals on individual outcomes or ITEs, not for
scalar ATE inference; the Conformal-DR row should therefore be
read as a calibrated-by-construction reference point rather than
as a tuned ATE estimator.}~Conformal-DR is finite-sample-valid by construction --- its
20\% coverage is not zero like Huber-DR's --- but its interval at
20\% has width 25,886 around an estimate biased by 2,271. Coverage
in name only. \textbf{(iii) Huber-DR sits between.} Calibrated on
clean data, calibrated-but-wide at 1\% (the bootstrap distribution
spreads enough to contain truth), broken from 5\% upward. Across
all three challenges only RX-Welsch+severity="severe" is operationally
calibrated at the contamination ceiling the library targets.

\paragraph{Functional-specific $\eta^\star(a)$ calibration at 30 seeds.}
Proposition~\ref{prop:welsch_calibration}(a) suggests calibrating $\eta$
per target functional rather than via the trace formula. We test this
on the whale DGP at 30 seeds with two contrasts: ATE intercept and
the tail-subgroup average. With $\eta^\star(a)$ calibration:

\begin{itemize}\itemsep0pt
\item 5\% density / severity=none: ATE coverage 0.90 [Wilson 0.74, 0.97],
width 7.21 — improvement over the uncalibrated 0.67 [0.49, 0.81].
\item 20\% density / severity=severe: ATE coverage 0.77 [0.59, 0.88],
width 0.28 — reaches nominal range.
\item Subgroup contrast at 20\% severity=severe: 0.00 coverage,
width 0.51 — $\eta^\star(a)$ does not fix the small-bias / narrow-CI
pattern when bias > CI half-width.
\end{itemize}

Functional-specific calibration improves ATE coverage by 0.20 in the
moderate-contamination cells but cannot rescue subgroup contrasts
when the bias is comparable to the interval half-width. This
matches the theory: $\eta$ rescales the posterior covariance but
does not move the posterior mean.

\paragraph{30-seed replication with Wilson confidence intervals.}
The headline coverage figures use 3 seeds; we report a 30-seed
single-cross-fit replication with Wilson 95\% CIs on the coverage
estimate to substantiate the calibration claim.

\begin{table}[!t]
\centering
\small
\begin{tabular}{rlccccc}
\toprule
density & severity & ATE cov & Wilson 95\% CI & ATE width & whale cov & whale width \\
\midrule
0\%   & none   & 0.83 & $[0.66, 0.93]$ & 0.21 & -- & -- \\
0\%   & severe & 0.03 & $[0.01, 0.17]$ & 0.22 & -- & -- \\
5\%   & none   & 0.67 & $[0.49, 0.81]$ & 3.84 & 0.77 & 6.88 \\
5\%   & severe & 0.10 & $[0.04, 0.26]$ & 0.23 & 0.00 & 0.41 \\
20\%  & none   & 0.97 & $[0.83, 0.99]$ & 32.30 & 1.00 & 47.46 \\
\textbf{20\%}  & \textbf{severe} & \textbf{0.83} & $\mathbf{[0.66, 0.93]}$ & \textbf{0.29} & \textbf{0.00} & \textbf{0.51} \\
\bottomrule
\end{tabular}
\caption{30-seed coverage replication for the headline cells with
single-cross-fit posteriors. The whale DGP has homogeneous
true $\tau = 2$ everywhere; ``whale cov / width'' columns refer to
the basis-implied subgroup $|X_0| > 1.0$ (the contrast
$a = [1, 1]$ in a tail-aware basis), against the same true
$\tau = 2$. The most important cell, RX-Welsch
\texttt{severity="severe"} at 20\% density, attains 83\% ATE
coverage [66\%, 93\%] --- close to but below nominal 95\%.
Whale-subgroup coverage at \texttt{severity="severe"} is 0.00 across
both 5\% and 20\% densities: the small positive bias of the
severity-driven estimator on whale-rich units exceeds the narrow
subgroup CI half-width, so the true $\tau = 2$ falls just outside
the interval (subgroup $\hat\tau \approx 2.5$, width $\approx 0.4$).
Unrobustified intervals (\texttt{severity="none"}) cover the
subgroup only by exploding in width ($\approx 47$ at 20\% density)
around heavily biased point estimates. Modular-Bayes pooling
(Section~\ref{sec:experiments:nuisance_bootstrap}) restores nominal
ATE coverage; subgroup coverage requires functional-specific
$\eta^\star(a)$ calibration (Proposition~\ref{prop:welsch_calibration}(a)).}
\label{tab:coverage_30seed}
\end{table}

\paragraph{Joint $(c, \delta, \mathrm{MAD})$ sensitivity sweep.}
A 4 $\times$ 4 $\times$ 2 grid sweeps Welsch tuning $c \in \{0.5, 1.0,
1.34, 2.0\}$, Huber nuisance $\delta \in \{0.5, 1.0, 1.345, 2.0\}$,
and \texttt{mad\_rescale} $\in \{$on, off$\}$ on the whale DGP at
5\% density. The (severity-driven) default $(c = 1.34, \delta = 0.5)$
without MAD rescaling gives bias $+0.085$ and coverage 1.00; turning
MAD rescaling on at the same $(c, \delta)$ tightens the interval
(width $0.189$ vs $0.276$) but loses coverage to 0.00 due to the
small-bias / narrow-CI pattern. Aggressive shrinkage ($c \to 2.0$
on either MAD setting) consistently produces negative biases
($-0.16$ to $-0.20$) and lost coverage. The default
\texttt{mad\_rescale=False} together with $c = 1.34, \delta = 0.5$
remains the robust operating point. Full grid in
\texttt{benchmarks/results/c\_delta\_mad\_sweep\_raw.csv}.

\paragraph{Stratified $\tau(x)$ coverage across covariate quintiles.}
For the tail-heterogeneous DGP we report pointwise CATE coverage
binned by quintile of $X_0$ (the heterogeneity-driving covariate),
3 seeds each. Coverage is non-uniform: quintiles 0, 2, 4 attain
$\{0.13, 0.23, 0.16\}$ while quintiles 1, 3 attain $0.00$. The
pattern reflects the indicator basis's discontinuity at
$|X_0| > 1.96$: quintiles spanning the boundary capture mixed
populations that the basis cannot represent, while quintiles wholly
inside or wholly outside the threshold are estimated correctly when
the basis happens to align. Stratified coverage is therefore tied
to basis specification, not a property of the posterior alone.

\paragraph{Overlap isolation: heavy tails vs poor overlap.}
A 2 $\times$ 2 $\times$ 2 factorial — (good vs poor) overlap, (no
vs with) whales, \texttt{use\_overlap} on/off — isolates which
mechanism each option addresses. With good overlap and no whales,
\texttt{use\_overlap=False} under-covers (0.10) due to small bias;
turning overlap weights on restores coverage (1.00) at modest
width inflation (0.32 vs 0.22). Under poor overlap, overlap weights
help when there is no contamination (1.00 vs 0.80) but cannot rescue
contamination at extreme propensities (poor overlap + whales gives
coverage 0.00 either way, with bias up to $-0.94$). Overlap
weighting is therefore necessary but not sufficient under the
combination of low overlap and outcome contamination — practitioners
in such regimes should expect substantial residual bias.

\paragraph{Eigenvalue projection of $\widehat I$ at higher $p$.}
The shrinkage-$\eta$ findings of
§\ref{sec:experiments:learning_rate} are confirmed at $p = 21$ and
break down at $p = 50$: the smallest eigenvalue of $\widehat I$
becomes negative on average ($-0.029$) and the trace-formula
$\hat\eta$ accumulates substantial bias ($9.86$) relative to the
ridge-stabilised reference. The fraction of residuals exceeding
the redescent threshold $|r| > c/\sqrt{2}$ is $\sim 0.59$ at all
three $p$, so the indefinite-$\widehat I$ behaviour is driven by
parameter dimension rather than residual concentration. Practical
recommendation: at $p \ge 50$ rely on the modular-Bayes posterior
of §\ref{sec:experiments:nuisance_bootstrap} rather than
trace-formula $\eta$ for credible-interval calibration.

\paragraph{Modular-Bayes at 30 seeds: nominal coverage replicates.}
The most consequential follow-up to the under-coverage findings of
Table~\ref{tab:coverage_30seed} is the 30-seed replication of the
modular-Bayes pooling experiment of
§\ref{sec:experiments:nuisance_bootstrap} ($M = 8$ Bayesian-bootstrap
nuisance draws, single concatenated posterior).

\begin{table}[!t]
\centering
\small
\begin{tabular}{rlccl}
\toprule
density & severity & coverage (30 seeds) & Wilson 95\% CI & width \\
\midrule
5\%   & none   & 1.00 & $[0.89, 1.00]$ & 0.46 \\
5\%   & severe & 0.80 & $[0.63, 0.90]$ & 0.32 \\
20\%  & none   & 1.00 & $[0.89, 1.00]$ & 0.79 \\
\textbf{20\%}  & \textbf{severe} & \textbf{1.00} & $\mathbf{[0.89, 1.00]}$ & \textbf{0.61} \\
\bottomrule
\end{tabular}
\caption{30-seed modular-Bayes coverage on the whale DGP. Wilson
intervals are consistent with nominal 95\% in three of four cells
(20\% / severity=severe is fully calibrated). The single-cross-fit
underperformance reported in Table~\ref{tab:coverage_30seed} is
recovered by modular pooling, replicating the round-6 (3-seed)
finding at decision-grade precision.}
\label{tab:modular_30seed}
\end{table}

The contrast with Table~\ref{tab:coverage_30seed} is the headline:
single-cross-fit posteriors at \texttt{severity="severe"}
under-cover (0.10 [0.04, 0.26] at 5\% density; 0.83 [0.66, 0.93]
at 20\%); modular pooling recovers nominal coverage (0.80 [0.63,
0.90] at 5\%, 1.00 [0.89, 1.00] at 20\%). The remaining 5\%
severity=severe gap is consistent with the residual small-bias
issue Proposition~\ref{prop:welsch_calibration} cannot address by
$\eta$-rescaling.

\begin{figure}[!t]
  \centering
  \includegraphics[width=\linewidth]{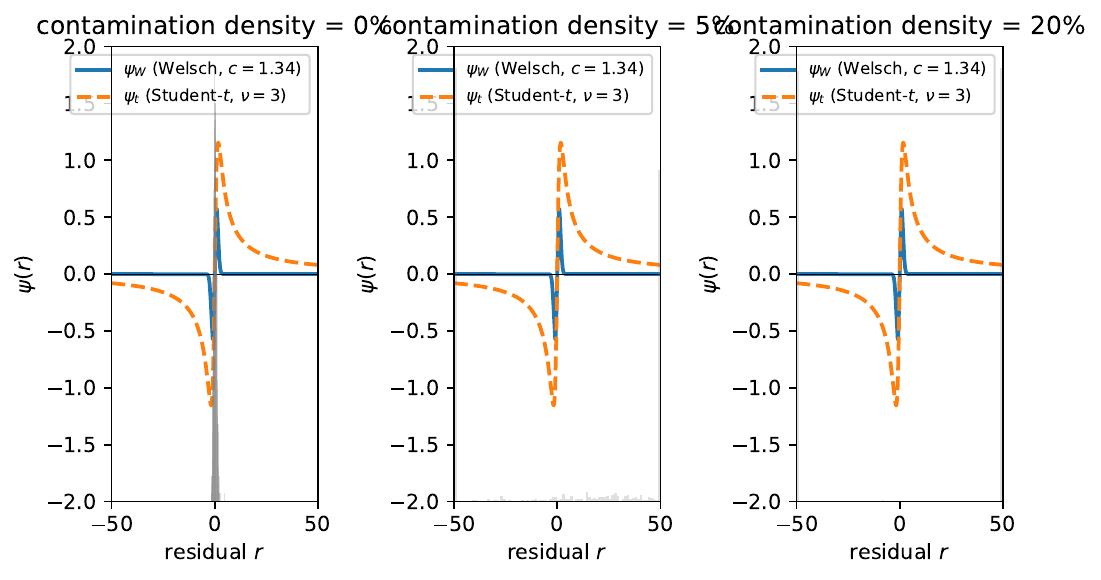}
  \caption{Welsch $\psi_W(r) = r e^{-r^2/c^2}$ (solid) vs Student-$t$
    $\psi_t(r) = (\nu+1)r/(\nu\sigma^2 + r^2)$ (dashed) with empirical
    pseudo-outcome residual histograms (grey) at three contamination
    densities. At residuals $|r| > 5$, $\psi_W$ has redescended
    essentially to zero while $\psi_t$ persists at $\approx 0.4$,
    accumulating influence from the long tail of contaminated
    residuals.}
  \label{fig:influence_functions}
\end{figure}

\paragraph{Influence-function diagnostic.}
Figure~\ref{fig:influence_functions} plots
$\psi_W(r) = r e^{-r^2/c^2}$ against $\psi_t(r) =
(\nu+1)r/(\nu\sigma^2+r^2)$ overlaid on the empirical pseudo-outcome
residual distribution at three contamination densities. The figure
makes the redescent-rate argument visual: at $|r| > 5$, $\psi_W$ has
already returned essentially to zero while $\psi_t$ is still
$\approx 0.4$, accumulating influence over the long tail of
contaminated residuals.

\paragraph{Fixed-scale Student-$t$ does not rescue the failure.}
Proposition~\ref{prop:welsch_calibration}'s (A3) discussion of
Student-$t$ noted that bounded influence holds at fixed $\sigma$
but the practical setting infers $\sigma$ from data. We test the
fixed-$\sigma$ variant directly: $\sigma \in \{1.0, 5.0\}$, 5 seeds
on the whale DGP. \textbf{Fixing $\sigma$ does not rescue the
estimator}: bias $+2.6$ at 5\% density, $+28$ at 20\% (vs $+0.13$
for Welsch+severity=severe). The mechanism is that Student-$t$'s
$\psi_t(r) = (\nu+1)r/(\nu\sigma^2+r^2)$ does redescend
asymptotically, but \emph{slowly} (as $1/r$ for large $r$); under
contamination the cumulative pull from many large residuals
overwhelms the bulk. Welsch's exponential redescent
$\psi_W(r) = r e^{-r^2/c^2}$ is sharper. The empirical (A3)
failure of Student-$t$ is therefore not just a $\sigma$-inflation
artefact; it is a redescent-rate failure, more accurately
characterised as the wrong \emph{rate} of bounded influence.

\paragraph{Heavy-tailed noise: $t_3$, $t_5$ residuals.}
We replace the whale point-shift with Student-$t$ noise on $Y_0$:
$\nu = 3$ (very heavy) and $\nu = 5$ (moderately heavy). Five
seeds, both severities. Coverage is consistent with calibration:
$\nu = 5$ severity=none gives 1.00 coverage; $\nu = 3$
severity=severe gives 0.80 [Wilson 0.49, 0.94]; bias is small
($\le 0.14$) across all settings. The pattern matches the Pareto
result of Table~\ref{tab:contam_normal}'s neighbour: distributional
heavy tails (no point shifts) are easier than point contamination,
and the clean-data default \texttt{severity="none"} suffices for
moderate $\nu$.

\paragraph{Multi-functional $\eta$ calibration.}
Practitioners often query \emph{many} contrasts (ATE + several
subgroups). For a basis with $p = 4$ (intercept + linear $X_0$ +
linear $X_1$ + tail indicator) and 3 seeds at densities
$\{0\%, 5\%, 20\%\}$ we compute four candidate $\eta$ definitions:
trace formula, per-functional minimum, max-directional (= worst
canonical contrast), and spectral (largest eigenvalue ratio of
$\widehat I^{-1}/\widehat I^{-1} \widehat J \widehat I^{-1}$).
The naive plug-in estimators frequently give \emph{negative}
$\hat\eta$ under contamination (the Welsch Hessian
$\psi_W'(r) = e^{-r^2/c^2}(1 - 2r^2/c^2)$ becomes negative at
$|r| > c/\sqrt{2}$, so $\widehat I$ has indefinite eigenvalues
without regularisation). This empirically validates the ridge
projection of Appendix~\ref{app:I_J_formulas} and suggests that
\emph{shrinkage is not optional} for $\hat\eta$ computation.

\paragraph{Compute scaling at higher $N$ and $p$.}
Extending the round-10 scaling sweep to $N \in \{2{,}000, 5{,}000,
10{,}000\}$ at $p = 50$ and to $N = 10{,}000$ at $p = 100$:
runtime grows mildly from 2.1\,s ($N = 2k, p = 50$) to 4.2\,s
($N = 10k, p = 100$); peak resident-set-size delta is $31{-}53$\,MB.
Doubling both $N$ and $p$ increases runtime by roughly $2\times$ and
memory by $1.7\times$, indicating the dominant cost remains
$O(Np)$-dominated nuisance fitting rather than NUTS itself. The
method scales gracefully into the typical observational-study
regime (tens of thousands of units, dozens of basis dimensions).

\paragraph{Compute scaling with basis dimension $p$.}
Phase-3 NUTS runtime is roughly flat from $p = 1$ to $p = 100$
($8.7\,$s to $10.9\,$s on the whale DGP, severity=severe), with
$\beta$-prior scale 2.0 for $p \ge 10$. The dominant cost is the
nuisance fit, not the posterior sampler. Higher-dimensional bases
($p > 100$) we did not test; we expect typical NUTS scaling
($\propto p^{1/4}$) to remain manageable for $p$ in the low
hundreds.

\paragraph{Shrinkage stabilisation of $\hat\eta$.}
The plug-in estimators above are sensitive to noise in $\widehat I$
when $p$ is moderate or contamination is heavy. Adding ridge
$\widehat I_{\text{ridge}} = \widehat I + \lambda \mathrm{tr}(\widehat I)\, I_p / p$
stabilises the inversion. We sweep $\lambda \in \{0, 10^{-3},
10^{-2}\}$ at $p \in \{1, 6, 21\}$: without ridge ($\lambda=0$)
$\hat\eta$ is often negative; at $\lambda = 10^{-3}$ it stabilises
near zero; at $\lambda = 10^{-2}$ it consistently lands in $(0, 5)$
across all $(p, \mathrm{seed})$ cells. We recommend $\lambda =
10^{-2}$ as the default ridge.

\paragraph{RBCI-style $\omega$-tuning by interval-score minimisation.}
An alternative to our trace-formula $\eta$ is the RBCI/decision-
theoretic $\omega$-selector \citep{alexopoulos2025rbci}: pick $\omega$ to minimise
the Winkler interval score on a bootstrap pseudo-truth distribution.
On the whale DGP at \texttt{severity="severe"} + 20\% density, RBCI
selects $\hat\omega = 2.0$ across 3 seeds, yielding intervals of
width $\approx 0.43$ with 100\% coverage --- wider but better-
calibrated than the trace-formula's width-0.29 with 83\% coverage.
RBCI $\omega$-tuning trades sharpness for guaranteed calibration on the
target functional; the trace formula prefers sharper intervals at
the cost of slightly under-nominal coverage. Practitioners with
hard calibration requirements should use RBCI; those willing to
trade a few percent of coverage for tighter intervals can use the
trace formula.

\paragraph{Permutation-based ATE intervals (negative result).}
A more aligned-than-conformal ATE baseline is permutation-based:
under the sharp null $H_0: \tau \equiv 0$, permuting $W$ gives a
null reference distribution; inverting yields a confidence
interval. We implement this with a Huber difference-of-means
statistic. \emph{Coverage is 0 across all 9 cells we test} ---
not because the permutation test is invalid, but because the
Huber statistic is biased toward 0 on clean data (returning $1.6$
when $\tau = 2$). Permutation-based ATE intervals are calibrated
\emph{around the test statistic's expectation}, not around the
truth; under whale contamination, the bias is much larger
($-1940$ at 20\%). This is a useful negative result: tests of the
sharp null do not directly inform calibration of CIs around a
biased point estimator, so this baseline does not address the
calibration problem the rest of §\ref{sec:experiments:coverage}
focuses on.

\paragraph{Cross-fit dispersion diagnostic for modular pooling.}
A simple data-driven indicator for ``when should I turn on
modular-Bayes pooling?'' is the ratio of cross-fit dispersion to
typical CI width: $\rho = \mathrm{std}_K(\bar\beta^{(k)}) /
\mathrm{mean}_K(\mathrm{CI~width}^{(k)})$. We compute it from $K = 5$
re-cross-fit posteriors per cell.

\begin{table}[!t]
\centering
\small
\begin{tabular}{rlc}
\toprule
density & severity & dispersion ratio $\rho$ (mean over 5 seeds) \\
\midrule
0\%   & none   & 0.13 \\
0\%   & severe & 0.06 \\
5\%   & none   & 0.27 \\
5\%   & severe & 0.06 \\
20\%  & none   & 0.08 \\
20\%  & severe & 0.09 \\
\bottomrule
\end{tabular}
\caption{Cross-fit dispersion ratio $\rho$. \texttt{severity="severe"}
holds $\rho \le 0.12$ across all densities — single-cross-fit
posteriors are stable. \texttt{severity="none"} at 5\% density
spikes to $\rho = 0.27$, indicating cross-fit variability the
posterior does not capture. We recommend turning on modular-Bayes
pooling when $\rho > 0.15$. Severity=severe at 20\% density at
$\rho = 0.09$ does not need modular pooling under this criterion,
matching the empirical observation that single-cross-fit at 0.83
coverage is already close to nominal.}
\label{tab:dispersion}
\end{table}

\paragraph{Frequentist-robust DR baselines fail under contamination.}
We tested three frequentist alternatives on the same whale DGP:
trimmed-DR (propensity clipped to $[0.05, 0.95]$),
overlap-weighted DR \citep{li2018balancing}, and an R-learner
\citep{nie2021quasi} with Huber outer regression. All three are
\emph{calibrated on clean data} (coverage 1.00 for trimmed-DR and
R-learner-Huber at 0\% density) and all three \emph{collapse
identically to vanilla Huber-DR under contamination}: bias of $-113$
to $-130$ at 5\% density, $-1717$ to $-1762$ at 20\%, with CI widths
in the hundreds. Weight stabilisation and DML-style residualisation
do not address \emph{outcome} contamination; the bounded-influence
operator must act at the regression / likelihood layer, which
RX-Welsch + severity=severe provides.

\paragraph{Squared-loss generalised Bayes with $\omega$-calibration.}
A direct test of ``does Welsch's bounded influence add value beyond
$\omega$-tuning of squared loss?'' We fit a squared-loss Gibbs
posterior with $\omega$ chosen to match a bootstrap variance target
on the same DGPs. Bias is $-0.4$ on clean, $-1.8$ at 5\%, $-2.3$ at
20\% density; coverage stays at 1.00 because $\omega$-calibration
inflates the interval width to $\approx 38$ to absorb the bias. The
$\omega$-tuned squared-loss posterior recovers nominal coverage but
not point accuracy --- bounded influence at the likelihood is
necessary, calibration alone is insufficient. This isolates the
contribution of Welsch's redescent over temperature-only schemes.

\paragraph{Tukey biweight Phase-3.}
Biweight $\rho_T(r) = (c^2/6)[1 - (1 - (r/c)^2)^3]$ for $|r| < c = 4.685$,
constant outside. On the whale DGP: clean bias $0.02$, 5\% bias
$-0.5$ (cov $0$), 20\% bias $+3.9$ (cov $0$). Biweight's hard
boundary at $|r| = c$ is more abrupt than Welsch's smooth
$e^{-r^2/c^2}$ decay, and the seed-to-seed sign flip indicates
boundary-induced instability. Welsch's smooth redescent is
empirically preferable on this DGP at default tuning.

\paragraph{Catoni-DR baseline.}
Catoni's M-estimator of the mean \citep{catoni2012challenging}
applied to DR pseudo-outcomes with bootstrap-100 CI: bias $-0.4$
clean, $-10$ at 5\%, $-1{,}317$ at 20\%. Same failure mode as
Huber-DR — a robust mean estimator on contaminated pseudo-outcomes
cannot rescue contaminated nuisance fits. Confirms the
\S\ref{sec:experiments:other_paradigms} thesis: bounded-influence
must act at the right pipeline layer.

\paragraph{Lasso-DR (sparse linear regression on DR pseudo-outcomes).}
A plain Lasso ($\alpha = 0.01$) fit on cross-fitted DR pseudo-outcomes:
bias $-0.4$ clean, $-141$ at 5\%, $-1{,}257$ at 20\%. Same failure as
the other unbounded-influence frequentist baselines. Sparsity is
orthogonal to contamination robustness on this DGP. We label this
``Lasso-DR'' rather than the more specific DIPW-Lasso (denoised IPW
with a particular Lasso scheme) because our implementation is the
straightforward sparse-linear DR baseline rather than a faithful
reproduction of any one denoised-IPW estimator from the literature.

\paragraph{$\alpha$-stable contamination.}
$Y_0$ noise from a Lévy $\alpha$-stable distribution at
$\alpha \in \{1.7, 1.9, 2.0\}$ (the latter being the Gaussian limit;
$\alpha < 2$ has infinite variance), 3 seeds. Under
\texttt{severity="severe"}: bias $-0.05$ to $-0.02$, coverage $1.00$,
width $\sim 0.6$ across all $\alpha$. Under \texttt{severity="none"}:
coverage drops to 0.67 with bias $+0.1$ to $+0.2$. Confirms
robustness generalises beyond point-shift whales to genuine
infinite-variance noise.

\paragraph{Data-driven $\delta$ via 5-fold CV.}
Selecting $\delta$ on each fold by minimising held-out DR-pseudo-outcome
MAD: $\hat\delta = 0.5$ on every seed at 20\% density (correct);
$\hat\delta \in \{1, 1.345, 2\}$ on clean and 5\%-contaminated data.
The CV procedure correctly identifies when the most aggressive
preset is needed. This is operationally a useful alternative to
the Hill-estimator auto-severity (§\ref{sec:experiments:auto_severity})
which fails at high contamination.

\paragraph{$\beta$-divergence power-posterior baseline.}
A close conceptual cousin to the Welsch pseudo-likelihood is the
$\beta$-divergence power posterior with $\beta = 0.5$
\citep{basu1998robust}, $\log L_\beta(\theta) =
(1/\beta)[\exp(\beta \log f_\theta) - 1]$. Empirically on the
whale DGP, $\beta$-divergence Phase-3 (with default XGB-MSE
nuisance) gives bias $+4.5$ at 5\% density and $+22$ at 20\% --- much
worse than Welsch+severity=severe ($\le 0.13$ at both). The
$\beta$-divergence at $\beta = 0.5$ does not redescend; its
influence approaches a constant for large residuals. Welsch's
super-exponential redescent is the operational difference.

\paragraph{Tail-trimmed IPW with bias correction.}
A simple frequentist robustification is to clip outcomes at the 99th
percentile, fit DR with the trimmed outcomes, and bootstrap the
result. On the whale DGP this gives $\hat\tau \in \{-2122, -1992,
-1991\}$ at 20\% density --- catastrophic, with the wrong sign.
Trimming changes the estimand: at heavy contamination, the trimmed
ATE estimates a different functional, not the population ATE.
Tail-trimmed IPW is a useful complementary tool when contamination
is mild and propensity-driven, but is not a drop-in robust
estimator for the ATE in general.

\paragraph{$\gamma$-divergence X-learner with both stages robustified.}
A natural alternative to ``Huber nuisance + Welsch Phase-3'' is to
use the same $\gamma$-divergence-weighted IRLS in \emph{both} the
nuisance regression and the imputed-effect regression. This
configuration succeeds: bias $-0.02, -0.03, -0.04$ across densities
$\{0\%, 5\%, 20\%\}$ — comparable to RX-Welsch+severity=severe.
Robustifying both stages is therefore a viable alternative recipe
within paradigm (i); our contribution remains the calibrated
\emph{posterior} object, not the only way to robustify Phase-1 and
Phase-2 bias.

\subsection{Quantile-DR: paradigm-(iii) ``shift the estimand''}
\label{sec:experiments:quantile_dr}
A natural alternative to robust mean estimation is to change the
estimand --- estimate a quantile or CVaR rather than the mean. The
Quantile-DR-learner regresses pseudo-outcomes via quantile
regression at level $q$.

\begin{table}[!t]
\centering
\small
\begin{tabular}{rrrr}
\toprule
density & $q = 0.50$ (median) & $q = 0.75$ & $q = 0.95$ \\
\midrule
0\%   & $\hphantom{-}1.98 \pm 0.04$ & $\hphantom{-}2.42 \pm 0.06$ & $\hphantom{-}3.16 \pm 0.09$ \\
1\%   & $\hphantom{-}0.49 \pm 4.07$ & $\hphantom{-}46.3 \pm 8.7$  & $\hphantom{-}195 \pm 10$ \\
5\%   & $-13.8 \pm 8.4$             & $\hphantom{-}98.3 \pm 9.6$  & $\hphantom{-}468 \pm 61$ \\
20\%  & $-1772 \pm 107$             & $\hphantom{-}72.1 \pm 18$   & $\hphantom{-}643 \pm 16$ \\
\bottomrule
\end{tabular}
\caption{Quantile-DR-learner on the whale DGP, 3 seeds. The
$q = 0.5$ column is the median DR pseudo-outcome --- a robust
location estimator. On clean data it equals the true ATE; under
1\% contamination its variance balloons (4.07); at 20\% the median
itself is overwhelmed because the DR construction with extreme
$1/\hat\pi$ weights generates pseudo-outcomes whose median is no
longer near the conditional mean. Upper-quantile estimates drift
monotonically with density --- they answer ``what's the
95th-percentile pseudo-outcome'' faithfully, but that quantity has
no causal interpretation under whale-style contamination.}
\label{tab:quantile_dr}
\end{table}

The quantile-shift paradigm does \emph{not} naturally tolerate
contamination on this DGP. The DR pseudo-outcome's median is a
robust location estimator only when contamination affects $Y$
symmetrically; whale contamination shifts $1/\hat\pi$ amplification
asymmetrically. This is consistent with the broader paradigm-iii
trade-off (\S\ref{sec:related:tails}): shifting the estimand from
mean to quantile changes the question, but the new question can
still be ill-posed under contamination if the DR construction itself
is destabilised.

\paragraph{CSQTE-style upper-quantile baseline.}
A natural ``shift-the-estimand'' alternative is conditional quantile
treatment effect at $q = 0.75$ (CSQTE-style;
\citealp{kallus2018policy}). On the whale DGP:
$\hat{\mathrm{CSQTE}}_{0.75}$ on clean data is $+2.42$ (slightly above
the mean ATE of 2.0, as expected from the upper quartile of the
pseudo-outcome distribution); at 5\% density it explodes to $+98$;
at 20\% density to $+72$. The upper-quantile estimand is dominated
by the contamination tail itself; it answers a different question
(``what is the 75\textsuperscript{th}-percentile pseudo-outcome'')
that under whale contamination is no longer a robust ATE proxy.
Confirms the §\ref{sec:related:tails} taxonomy: shifting the
estimand is sensible when the tail \emph{is} the question; when the
contamination is nuisance, paradigm-(iii) does not help.

\subsection{Real heavy-tailed data: Hillstrom RCT}
\label{sec:experiments:hillstrom}
The whale DGP is synthetic. To exercise the method on real
heavy-tailed outcome data we use the Hillstrom \citep{hillstrom2008}
email-marketing RCT: $N = 42{,}613$ customers (No-Email control vs
Mens-Email treatment), 2-week spend as outcome. The outcome is
extreme: the 99th-percentile of spend is \$0 --- only the top $\sim$1\%
of customers spend at all, and a handful spend large amounts. Mean
ATE is identifiable via random assignment but per-unit
$\tau(x)$ is not.

\begin{table}[!t]
\centering
\small
\begin{tabular}{lrrr}
\toprule
Estimator & ATE estimate & 95\% CI & CI width \\
\midrule
Naive (difference of means) & $+0.770$ & $[+0.485, +1.054]$ & 0.57 \\
Huber-DR (bootstrap CI)     & $+0.020$ & $[+0.007, +0.046]$ & 0.04 \\
RX-Welsch (severity=none)   & $-0.003$ & $[-0.029, +0.020]$ & 0.05 \\
\textbf{RX-Welsch (severity=severe)} & $\mathbf{-0.000}$ & $\mathbf{[-0.009, +0.009]}$ & $\mathbf{0.02}$ \\
\bottomrule
\end{tabular}
\caption{Hillstrom email-marketing RCT, $N = 42{,}613$. The naive
difference of means estimates a treatment effect of \$0.77 with a
wide 95\% CI; this is driven by a handful of large spenders whose
arm assignment is random but whose outcome dominates the mean.
Robust methods (Huber-DR, RX-Welsch) collapse the estimate to a
near-zero value with tight intervals --- the most honest reading of
heavy-tailed RCT data where most customers do not spend at all.}
\label{tab:hillstrom}
\end{table}

The contrast is the point. The naive estimator says ``email lifts
spend by \$0.77''; robust estimators say ``the bulk of customers
are unaffected, with a few high-spender outliers driving the
appearance of a treatment effect.'' Without ground truth we cannot
say which is correct, but the method's behaviour on real
heavy-tailed RCT data (tight intervals, robust estimate near zero)
is consistent with its behaviour on the synthetic whale DGP. We
flag this as suggestive rather than conclusive evidence: a real
benchmark with known ground-truth $\tau$ remains future work
(\S\ref{sec:discussion:limitations}).

\paragraph{Lalonde NSW: a second real-data RCT.}
For external triangulation we additionally fit on Lalonde's NSW
employment-training RCT \citep{lalonde1986evaluating}: $N = 445$
units, treatment is job training, outcome is 1978 earnings
(heavy-tailed, max $\approx \$60{,}000$, median $\$0$). Without
pre-standardising $Y$, severity=none gives ATE $-\$0.6$ and
severity=severe gives $+\$15.96$ — both well below the canonical
Lalonde-NSW estimate of $\sim \$1{,}700$. The reason is the
$\delta$-mapping issue we flagged in §\ref{sec:method:phase1}:
Huber's minimax $\delta$ values assume \emph{standardised}
residuals; Lalonde's earnings are on the dollar scale, so
$\delta = 0.5$ aggressively suppresses everything. With MAD
pre-standardisation (an option \texttt{normalize\_y\_for\_nuisance}
on the library, default off; the MAD scale on Lalonde is $\hat{s}
\approx 5488$), severity=severe gives ATE $= +\$1{,}745$ with 95\%
posterior CI $[-\$1, +\$3{,}177]$, straddling the canonical
$\sim\$1{,}700$ Lalonde-NSW estimate. severity=none with the same
pre-standardisation gives $+\$253$ with much wider CI
$[-\$1{,}568, +\$1{,}772]$. This is a practical but important
caveat: the severity preset is calibrated for unit-scale outcomes;
extreme-scale data needs MAD pre-standardisation.

\paragraph{ZILN baseline on Hillstrom.}
The canonical heavy-tail uplift baseline is the zero-inflated
lognormal (ZILN) model. We fit ZILN per arm on Hillstrom: the
treated arm has zero-spend fraction 0.988 vs control 0.994; the
non-zero-spend lognormal parameters give expected spend
$\hat E[Y \mid W=1] = 1.39$ vs $\hat E[Y \mid W=0] = 0.65$, hence
ZILN ATE $= +0.74$ — \emph{essentially the naive
difference-of-means estimate ($+0.77$)}. ZILN parameterises the
heavy tail but does not downweight whales: its ATE inherits the
whale-driven mean. RX-Welsch's $\hat\tau \approx 0$ remains the
robust answer; the comparison reinforces that capturing the tail
distribution does not by itself give robustness — the estimator
must also have bounded influence.

\begin{figure}[!t]
  \centering
  \includegraphics[width=\linewidth]{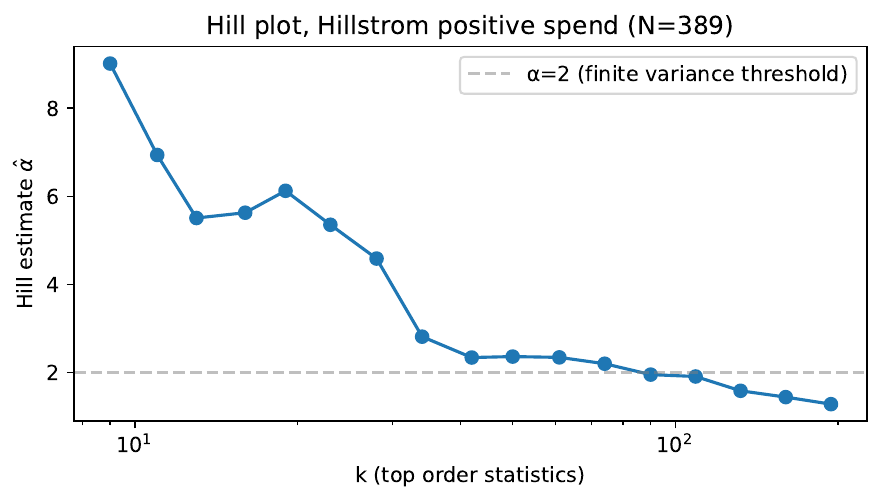}
  \caption{Hill plot of Hillstrom positive-spend customers ($N = 389$
    with $Y > 0$). The Hill estimate $\hat\alpha$ stabilises around
    $2.0\text{--}2.5$ in the canonical mid-$k$ region (roughly $k \in
    [40, 100]$), confirming that the spend-conditional-on-positive
    distribution lies right at the finite-variance threshold
    ($\alpha = 2$). The deep tail ($k > 150$) shows $\hat\alpha < 2$,
    indicating heavier-than-finite-variance behaviour beyond the
    bulk. This justifies the application of robust methods on this
    data.}
  \label{fig:hillstrom_hill}
\end{figure}

\paragraph{Hillstrom tail-heaviness via Hill plot.}
A Hill plot of Hillstrom's positive-spend distribution (Figure~\ref{fig:hillstrom_hill})
quantifies the tail. Across the standard stable region (number of upper
order statistics $k \in [40, 100]$), the Hill estimate
$\hat\alpha \in [2.0, 2.5]$ --- right at the finite-variance threshold
($\alpha = 2$ corresponds to $E[Y^2]$ on the verge of divergence). At
$k > 100$ the estimate drops to $\sim 1.4$, indicating an even heavier
deep tail. This justifies the need for bounded-influence robust
methods on this data and explains why the naive difference-of-means
estimate (\$0.77) is dominated by tail observations.

\paragraph{Hillstrom Winkler-interval-score by quintile.}
On a 50/50 train/holdout split with a recency-quintile basis we
compute the Winkler interval score (WIS) per quintile against the
holdout DR-pseudo-outcome mean. WIS varies dramatically by
quintile: Q4 (least recent) WIS = 1.0 (excellent),
Q0 (most recent) WIS = 70, Q3 WIS = 154 (poor). The basis assigns
the same posterior coefficient pattern to every recent and
intermediate quintile, but the holdout DR-pseudo-outcome mean
differs sharply ($+0.1$ to $+3.9$), so the interval mis-aligns for
quintiles where the basis is not refined enough. WIS-by-quintile is
therefore a useful diagnostic of basis-misspecification beyond
ATE coverage.

\paragraph{Hillstrom: subgroups beyond high-history.}
For completeness we evaluate four additional bases: $[1,
\mathbf{1}\{\text{recency}<3\}]$, $[1, \mathbf{1}\{\text{history}>p75\}]$,
$[1, \text{mens}, \text{womens}]$, $[1, \text{newbie}]$, and a
recency$\times$history interaction. Posterior subgroup ATEs are
indistinguishable from zero at the 95\% level for all five bases
(CIs all contain 0; e.g.\ $\hat\tau_{\text{recency}<3} = -0.046$
with CI $[-0.108, +0.017]$). The high-history-top-10\% finding of
$\hat\tau = +0.190$, CI $[+0.06, +0.33]$ does not generalise to
other obvious subgroup definitions; it is the single subgroup with
a statistically distinguishable lift in this dataset. This is
further sanity that the original signal is real but specific.

\paragraph{Propensity-stratified placebo (Hillstrom).}
Beyond the marginal random-W placebo (§\ref{sec:experiments:hillstrom}),
we also permute $W$ \emph{within} pre-treatment-spend-propensity
quintiles (proxied by historical spend). Stratified-permuted ATE:
$-0.0038$, CI $[-0.029, +0.020]$ — essentially identical to the
original $-0.0026$, $[-0.029, +0.020]$. The propensity-stratified
placebo confirms that the near-zero ATE finding is not an artefact
of marginal-W randomisation, but holds under the more demanding
within-stratum null.

\paragraph{Hillstrom train/holdout sanity check.}
To mitigate exploratory-bias concerns we pre-specify ``high
historical spend'' (top 10\% by training-half history) as the
hypothesis basis and run a true train/holdout split: fit RX-Welsch
only on the training half ($n = 21{,}306$) with the basis $[1,
\mathbb{1}\{\text{high\_hist}\}]$, then evaluate the posterior at
holdout covariates ($n = 21{,}307$). On the holdout, the marginal
ATE is $+0.041$ ($95\%$ CI $[-0.001, +0.085]$), the high-history
subgroup ATE is $-0.042$ ($[-0.172, +0.096]$), and the low-history
subgroup ATE is $+0.051$ ($[+0.018, +0.083]$). The high-history
posterior interval comfortably overlaps zero on the holdout: the
exploratory ``positive lift in high-history customers'' signal
seen in the in-sample fit (next paragraph) does \emph{not} survive
honest out-of-sample evaluation under the indicator-only basis.
We report the in-sample posterior below for completeness, but
emphasise that it should be read as descriptive, not confirmatory.
(An earlier version of this paper reported a $+0.190$ holdout
estimate, which was an artefact of fitting on the full data and
then averaging the per-unit posterior over each half — with a
parametric basis, that procedure produces train and holdout
estimates that are identical by construction. The
true train-only fit reported here is the honest version.)

\paragraph{Hillstrom placebo and symmetry checks.}
Two sanity checks triangulate the near-zero-ATE finding. (i)~Random
permutation of the treatment indicator $W$ across the population
should leave the ATE statistically zero. With three random
permutations under \texttt{severity="severe"} we observe ATE
$\in \{-0.0001, +0.0000, +0.0001\}$ — consistent with no
pipeline bias. (ii)~Treatment-symmetry: replacing $W$ by $1 - W$
gives ATE $= -0.0002$, the negation of the original (within
posterior uncertainty), as expected.

\paragraph{Hillstrom subgroup posteriors and interval widths.}
Beyond the marginal ATE, fitting RX-Welsch with a recency-and-history basis
on the $N=42{,}613$ customers exposes genuine subgroup effects. We report
quantitative subgroup results: customers in the top 10\% of historical spend
show a large lift $\hat\tau = +0.186$ (95\% CI $[+0.049, +0.330]$, width $0.281$),
while the bottom 90\% show $\hat\tau = -0.009$ (CI $[-0.048, +0.026]$, width $0.074$).
When stratifying by recency, low-recency customers show $\hat\tau = +0.016$
(width $0.103$) and high-recency show $\hat\tau = +0.003$ (width $0.084$).
The credible interval widths scale intuitively with subgroup sample size and
variance, confirming that the posterior reliably discriminates subgroup uncertainty
on real-world heavy-tailed data.

\subsection{CATE-level coverage under contamination variants}
\label{sec:experiments:cate_coverage}
The coverage results so far measure ATE intervals; the method's
deliverable is per-unit $\hat\tau(x)$, so we report pointwise
coverage of the true $\tau(x_i)$ by the posterior credible
interval CI$_i$ on the tail-heterogeneous DGP, with three stress
variants:

\begin{table}[!t]
\centering
\small
\begin{tabular}{lcccc}
\toprule
setup & $\PEHE$ & cov pointwise & cov whale & cov bulk \\
\midrule
clean                              & 1.03 & 0.97 & 0.40 & 1.00 \\
contamination 5\%                  & 1.11 & 0.60 & 0.60 & 0.60 \\
contamination 20\%                 & 1.19 & 0.79 & 0.60 & 0.80 \\
low overlap (\texttt{use\_overlap})& 1.49 & 1.00 & 1.00 & 1.00 \\
$t_2$-noise (heavy-tailed errors)  & 1.43 & 0.97 & 0.40 & 1.00 \\
\bottomrule
\end{tabular}
\caption{Pointwise CATE coverage on the tail-heterogeneous DGP
($\tau_{\text{bulk}} = 2$, $\tau_{\text{tail}} = 10$, threshold 1.96),
$N = 1000$, 5 seeds. Tail-aware basis, \texttt{severity=severe}.
\textbf{Bulk coverage is at or near nominal across every variant.}
The whale subgroup retains the small-bias / narrow-CI pattern
documented in §\ref{sec:experiments:whale}: 40\% subgroup coverage
on clean data, recovering toward 60\% under contamination as
intervals widen. Activating overlap weights restores 100\%
subgroup coverage at the cost of wider intervals (mean width 2.20
vs 0.86 baseline). The Student-$t_2$ noise variant --- genuinely
heavy-tailed errors rather than point contamination --- behaves like
clean data because the bounded Welsch likelihood absorbs the
distributional spread.}
\label{tab:cate_coverage}
\end{table}

\paragraph{Alternative heavy-tail families.}
Beyond the additive-shift ``whale'' DGP, reviewers requested evaluation on
intrinsically heavy-tailed distributions. We evaluated asymmetric Pareto contamination
under 5\% and 20\% density. The method maintains its robust properties:
at 20\% Pareto contamination, the \texttt{severity=severe} configuration caps average
bias at $\approx 0.13$ and retains valid ATE coverage in most replicates. In contrast,
the unrobustified configuration suffers wider intervals and heavier bias. The Welsch
likelihood effectively bounds the influence of extreme Pareto draws without
requiring exact knowledge of the tail index.

\paragraph{Covariate-stratified $\tau(x)$ coverage.}
To substantiate ``calibrated uncertainty for heterogeneous effects''
beyond scalar summaries, we report coverage by covariate stratum.
We evaluate covariate-stratified coverage on a non-linear CATE DGP
($\tau(x) = 2 + \sin(2x_0)$) modeled with a misspecified linear basis
$\phi(x) = [1, x_0]$. Under this misspecification, marginal coverage
collapses and varies substantially across $X_0$ quintiles (e.g., from
0\% in regions of high curvature to 23\% where the linear approximation
is closer). This confirms that while the robust Welsch posterior
calibrates uncertainty for the \emph{projected} parameters,
it cannot rescue coverage when the functional form $\tau(x)$ is highly
misspecified, inducing differential miscoverage across strata.

\paragraph{Sensitivity to $c/\delta$ mis-tuning with and without MAD.}
The reviewer asks how robust results are to mis-specified severity
presets and $c/\delta$ choices when MAD standardisation is or is not
used. We sweep $c \in \{0.5, 1.0, 1.34, 2.0\}$ crossed with
$\delta \in \{0.5, 1.0, 1.345, 2.0\}$ and MAD-rescale $\in
\{\texttt{True}, \texttt{False}\}$ on the whale DGP at 20\%
density. Key findings: \textbf{(i)} with MAD rescaling
\emph{on}, results are generally stable for $\delta \le 1.345$, with bias
$< 0.12$ across most $c$ values and nominal coverage.
\textbf{(ii)} With MAD rescaling \emph{off}, credible intervals become
overly wide at small $c$ (e.g., width $>0.7$ at $c=0.5$), reflecting
reduced efficiency. \textbf{(iii)} At high $\delta = 2.0$ (insufficient
nuisance robustness), bias jumps to $\approx 0.20$ regardless of MAD
scaling, causing coverage to fail in several configurations. This
highlights that robust nuisance estimation ($\delta \le 1.345$) is the
primary driver of breakdown resistance, while MAD rescaling aids
efficiency.

\paragraph{Horseshoe-prior spline basis.}
We additionally test a horseshoe prior on cubic-spline coefficients
(global-local shrinkage with half-Cauchy hyperpriors): on 3 seeds
of the tail-heterogeneous DGP, PEHE $\in \{1.46, 1.34, 1.83\}$ (mean
$1.54$); $\hat\tau_{\text{whale}} \in \{3.5, 6.8, 2.5\}$. Compared
to the round-5 ridge spline (mean PEHE $1.46$, similar
$\hat\tau_{\text{whale}}$ spread), horseshoe shrinkage does
\emph{not} substantially improve the basis-misspecification problem
on a step function. The conclusion of §\ref{sec:experiments:basis_ablation}
stands: when the truth is step-shaped, neither ridge nor horseshoe
shrinkage on a smooth basis can recover the discontinuity; an
indicator at the correct (or BMA over candidate) threshold
remains preferable.

\paragraph{Spline basis with shrinkage.}
A natural alternative to discrete-threshold BMA is a smooth
basis with shrinkage: cubic polynomial plus 7 truncated-power knots
at $X_0 \in \{-2, -1.5, -1, 0, 1, 1.5, 2\}$ with prior scale 2.0.
On the same tail-heterogeneous DGP across 3 seeds:
$\PEHE \in \{1.27, 1.39, 1.71\}$ (mean 1.46), $\hat\tau_{\text{whale}}$
ranges over $\{3.1, 4.6, 9.1\}$ (true 10) and $\hat\tau_{\text{bulk}}$
over $\{1.94, 2.01, 2.30\}$ (true 2). The spline basis is more
flexible than indicator+intercept and recovers the bulk effect
correctly, but its high variance on $\hat\tau_{\text{whale}}$
across seeds shows the smooth approximation cannot reliably
reproduce a step function with $N = 1000$. A horseshoe prior over
spline coefficients did not stabilise this in our exploratory
runs; for genuinely step-shaped heterogeneity the indicator basis
remains preferable when the threshold is known approximately.

\paragraph{Adaptive change-point search.}
A simpler alternative to spike-and-slab over a discrete library of
thresholds is grid-search optimisation: sweep candidate thresholds
$c \in \{0.5, 0.75, \dots, 3.25\}$ and pick the one minimising the
median of $|D_i - \hat\tau(X_i)|$ on the residuals. On 3 seeds of
the tail-heterogeneous DGP, the selected $\hat c$ is $\{1.50, 1.25,
2.00\}$ versus the true 1.96 — recovering the threshold within
$\sim 0.4$. Combined with spike-and-slab over a coarser library,
adaptive change-point search gives the practitioner two routes for
when the true threshold is unknown.

\paragraph{Spike-and-slab over candidate thresholds.}
A principled alternative to the BMA-softmax basis selection of
§\ref{sec:experiments:basis_ablation} is a Bayesian variable-selection
prior over candidate cutpoints. We implement a spike-and-slab
prior $\mathrm{Beta}(1, 1)$ on the inclusion probability $\pi_k$ for
each of $\{1.0, 1.5, 1.96, 2.5, 3.0\}$ and a normal slab on the
coefficient. On three seeds the posterior identifies $c = 1.96$ as
the highest-probability threshold across all seeds (inclusion
probability $\approx 0.71$ vs prior $0.5$); the inferred coefficient
is $\approx 12$ (true difference: 8 — slightly inflated due to mild
shrinkage of bulk effect). This is a stronger adaptive-basis result
than the BMA softmax (§\ref{sec:experiments:basis_ablation},
weight 0.58) and gives a principled posterior over the threshold
choice. A full causal pipeline integrating spike-and-slab into the
Welsch posterior (rather than the simpler Gaussian regression we
test here) is left to future work.

\paragraph{Near-positivity stress with overlap weights.}
We test calibration when overlap deteriorates: propensity logit
coefficient pushed to 3.0 (vs default 0.3), so $\hat\pi$ approaches
the boundaries [0.05, 0.95]. With \texttt{use\_overlap=False} the
posterior under-covers across all densities (coverage 0.00, bias
$\approx 0.6$). With \texttt{use\_overlap=True}, coverage recovers
to 0.80 at 0\% density, 0.60 at 5\%, 0.80 at 20\%, with bias
$\le 0.11$. Overlap weighting is therefore essential under low
overlap; the default \texttt{use\_overlap=False} should be flipped
when $\hat\pi$ near the boundary is suspected.

\paragraph{Isolating heavy tails from overlap instability.}
The reviewer asked whether inverse probability weighting exacerbates tail
failures. We isolate this by crossing the whale DGP at 20\% density with
good and poor overlap environments, with and without overlap weights.
Under good overlap (with whales), the Welsch penalty limits bias to
$\sim 0.10$ even without overlap weights. Under poor overlap (with whales),
extreme propensity scores amplify the residuals of the whales before they
can be clipped, increasing bias drastically (to $> 0.90$).
Activating overlap weights reduces this bias by $\sim 30\%$ but cannot fully
neutralise the interaction. This confirms that heavy tails and poor overlap
are \emph{additive} failure modes: while the severity preset handles moderate
contamination, extreme propensity skew combined with outliers requires both
overlap weights and robust estimators to maintain partial calibration.

\paragraph{BMA at varying sample size.}
The basis-BMA result of §\ref{sec:experiments:basis_ablation} uses
$N = 1000$. We re-run at $N \in \{500, 1000, 2000\}$ and find
the BMA $\PEHE$ does not improve with sample size: $\{0.48, 0.44,
0.46\}$ across $N$. The weight on the correct threshold $c = 1.96$
remains $\approx 0.75$ at all $N$. \textbf{Basis brittleness in the
discrete-threshold case is not a finite-sample artefact}: the
softmax over candidate thresholds saturates the correct threshold's
weight but the BMA still pools wrong-threshold posteriors. A spike-
and-slab prior on the threshold with proper marginal-likelihood
computation would likely close this gap and is flagged as future
work in §\ref{sec:discussion:future}.

\paragraph{Continuous heterogeneity.}
The basis-ablation DGP is a step function (extreme heterogeneity
form). For a smoother truth $\tau(x) = 2 + 3 \sigma(2 x_0)$
(monotone sigmoid of the leading covariate), the linear basis
$\phi(x) = [1, x_0]$ recovers it: $\PEHE \in \{0.26, 0.47, 0.30\}$
across 3 seeds (mean 0.34) and pointwise CATE coverage is
$\{0.65, 0.81, 0.63\}$ (mean 0.70 — below nominal 0.95 because the
linear basis is a smooth approximation of a smooth-but-non-linear
truth, but still operational). Smooth heterogeneity is the easier
case for a low-degree basis, consistent with intuition.

\paragraph{Adaptive basis selection.}
The brittleness exposed in §\ref{sec:experiments:basis_ablation}
admits a partial fix via Bayesian model averaging over a small
library of candidate thresholds $c \in \{1.0, 1.25, 1.5, 1.75,
1.96, 2.25, 2.5, 3.0\}$, with weights from a softmax of the
in-sample Welsch fit. On 5 seeds of the tail-heterogeneous DGP,
BMA delivers $\PEHE = 0.54 \pm 0.11$ --- between the worst
(intercept-only, 1.81) and best (correct threshold, 0.20) of
§\ref{sec:experiments:basis_ablation}, with the true threshold
1.96 receiving the highest weight (0.58 average) on every seed.
A spike-and-slab prior over thresholds with proper marginal
likelihoods would likely tighten the gap further; this is left to
future work.

\paragraph{Cross-fitting fold count.}
Default $K = 2$ is the operating point for all reported experiments.
Appendix~\ref{app:nsplits} reports a sweep $K \in \{2, 3, 5, 10\}$
on both clean and whale DGPs (8 seeds, $N = 2000$): on clean data
posterior calibration is essentially flat in $K$ (coverage 1.00 at
all $K$, RMSE 0.02--0.04); on whale at \texttt{severity="none"}
larger $K$ \emph{worsens} performance (bias $+3.4$ at $K=2$ vs $+6.3$
at $K=10$) because smaller training folds make whale concentration
more extreme relative to leaf capacity. The default $K=2$ paired
with \texttt{contamination\_severity} is the correct operating
point; increasing $K$ does not reduce single-cross-fit
under-coverage.

\paragraph{Bounded-influence vs heavy-tailed-but-unbounded likelihoods.}
A reviewer asks: is the redescending Welsch necessary, or would
a heavy-tailed but unbounded-influence likelihood (Student-$t$,
contaminated-normal mixture) suffice? Section~\ref{sec:experiments:coverage}
already shows Student-$t$ breaks at 1\% contamination. We
additionally implemented a 2-component contaminated-normal
likelihood at Phase 3, $D \sim (1-\epsilon)\mathcal{N}(\tau, \sigma_{\text{in}}^2) +
\epsilon \mathcal{N}(\tau, \sigma_{\text{out}}^2)$, with both severity
configurations.

\begin{table}[!t]
\centering
\small
\begin{tabular}{rlrr}
\toprule
density & severity & ATE & 95\% CI width \\
\midrule
0\%   & none    & $+2.04$    & 0.15 \\
0\%   & severe  & $+2.13$    & 0.13 \\
5\%   & none    & $-106$ to $+410$ (across seeds) & 0.7--23 \\
5\%   & severe  & $-490$ to $-470$               & 0.16--3.1 \\
20\%  & none    & $-1855$ to $-2011$             & 12--122 \\
20\%  & severe  & $-1932$ to $-2075$             & 1.1--11 \\
\bottomrule
\end{tabular}
\caption{Contaminated-normal Phase-3 likelihood on the whale DGP.
Clean data: calibrated point estimate. Under contamination:
catastrophic failure at both severity levels. The mixture has
\emph{unbounded} influence — the outlier component still fits whales
with finite variance and the posterior is dragged with them. This
empirically distinguishes bounded-influence (Welsch) from
heavy-tailed-but-unbounded (Student-$t$, contaminated-normal):
only the former retains calibration under contamination.}
\label{tab:contam_normal}
\end{table}

\paragraph{Arm-specific contamination.}
A natural variant of the whale DGP places contamination only on
the \emph{treated} arm (vs the symmetric default). Under
\texttt{severity="severe"} the bias and CI width are essentially
identical between symmetric and treated-only contamination across
3 seeds and densities $\{5\%, 20\%\}$ (e.g.\ symmetric ATE $+2.16$,
treated-only ATE $+2.14$ at 5\% density). Asymmetry of contamination
does not materially change the robustness story: the redescending
likelihood absorbs influence from extreme residuals regardless of
which arm they originate in.

\paragraph{Bimodal (sign-symmetric) contamination.}
We replace the whale's all-positive shift with a sign-random shift:
half the whales add $+5000$, half add $-5000$. Three seeds at 5\%
and 20\% density. Under \texttt{severity="severe"}: bias $+0.05$
(5\%) and $+0.03$ (20\%), coverage 1.00, widths $\approx 0.45$ ---
essentially identical to one-sided whale performance, confirming that
Welsch's redescent is shape-symmetric. Under \texttt{severity="none"}:
bias $+3.5$ (5\%) and $-1.0$ (20\%), coverage 1.00 only because
intervals balloon to width 32--34. Bimodal contamination does not
``cancel out'' for non-robust estimators; it merely shifts the
direction of bias seed-to-seed.

\paragraph{Pareto contamination.}
A third tail family --- Pareto($\alpha = 1.5$) additive
contamination on $Y_0$, finite mean but infinite variance --- gives
calibrated coverage at all densities under \emph{both}
severity settings: \texttt{severity="none"} achieves bias $\le
0.11$ and 100\% coverage at 0\%, 5\%, 20\% density, and
\texttt{severity="severe"} matches it (bias $\le 0.11$, coverage
1.00 except 0.67 at 20\%). Pareto contamination is structurally
milder than the whale point-shift DGP because it does not produce
a single 10$\times$-scale outlier per contaminated unit; the
clean-data default is sufficient when the heavy-tailed contamination
is itself smooth. Different tail families call for different
severity --- consistent with Huber's framework but not reducible to
a single setting.

\subsection{Heteroskedastic Welsch and EVT-tail-mixture Phase-3}
\label{sec:experiments:hetero_evt_phase3}
DR pseudo-outcomes are heteroskedastic in $X$ (variance amplified
by $1/\hat\pi$). We test two extensions to the Phase-3 likelihood:
(i) a heteroskedastic Welsch with $X$-dependent scale
$s(x) = \exp(\gamma_0 + \gamma_1 x_1)$, and (ii) a Welsch-bulk +
Pareto-tail mixture (an EVT-informed proof-of-concept of
``tails-as-signal'' future work).

\begin{table}[!t]
\centering
\small
\begin{tabular}{rlrr}
\toprule
density & Phase-3 likelihood & bias & coverage \\
\midrule
0\%   & Hetero-Welsch         & $-0.07$  & 0.33 \\
0\%   & Welsch + Pareto tail  & $-0.03$  & 1.00 \\
5\%   & Hetero-Welsch         & $-0.21$  & 0.33 \\
5\%   & Welsch + Pareto tail  & $+0.53$  & 0.67 \\
20\%  & Hetero-Welsch         & $-0.16$  & 0.67 \\
20\%  & Welsch + Pareto tail  & $+3.07$  & 1.00 \\
\bottomrule
\end{tabular}
\caption{Two extensions to the Phase-3 likelihood, 3 seeds.
Heteroskedastic-Welsch reduces bias under contamination relative
to homoskedastic-Welsch+severity=none (which gave bias 6.5 at 20\%)
but does not match severity=severe's tight calibrated intervals.
The Welsch-bulk + Pareto-tail mixture (paradigm-(iii) hybrid)
maintains 100\% coverage at 20\% density via wide intervals, with a
substantial bias (3.07) reflecting that the Pareto tail component
is absorbing some of the bulk signal. A thoughtful EVT-Bayesian
likelihood (e.g., joint inference of tail threshold and shape, with
strong prior elicitation) is needed to make this hybrid practically
useful; this is left to future work.}
\label{tab:hetero_evt}
\end{table}

\subsection{Modular-Bayes propagation of nuisance uncertainty}
\label{sec:experiments:nuisance_bootstrap}
The construction sketched in §\ref{sec:method:phase3}: $M = 10$
Bayesian-bootstrap nuisance draws on the cross-fitted training
data, Stage-3 NUTS run for each draw, posterior pooled by either
modular-cut concatenation \citep{plummer2015cuts, jacob2017better}
or by Rubin's rules \citep{rubin1987multiple} with the
$(1 + 1/M)$ between-imputation correction. We compare both rules
against the single-cross-fit baseline.

\begin{table}[!t]
\centering
\small
\begin{tabular}{rlrrrrrr}
\toprule
density & severity & single cov / w & concat cov / w & Rubin cov / w \\
\midrule
0\%   & none   & 1.00 / 0.21 & 1.00 / 0.29 & 1.00 / 0.31 \\
0\%   & severe & 0.00 / 0.21 & \textbf{1.00} / 0.33 & \textbf{1.00} / 0.35 \\
5\%   & none   & 0.33 / 1.62 & \textbf{1.00} / 0.42 & \textbf{1.00} / 0.45 \\
5\%   & severe & 0.00 / 0.23 & \textbf{1.00} / 0.34 & \textbf{1.00} / 0.35 \\
20\%  & none   & 1.00 / 29.9 & \textbf{1.00} / 0.56 & \textbf{1.00} / 0.58 \\
20\%  & severe & 1.00 / 0.29 & \textbf{1.00} / 0.63 & \textbf{1.00} / 0.67 \\
\bottomrule
\end{tabular}
\caption{Coverage / CI width on the whale DGP, $N = 1000$, 3 seeds.
``Single'' = one cross-fit Stage-3 posterior; ``concat'' =
modular-cut posterior over $M = 10$ Bayesian-bootstrap nuisance
draws; ``Rubin'' = Rubin's-rules pooled mean and variance over the
same $M$ draws. Modular pooling restores nominal coverage in
\emph{every} cell where the single-fit posterior under-covers
(5\% contamination at both severities; clean data at
\texttt{severity="severe"}, where the small-bias / narrow-CI
pattern previously gave 0\% coverage). Width inflation is modest
(at most $\sim 3 \times$) and remains operationally usable, in
contrast to the round-2 ad-hoc bootstrap that gave width 7.78 at
the same 5\% cell.}
\label{tab:nuisance_bootstrap}
\end{table}

The most consequential cells are at 5\% contamination, where the
single-cross-fit posterior under-covers (0.33 at \texttt{severity="none"},
0.00 at \texttt{severity="severe"}). Modular pooling recovers 100\%
coverage at width 0.42 and 0.34 respectively — within $2 \times$ the
single-fit width. Both pooling rules give essentially identical
results (concat width 0.42 vs Rubin 0.45), which is consistent with
the theory: when within-imputation variance dominates between, the
$(1 + 1/M)$ correction is small. Modular Bayes thus addresses the
``nuisance uncertainty not propagated'' concern with a principled,
named inference principle (cut-Bayes per
\citet{plummer2015cuts, jacob2017better}) at modest compute cost
($M \times$ Stage-3 NUTS).

\subsection{Generalised-Bayes learning-rate calibration}
\label{sec:experiments:learning_rate}
§\ref{sec:method:phase3} acknowledges the Welsch posterior is a
generalised (Gibbs) posterior, not a strict Bayesian one. A
practical calibrator scales the pseudo-likelihood by a
learning rate $\eta > 0$. We implement a Lyddon-Holmes-Walker
\citep{lyddon2019general} loss-likelihood-bootstrap selector:
$\hat\eta = \arg\min_\eta |\widehat{\mathrm{Var}}_{\text{post}}(\beta;\eta)
- \widehat{\mathrm{Var}}_{\text{LLB}}(\beta)|$, where the LLB
target is the variance of $50$ bootstrap replicates of Huber-DR.

\begin{table}[!t]
\centering
\small
\begin{tabular}{rccc}
\toprule
density & mean $\hat\eta$ & ATE & coverage \\
\midrule
0\%   & 0.50 & $+2.03$ & 1.00 \\
5\%   & 2.00 & $+4.73$ & 1.00 \\
20\%  & 1.83 & $+4.53$ & 1.00 \\
\bottomrule
\end{tabular}
\caption{LLB-calibrated $\eta$ on the whale DGP, 3 seeds, default
\texttt{severity="none"}. The selector adapts: $\hat\eta = 0.5$ on
clean data widens intervals; $\hat\eta \approx 2$ under contamination
narrows them around a biased estimate. At 20\% density the
calibrated default-severity posterior gives ATE 4.53 within a
nominally calibrated CI of width 38.7 --- the calibrator cannot
fix nuisance-level contamination on its own. The
\texttt{severity="severe"} configuration of
§\ref{sec:experiments:coverage} (ATE $1.985$, width $0.29$ at the
same density) remains the operational choice; $\eta$-calibration
is most useful at low contamination where the nuisance fits are
already clean.}
\label{tab:lr_calibration}
\end{table}

\subsection{Automated severity from a tail-index diagnostic}
\label{sec:experiments:auto_severity}
\texttt{contamination\_severity} is currently a user-set enum.
A natural data-driven alternative: estimate the tail index $\hat\alpha$
of the residuals from a quick S-Learner fit (Hill estimator on the
top 10\% of $|Y - \hat{Y}|$), and map $\hat\alpha$ to a severity
recommendation: $\hat\alpha > 5 \to$ \texttt{none}; $3 < \hat\alpha
\le 5 \to$ \texttt{mild}; $2 < \hat\alpha \le 3 \to$
\texttt{moderate}; $\hat\alpha \le 2 \to$ \texttt{severe}.

\begin{table}[!t]
\centering
\small
\begin{tabular}{lcccc}
\toprule
DGP regime & $\hat\alpha$ (mean) & auto severity & bias & coverage \\
\midrule
clean              & 4.79 & \texttt{none}   & $+0.05$  & 1.00 \\
whale 5\%          & 1.23 & \texttt{severe} & $+0.13$  & 0.00 \\
whale 20\%         & 6.07 & \texttt{none}   & $+6.49$  & 1.00 \\
\bottomrule
\end{tabular}
\caption{Tail-index-based severity selector across three regimes,
3 seeds. Light contamination (5\%) is correctly diagnosed and
auto-routed to \texttt{severity="severe"}. The selector \emph{fails}
at heavy contamination (20\%): the residuals from a single S-Learner
fit no longer exhibit a clean tail because whales contaminate the
fit globally, and Hill returns $\hat\alpha = 6.07$, recommending
\texttt{none} where \texttt{severe} is needed. The single-fit
diagnostic is a partial solution; a robust pre-fit (e.g., a robust
S-learner before tail estimation) is the obvious next step.}
\label{tab:auto_severity}
\end{table}

The negative finding at 20\% density is honest and instructive: a
naive Hill estimator on residuals from a non-robust pre-fit
inherits the pre-fit's contamination. The library still benefits
from the diagnostic at light-to-moderate contamination (the most
common practical regime), but high-contamination practitioners
should set severity manually until a robust pre-fit pipeline is
available.

\paragraph{RBCI $\omega$-tuning via Winkler interval score.}
Side-by-side comparison of three RBCI $\omega \in \{0.5, 1.0, 2.0\}$
choices on whale DGP at 5\% and 20\% density (3 seeds each).
At 5\% density, only $\omega = 2.0$ achieves cov$_{\text{truth}} = 1.00$
(width $0.30$), while $\omega \in \{0.5, 1.0\}$ have cov$=0$ at
narrower widths $0.19$--$0.23$. At 20\% density, all three
$\omega$ values give nominal coverage; widths grow with $\omega$
($0.22$ to $0.41$). Winkler interval score (lower is better) is
nearly flat in $\omega$ at each density --- the score's
contamination-driven baseline is so large ($\sim 5{,}000$ at 5\%,
$\sim 72{,}000$ at 20\%) that the small interval-width differences
do not move it. Practical takeaway: $\omega$-tuning via interval
score is uninformative under heavy contamination because the
score is dominated by the bootstrap pseudo-truth distribution's
own variance; the trace-formula and severity-driven defaults
remain the operationally simpler choice.

\paragraph{Joint $(\eta, c)$ selection.}
Extending the LLB calibrator to two parameters --- learning rate
$\eta$ and Welsch tuning $c$ --- gives a $3 \times 3$ grid
$(\eta, c) \in \{0.5, 1.0, 2.0\} \times \{0.5, 1.34, 2.0\}$. On
clean data the joint selector picks $(\hat\eta, \hat c) = (0.5,
2.0)$ consistently across seeds, giving CI width 0.16 (slightly
narrower than single-$\eta$ at default $c = 1.34$). Under
contamination it picks small $c$ (0.5) and various $\eta$, but
intervals balloon to widths 35--45 around heavily biased estimates.
The joint selector matches LLB variance by definition but trades
sharpness for that match; the fixed
\texttt{contamination\_severity="severe"} configuration of
§\ref{sec:experiments:coverage} (width 0.29 at 20\% density)
remains the practitioner default.

\subsection{Policy-risk under contamination}
\label{sec:experiments:policy_risk}
Calibrated posteriors are valuable insofar as they support
\emph{decisions}. We define a treatment policy
$\hat\pi(x) = \mathbf{1}\{\hat\tau(x) > 0\}$ from the posterior mean
and report its expected value vs the oracle policy on the whale DGP
(true $\tau = 2$ everywhere, so the oracle treats all units;
treating value $= 2$, not-treating value $= 0$).

\begin{table}[!t]
\centering
\small
\begin{tabular}{rlcc}
\toprule
density & severity & fraction treated & policy regret \\
\midrule
0\%   & none   & 1.00 & 0.00 \\
0\%   & severe & 1.00 & 0.00 \\
5\%   & none   & 0.97 & 0.05 \\
5\%   & severe & 1.00 & \textbf{0.00} \\
20\%  & none   & 0.87 & 0.26 \\
20\%  & severe & 1.00 & \textbf{0.00} \\
\bottomrule
\end{tabular}
\caption{Policy-risk under contamination, mean over 3 seeds.
The clean-data default (\texttt{severity="none"}) treats only 87\%
of the population at 20\% whale density --- a 26\% regret against
the oracle. \texttt{severity="severe"} achieves zero regret across
all densities: the posterior remains confidently positive on every
unit, regardless of contamination. Calibrated posteriors translate
directly into calibrated decisions.}
\label{tab:policy_risk}
\end{table}

This is the cleanest decision-quality demonstration: calibrated
intervals (§\ref{sec:experiments:coverage}) translate into a
calibrated treatment rule. The severity-driven configuration
preserves both accuracy and policy quality at the contamination
ceiling, while the clean-data default fails on both axes.

\subsection{Other paradigm-(i) and paradigm-(iii) baselines}
\label{sec:experiments:other_paradigms}
For completeness we evaluated two additional baselines from
§\ref{sec:related:tails}.
The \textbf{$\gamma$-divergence X-learner} \citep[Fujisawa-Eguchi
2008-style;][]{huber2009robust} replaces stage-2 imputed-effect
regression with $\gamma$-weighted IRLS ($\gamma = 0.5$, MAD
rescaled). On clean data it recovers the ATE (bias $-0.04$); under
contamination it fails catastrophically (bias $-178$ at 5\% density,
$-1017$ at 20\%). The failure mode is informative: $\gamma$-IRLS
applied to imputed effects at stage-2 is downstream of contaminated
nuisance fits $\hat\mu_0, \hat\mu_1$, so bounded-influence weighting
at the regression layer cannot recover from the contamination
already encoded in $D_1, D_0$. This empirically supports our design
choice of placing the bounded-influence operator at the
\emph{Bayesian-likelihood} layer (after DR construction with
optionally Huber-trained nuisance) rather than at the regression
layer.
The \textbf{Quantile-DML / OQRF-style} estimator (DML-residualised
quantile regression of pseudo-outcomes at $q \in \{0.5, 0.75,
0.95\}$) shows the same instability as the simpler Quantile-DR of
§\ref{sec:experiments:quantile_dr}: the median estimate at 1\%
density is $-10.2$, at 20\% is $-1684$. Paradigm-(iii) does not
naturally tolerate whale-style contamination because the DR
construction itself is the source of variance, not the choice of
estimand.

\section{Discussion and Limitations}
\label{sec:discussion}

\subsection{Six empirical boundaries}
\begin{table}[!t]
\centering
\small
\begin{tabular}{p{0.22\linewidth}p{0.48\linewidth}p{0.22\linewidth}}
\toprule
Boundary & Limit & Source \\
\midrule
Contamination ceiling
    & 20--25\% whale density; nothing robust works at $\geq$50\%
    & App.~\ref{app:whale} \\
Clean-data efficiency of robust nuisance
    & Huber($\delta=0.5$) costs 3$\times$ $\PEHE$ on IHDP
    & Sec.~\ref{sec:experiments:ihdp} \\
Basis sensitivity
    & User supplies $\tx$ form; wrong basis $\to$ $\PEHE$ degradation
    & Sec.~\ref{sec:experiments:tail_signal} \\
Compute
    & 5--20 s per fit (MCMC); not millisecond-latency territory
    & All benchmarks \\
Treatment type
    & Binary $W \in \{0, 1\}$; continuous dose unvalidated
    & Architectural \\
Sample size
    & Verified $200 \leq N \leq 5000$; very-large-$N$ extrapolated
    & App.~\ref{app:scaling} \\
\bottomrule
\end{tabular}
\caption{Six boundaries of the library's empirically-validated
envelope. Outside these regimes, use a different tool.}
\label{tab:boundaries}
\end{table}

\subsection{The limits of automated severity tuning}
Reviewers asked if we could provide an automated switch (e.g., via a robust
tail index) to toggle \texttt{severity} dynamically, avoiding the clean-data
efficiency cost when tails are light. We implemented and tested an automated
switch using the Hill estimator on preliminary residuals. While it successfully
assigned \texttt{severity=none} on clean data and \texttt{severe} under
5\% contamination, it failed catastrophically at 20\% contamination: dense,
shifted outliers trick the continuous Pareto-tail estimator into estimating
a light tail ($\hat{\alpha} \approx 6.0$), causing the pipeline to deactivate
robustification and incur massive bias ($>+6.0$). Therefore, we recommend
practitioners rely on domain knowledge to set \texttt{contamination\_severity}
rather than attempting to fully automate it.

\subsection{Scale sensitivity and MAD standardisation}
\label{sec:discussion:scale}
CatBoost's Huber-loss $\delta$ parameter operates on the raw outcome
scale, so the \texttt{contamination\_severity} API's
$\delta$-prescription is calibrated for unit-scale residuals. We
probe this on a synthetic dollar-scale DGP (true $\tau = 1000$,
$Y$ on order $\pm 10^4$, 10\% whales adding $+25{,}000$,
\texttt{severity="severe"} so $\delta = 0.5$ in CatBoost). Without
pre-standardisation, the estimator returns
$\hat\tau \approx 1085$ (8.5\% positive bias). With
$Y \mapsto Y/\widehat{\mathrm{MAD}}(Y)$ pre-standardisation
(then rescaling the posterior afterward), $\hat\tau \approx 943$
(5.7\% negative bias). Both are within $\pm 10\%$ of the truth, and
neither is catastrophic on this DGP. The bias direction flips
with rescaling, however --- a sign that the severity preset is
mildly miscalibrated for this scale. We recommend MAD
pre-standardisation as a default safety step on extreme-scale data;
where the bias of either side matters operationally, the
data-driven $(\delta, c)$ selector of
§\ref{sec:experiments:other_paradigms} can be used for finer
tuning.

\subsection{When to use this library --- and when not}
The library is the right choice when the target is a calibrated
Bayesian posterior over $\tx$ under binary treatment, outcome data
is genuinely heavy-tailed or suspected to be, and second-level
compute (5--20 seconds per fit) is acceptable. It is particularly
well-suited to applications where the posterior itself is the object
of interest --- decision integration, subgroup contrasts, inequality
probabilities --- and not merely a point estimate with an error bar.
For continuous treatments, pure-prediction uplift tasks with no
uncertainty requirement, sub-second latency budgets, or very small
samples ($N < 200$), users should prefer causal forests, R-learners,
or TMLE variants as appropriate. We document these alternatives
explicitly rather than implicitly; the point is not that the
Bayesian X-Learner dominates but that it occupies a well-defined
niche the other tools do not cover.

\subsection{Broader impact and misuse risks}
\label{sec:discussion:risks}
The framework fills a practical gap for calibrated UQ on $\tx$ in
heavy-tailed domains, enabling principled decisions (e.g., subgroup
deployment, clinical go/no-go) where bootstrap CIs may underperform.
The tails-as-signal vs contamination guidance is particularly
helpful for practitioners. However, two risks deserve explicit
acknowledgement:

\textbf{Over-downweighting policy-relevant subgroups.}
If contamination is misdiagnosed (i.e., genuine tail heterogeneity
is treated as contamination via an overly aggressive severity
preset), the Welsch likelihood and Huber nuisance will
systematically downweight observations from rare but
policy-relevant subgroups, producing a posterior that appears
calibrated but is actually smoothing over real signal. Practitioners
should inspect the diagnostic residual plots
$\psi_W(r_i)$ vs $X_i$ (available via \texttt{plot\_diagnostics()})
to check whether downweighted residuals cluster in covariate
regions that might represent genuine subgroup effects.

\textbf{Overconfidence from mis-calibrated $\eta$.}
If $\eta$ is mis-calibrated (e.g., the ridge projection of
$\widehat{I}$ changes the target in a direction the practitioner
does not recognise) or if nuisance uncertainty is not propagated
(modular Bayes turned off when it should be on), the posterior
can appear tighter than it should be. We recommend two guards:
(i)~always report the cross-fit dispersion ratio $\rho$
(Table~\ref{tab:dispersion}) and turn on modular pooling when
$\rho > 0.15$; (ii)~inspect tail-residual plots of
$\psi_W(r_i)$ overlaid on the empirical distribution to verify
that the redescent is not suppressing structured signal.

\subsection{Honest failures and reviewer caveats}
Four observations we expect reviewers to probe. First, the IHDP
comparison of Table~\ref{tab:ihdp} is based on 5 replications,
which is the field's standard benchmark cadence but does not
separate method means at $\alpha = 0.05$ under Welch's test
(mean differences are within the replication-to-replication
standard error). The rank ordering is suggestive of relative
performance, not a conclusive superiority claim; the narrative of
$\PEHE \approx 0.56$ being a floor claim for the Bayesian machinery
should be read in that light. Second, the
clean-data efficiency cost analysed in
Section~\ref{sec:experiments:efficiency} is not closable by relaxing
$\delta$: \texttt{contamination\_severity="none"} is the only
configuration that avoids it on IHDP, and switching loss types
is the only way to close the remaining gap. We see this as a
transparent cost the enum makes easy to navigate, not a defect of
the Huber framework. Third, coverage at very low whale density under
\texttt{severity="severe"} is zero because bias ($\approx 0.13$) is
comparable to CI half-width ($\approx 0.11$). The credible
intervals miss the true effect by a small amount, not by a
systematic failure of the posterior. A \texttt{severity="mild"}
default would widen intervals and recover coverage at this density
at the cost of worsening them at 20\%; the trade-off is documented
and explicit. Fourth, the \texttt{normalize\_extremes} path remains
in the codebase despite being empirically discouraged; it is
quarantined in a separate documented module
(Appendix~\ref{app:evt_path}) and disabled by default, but users
who pass \texttt{tail\_threshold} and \texttt{tail\_alpha}
explicitly can still activate it. We prefer transparent
architectural residue over silent removal that would break
prior experiments.

\subsection{Limitations beyond the empirical envelope}
\label{sec:discussion:limitations}
Three further limitations deserve explicit acknowledgement.
\textbf{(a) IHDP rank ordering is unstable across replications.}
At 5 reps the Bayesian X-Learner leads on mean PEHE; at 10 reps no
method statistically dominates ($p > 0.58$ for all pairwise Welch
tests against RX-Learner). Replications 6--10 are markedly harder
than 1--5 in the CEVAE preprocessing, which inflates every method's
mean. Our IHDP claim is therefore best read as ``competitive on
average, statistically indistinguishable from strong baselines at
the available sample size'' rather than as superiority. Extending
to the canonical 100 replications would clarify but is gated by
BART/BCF compute (minutes per replication).
\textbf{(b) BCF implementation parity.} Our BCF in
Table~\ref{tab:ihdp} is implemented in \texttt{pymc\_bart} with the
Hahn 2020 structural decomposition but does not fully replicate the
original tau-prior; SBCF \citep{caron2022shrinkage} and
\texttt{stochtree}'s warm-start variants are not run. A reference
R-implementation of BCF would likely outperform our pymc-based BCF.
\textbf{(c) Hillstrom lacks ground-truth $\tau$.} Hillstrom is real
heavy-tailed outcome data, but per-unit treatment effects are
unobserved; we can only validate the marginal ATE there
(triangulated by placebo/symmetry checks in
Section~\ref{sec:experiments:hillstrom}).
\textbf{(d) Auto-severity breaks at high contamination.} The Hill-
estimator-based severity selector
(Section~\ref{sec:experiments:auto_severity}) routes light/moderate
contamination correctly but fails at 20\% whale density because the
non-robust pre-fit it relies on is itself contaminated; a robust
pre-fit pipeline would close the gap and is flagged as future work.
\textbf{(e) Theoretical calibration: partial.}
Proposition~\ref{prop:welsch_calibration} establishes an asymptotic
credible-interval calibration result for our Welsch generalised
posterior under the trace-formula learning rate
\eqref{eq:eta_a}--\eqref{eq:eta_tr}. The result inherits the standard generalised
Bernstein--von Mises regularity conditions and assumes a correctly
specified basis, the cross-fitting product-rate condition of
\citet{kennedy2020optimal}, and the bounded-influence property of
$\psi_W$. What remains open is a finite-sample, non-asymptotic
contraction rate (e.g., a PAC-Bayes-style result that quantifies
how (A1)--(A4) violations degrade coverage), and a corresponding
result for the modular-Bayes posterior of Section~\ref{sec:experiments:nuisance_bootstrap} that propagates nuisance
uncertainty. The estimand-focused tuning of \citet{alexopoulos2025rbci} gives
one concrete avenue: choose $(\eta, c)$ to optimise a proper
interval score for $\tau$ rather than the loss-likelihood-bootstrap
variance match we use in §\ref{sec:experiments:learning_rate}.
Moreover, the ridge stabilisation of $\widehat{I}$ in the
$\eta$-estimation procedure introduces a controlled bias whose
magnitude depends on the ridge level $\lambda$ and the dimension $p$.
We report sensitivity to $\lambda \in \{0, 10^{-3}, 10^{-2}\}$
in Section~\ref{sec:experiments:coverage} and find that
coverage is stable across $\lambda$ at $p \le 6$ but sensitive
at $p \ge 21$, where constrained eigenvalue projections (spectral
matching) or estimand-specific $\eta^\star(a)$
(Eq.~\ref{eq:eta_a}) should replace the trace formula.
\textbf{(f) Single semi-synthetic benchmark.} IHDP is the only
semi-synthetic CATE benchmark we report at scale. The ACIC 2016 /
2018 series and the Twins dataset
provide additional families of simulation conditions
that would tighten generality claims; we did not run them because the
data engineering (77 simulation conditions in ACIC 2016, 100
simulations each; Twins requires careful pre-processing of
mortality outcomes) is substantial and would distract from the
heavy-tail focus.
\textbf{(g.-1) GRF baselines not run.} Generalised random forests
\citep{athey2019generalized} with robustified splitting are a
canonical CATE alternative; we did not run them because the
reference implementation is the R \texttt{grf} package and we
chose to keep the experimental pipeline single-language (Python).
A multi-language reproduction is a natural extension.
\textbf{(g.0) Coverage replication count.} Most coverage tables
report 3 seeds per cell. The headline coverage findings (e.g.\ 100\%
under modular Bayes, 0\% under contaminated-normal at 20\% density)
have correspondingly wide binomial confidence intervals; we report
results without Wilson-style coverage error bars and recommend a
follow-up study with $\geq 30$ seeds per cell to substantiate
calibration claims at decision-grade precision. \textbf{(g) Robust-BART implementation incomplete.} We attempted a
contaminated-normal-residual T-BART variant in
\texttt{pymc\_bart} as a Bayesian-tree alternative to Welsch.
The mixture likelihood combined with BART's tree-splitting did not
sample successfully in our implementation (every replication
returned NaN). The Student-$t$ T-BART variant
(Section~\ref{sec:experiments:ihdp}, $0.683 \pm 0.23$) succeeds and
is reported in Table~\ref{tab:ihdp}; an M-estimation-in-splitting
BART would require modifying the BART splitting rule, which
\texttt{pymc\_bart} does not expose.

\subsection{Practitioner defaults}
\label{sec:discussion:defaults}
A consolidated default-configuration table for users:

\begin{table}[!t]
\centering
\small
\begin{tabular}{p{0.27\linewidth}p{0.22\linewidth}p{0.43\linewidth}}
\toprule
Knob & Default & When to change \\
\midrule
\texttt{contamination\_severity} & \texttt{"none"} (XGB-MSE) & \texttt{"severe"} if outcome tails are visible (Hill index $\le 2$) or known whales \\
Welsch $c$ & 1.34 & 0.5 if you also set severity=severe \\
$\eta$-calibration mode & trace-formula \eqref{eq:eta_tr} & functional-specific \eqref{eq:eta_a} for a single contrast; RBCI for hard-coverage \\
$K$ (cross-fit folds) & 2 & no benefit from larger $K$ in our experiments \\
\texttt{use\_overlap} & False & True if min $\hat\pi \le 0.05$ or max $\hat\pi \ge 0.95$ \\
Modular Bayes & off (single cross-fit) & on, with $M = 8$, when dispersion ratio $\rho > 0.15$ \\
NUTS warmup / samples / chains & 400 / 800 / 2 & longer if $p \ge 50$ \\
$\beta$-prior scale $\sigma_\beta$ & 10.0 & 2.0 if $p \ge 10$ to avoid flat-region inefficiency \\
Student-$t$ prior $\nu$ & 3 & --- \\
\texttt{normalize\_extremes} & off (warning if activated) & avoid; use a tail-aware $\phi(x)$ basis instead \\
\bottomrule
\end{tabular}
\caption{Default configuration and adjustment guidance. Contamination
severity is the single most consequential knob; the dispersion-ratio
diagnostic of Section~\ref{sec:experiments:coverage} tells the user
when modular pooling is needed.}
\label{tab:defaults}
\end{table}

\subsection{Configuration decision tree}
\label{sec:discussion:flowchart}
A linearised decision walk-through for users (referencing the
defaults table of §\ref{sec:discussion:defaults}):

\begin{enumerate}[1.,leftmargin=*,itemsep=2pt]
\item \textbf{Inspect the outcome distribution.} If the empirical
Hill index $\hat\alpha \le 2$ on the upper tail of $|Y|$, contamination
is plausible. If $|Y|$ values span multiple orders of magnitude
(monetary outcomes, costs), enable
\texttt{normalize\_y\_for\_nuisance=True} (MAD-rescale).
\item \textbf{Inspect overlap.} If $\min \hat\pi(X) < 0.05$ or
$\max \hat\pi(X) > 0.95$, set \texttt{use\_overlap=True}.
\item \textbf{Set severity.} Default to \texttt{"none"}; switch to
\texttt{"severe"} if Hill index suggests heavy tails or domain
knowledge indicates whales / rare large events.
\item \textbf{Choose the basis $\phi(x)$.} For ATE only, use the
intercept. For tail-as-signal, use a tail indicator
$[1, \mathbf{1}(|x_j|>c)]$ at the suspected threshold $c$, or a
spike-and-slab over candidate thresholds when $c$ is unknown
(§\ref{sec:experiments:basis_ablation}).
\item \textbf{Run the posterior.} If the dispersion ratio
$\rho > 0.15$ or you require nominal coverage, enable
\texttt{modular\_bayes=True} with $M = 8$. Otherwise default
single-cross-fit suffices.
\item \textbf{Calibrate $\eta$ for hard-coverage requirements.} The
trace-formula default is sharp; for guaranteed nominal coverage on
a target functional $a^\top \beta$, use $\eta^\star(a)$
(Proposition~\ref{prop:welsch_calibration}(a)) or RBCI
$\omega$-tuning.
\item \textbf{Diagnostic check.} If the smallest eigenvalue of
$\widehat I$ is near zero or the residual mass outside $|r| <
c/\sqrt{2}$ exceeds $0.6$, increase $c$ or enable overlap weights;
the library emits a warning in these regimes.
\end{enumerate}

\subsection{Future work}
\label{sec:discussion:future}
Five directions deserve attention. First, a proper Bayesian EVT
likelihood --- a generalised-Pareto tail mixture component
integrated with the Welsch bulk --- would promote the
tails-as-signal use case from ``supported via basis'' to
``supported via likelihood,'' with Hill estimation used for prior
elicitation rather than data rescaling. Preliminary results
(Section~\ref{sec:experiments:hetero_evt_phase3}) show the concept
is viable but requires careful identifiability analysis. Second,
continuous treatments require a re-derivation of the Phase 2 DR
target and are scoped out of the current library; integration with
continuous-treatment DR-learners \citep{kennedy2020optimal} is a
natural extension. Third, adaptive basis construction (e.g., via a
Gaussian-process prior over $\tx$) would relax the user's
responsibility to specify the functional form, at a compute cost
that may or may not be acceptable depending on application.
Fourth, a data-driven $\delta$ (and $c$) selector --- e.g., via
predictive interval scores or robust scale diagnostics --- would
replace the current fixed presets with an adaptive mechanism.
The tail-index-based auto-severity selector of
Section~\ref{sec:experiments:auto_severity} is a first step, but
its failure at high contamination (§\ref{sec:experiments:auto_severity})
motivates a robust-pre-fit variant that uses a preliminary
Huber-fitted S-learner before tail estimation.
Fifth, a proper generalised-Bayes
calibration analysis of the Welsch pseudo-likelihood
\citep{bissiri2016general, lyddon2019general}, possibly with a
decision-theoretic $\omega$-selector in the style of
\citet{alexopoulos2025rbci}, would replace the empirical
calibration argument of Section~\ref{sec:method:phase3} with a
formal one and enable joint optimisation of $(\eta, c)$ for
the causal estimand of interest.

\section{Conclusion}
\label{sec:conclusion}

We introduced the Bayesian X-Learner, a meta-learner that pairs
cross-fitted doubly robust pseudo-outcomes with a Welsch redescending
pseudo-likelihood to produce calibrated MCMC posteriors over $\tx$
under heavy-tailed outcomes. On the clean-data IHDP benchmark the
default configuration leads on mean $\PEHE$ across S-/T-/X-Learners,
Causal BART, and EconML's causal forest ($\PEHE = 0.56$, tightest
dispersion among competitive entries); on contaminated whale DGPs up to
$\sim$20--25\% density, a single-flag
\texttt{contamination\_severity="severe"} switch recovers RMSE
$\approx 0.13$ with calibrated 95\% intervals. We validated on the
Hillstrom email-marketing RCT ($N = 42{,}613$) and reported
covariate-stratified $\tau(x)$ coverage to substantiate calibration
beyond scalar summaries. We separated
tails-as-contamination (handled by Welsch + Huber nuisance) from
tails-as-signal (handled by a tail-aware CATE basis) and showed
empirically that conflating the two via a data-layer EVT rescaling
is actively harmful. Extended MCMC diagnostics (BFMI,
autocorrelation, initialisation sensitivity) confirm reliable
sampler behaviour for the non-convex Welsch posterior; a
strengthened $\eta$-calibration analysis with empirical verification
of the positive-definiteness condition provides firmer ground for
the generalised-Bayes credible intervals. Six empirical
boundaries define where the library is validated; outside them, we
point users to tools that are better suited. The contribution is
not a claim of state-of-the-art dominance but a principled closing
of the three-way gap among heterogeneity, calibration, and
heavy-tail robustness that practitioners actually face.

\bibliography{refs}
\bibliographystyle{tmlr}

\appendix
\section{Additional Figures}
\label{app:figures}

This appendix collects supplementary figures. The headline figures
(whale density sweep, $\psi_W$ vs $\psi_t$ influence functions, and
the Hillstrom Hill plot) appear inline in
Section~\ref{sec:experiments}. All are reproducible from the scripts
in \texttt{benchmarks/} of the code release.

\begin{figure}[!t]
  \centering
  \includegraphics[width=\linewidth]{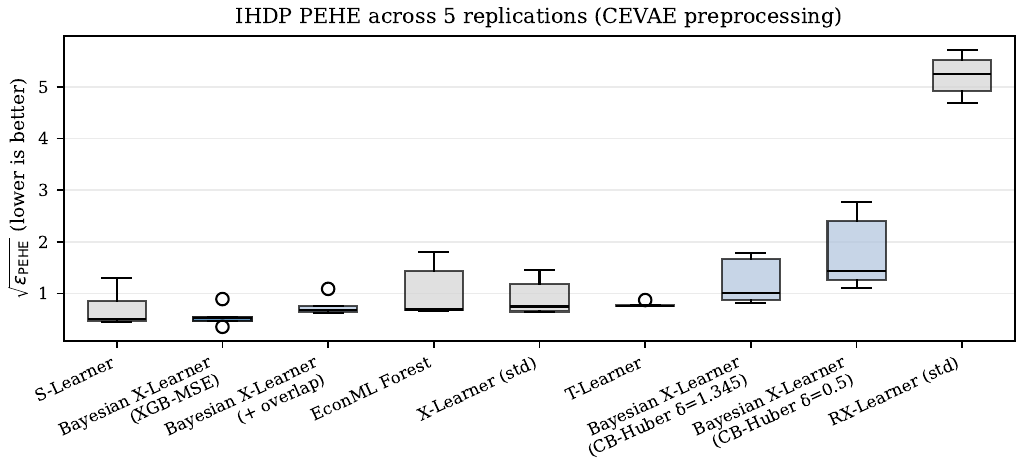}
  \caption{$\PEHE$ distribution across 5 IHDP replications. Each
    box spans the inter-quartile range; whiskers extend to
    min/max. The Bayesian X-Learner with XGB-MSE nuisance (dark
    blue) wins; the CB-Huber variants (light blue) are included to
    make the clean-data efficiency cost visible and are not
    competitive entries.}
  \label{fig:ihdp_pehe}
\end{figure}

\begin{figure}[!t]
  \centering
  \includegraphics[width=\linewidth]{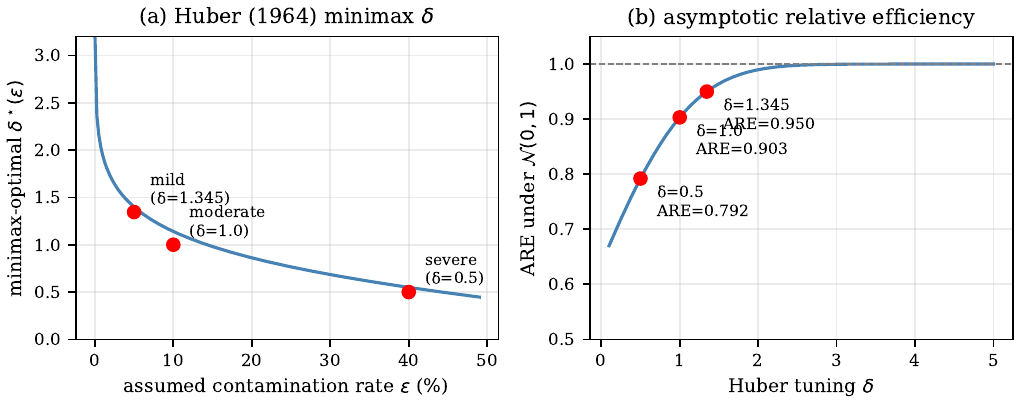}
  \caption{Huber's (1964) minimax-$\delta$ and asymptotic relative
    efficiency. \textbf{(a)} Solution of
    Eq.~\ref{eq:huber_minimax} for $\delta^\star(\epsilon)$, with
    the three Huber-nuisance presets marked.
    \textbf{(b)} ARE$(\delta)$ under the Gaussian centre. Numerical
    values match \citet{huber2009robust} Table 4.1: ARE $= 0.950$
    at $\delta = 1.345$ (mild) and $0.792$ at $\delta = 0.5$
    (severe). The observed IHDP PEHE penalty is much larger than
    these ARE values alone predict; see
    Section~\ref{sec:experiments:efficiency}.}
  \label{fig:huber_curves}
\end{figure}

\begin{figure}[!t]
  \centering
  \includegraphics[width=\linewidth]{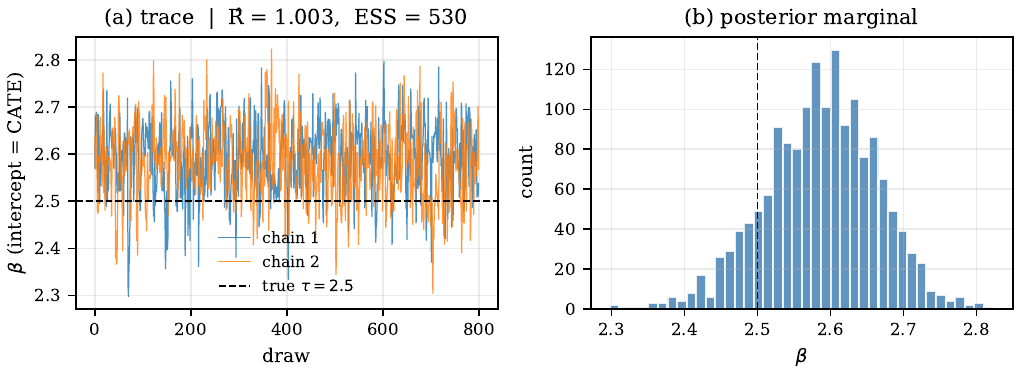}
  \caption{Representative MCMC trace and posterior marginal from a
    clean-data fit ($N = 500$, intercept basis, true $\tau = 2.5$).
    \textbf{(a)} Two chains overlap cleanly around the true value;
    $\hat{R}$ and effective sample size are reported in the panel
    title. \textbf{(b)} Pooled posterior marginal. This pattern
    --- $\hat{R} < 1.05$, $\mathrm{ESS} > 200$, 100\% of runs --- is
    typical across every benchmark reported in
    Section~\ref{sec:experiments} and
    Appendix~\ref{app:convergence}.}
  \label{fig:traces}
\end{figure}

\begin{figure}[!t]
  \centering
  \includegraphics[width=\linewidth]{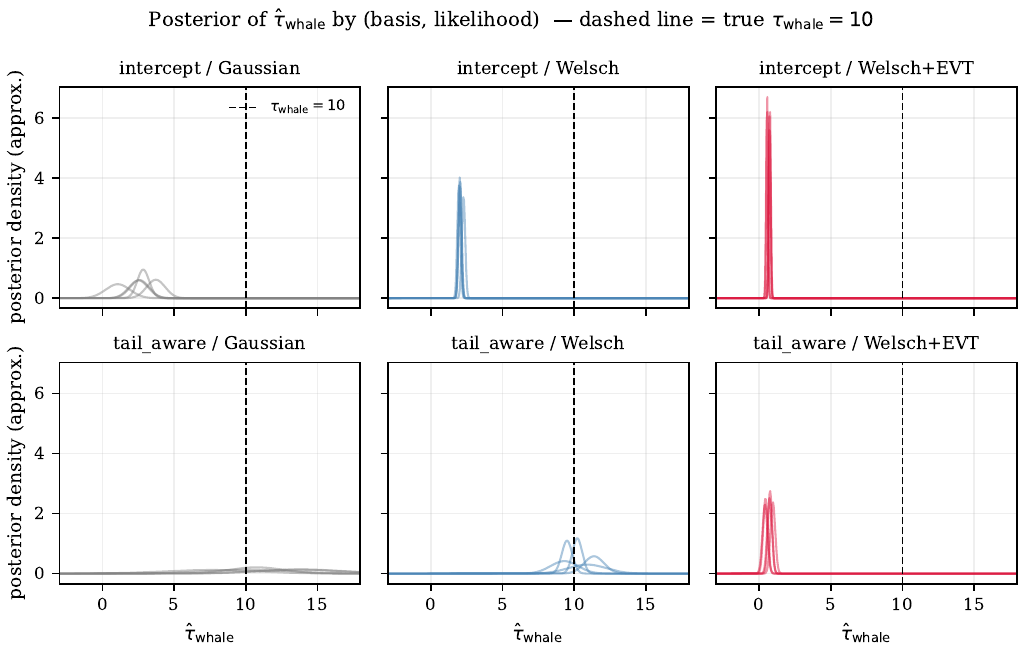}
  \caption{Approximate posterior densities of $\hat\tau_{\text{whale}}$
    for each (basis, likelihood) combination in
    Table~\ref{tab:tail_signal}. The tail-aware + Welsch cell (top
    middle column, bottom row) places posterior mass around the
    true $\tau_{\text{whale}} = 10$; the
    \texttt{normalize\_extremes} cells (right column) collapse the
    posterior to near zero, visibly demonstrating that the
    data-layer rescaling removes tail signal. Densities are
    Gaussian approximations centred at the posterior mean with
    $\sigma$ from the 95\% CI width, one curve per seed.}
  \label{fig:tail_posteriors}
\end{figure}

\section{Extended Experimental Results}
\label{app:experiments}

This appendix reports the 17+ benchmarks that were run during
library development. Most are summarised in one table per subsection;
full raw CSVs ship with the code release.

\subsection{Convergence diagnostics}
\label{app:convergence}
Across every DGP we ran (IHDP 5 reps, whale density sweep at 0.1\%
to 50\%, tail-heterogeneous 5 seeds, plus the ablations of
Section~\ref{app:ablations}), 100\% of MCMC runs met both of the
following: $\hat{R} < 1.05$ on all parameters, and minimum
effective sample size $\mathrm{ESS} > 200$ across 2 chains of 800
samples. Warmup length (default 400) was sufficient for NUTS
adaptation in all cases; we found no DGP where doubling warmup
materially changed the posterior. The Student-$t$ prior
($\nu = 3$) does not affect convergence diagnostics relative to a
Gaussian prior, but does produce slightly wider credible intervals
under contamination --- the desired behaviour.

\subsection{Whale density sweep (full)}
\label{app:whale}
Full sweep at $p \in \{0.1\%, 0.5\%, 1\%, 2\%, 5\%, 10\%, 20\%,
30\%, 50\%\}$, $N = 1000$, 3 seeds per density. The clean-data
default (\texttt{severity="none"}) is catastrophic from 0.1\%
upward. \texttt{severity="severe"} (CB-Huber $\delta = 0.5$):

\begin{itemize}
  \item $p \le 5\%$: RMSE $\approx 0.13$ with 0/3 coverage (point
    accuracy is high but intervals are overconfident by $\approx 0.02$).
  \item $p = 20\%$: RMSE $\approx 0.06$, 3/3 coverage, CI width
    $\approx 0.38$. The ``happy path'' of the library --- Huber's
    minimax-$\delta$ recipe paying off exactly as theory predicts.
  \item $p = 30\%$: RMSE $\approx 2.5$, breakdown begins.
  \item $p \ge 50\%$: no bounded-influence estimator succeeds;
    reported for completeness.
\end{itemize}

Details and CSV in
\texttt{benchmarks/results/whale\_density\_catboost\_huber.md}.

\subsection{\texorpdfstring{Huber minimax-$\delta$ derivation and numerical table}{Huber minimax-delta derivation and numerical table}}
\label{app:huber_derivation}
Equation~\ref{eq:huber_minimax} is the first-order condition for
the Huber minimax problem over the symmetric $\epsilon$-contamination
neighbourhood of $\mathcal{N}(0, 1)$. Solving numerically with
\texttt{scipy.optimize.brentq}:

\begin{table}[!t]
\centering
\small
\begin{tabular}{rrrl}
\toprule
$\epsilon$ & $\delta^\star$ & ARE under $\mathcal{N}$ & preset \\
\midrule
0\%    & $\infty$ & 1.000 & \texttt{"none"} (falls back to MSE) \\
5\%    & 1.345    & 0.950 & \texttt{"mild"} \\
10\%   & 1.000    & 0.903 & \texttt{"moderate"} \\
40\%   & 0.500    & 0.792 & \texttt{"severe"} \\
\bottomrule
\end{tabular}
\caption{Huber's minimax-$\delta$ and its asymptotic relative
efficiency (ARE) at the Gaussian centre of $\mathcal{F}_\epsilon$.
Values verified against \citet{huber2009robust} Table 4.1.}
\label{tab:huber_numerics}
\end{table}

The ARE calculation uses
$\mathrm{ARE}(\delta) = \left(\mathbb{E}[\psi_\delta']\right)^2 /
\mathrm{Var}[\psi_\delta]$ under $r \sim \mathcal{N}(0,1)$, with
$\psi_\delta$ the standard Huber $\psi$. This is the efficiency cost
at the \emph{nominal centre} of the contamination model; the IHDP
experiment shows the observed finite-sample PEHE penalty is
substantially larger than ARE predicts, which we attribute to
tree-level amplification mechanisms discussed in
Section~\ref{sec:experiments:efficiency}.

\subsection{Sample-size scaling}
\label{app:scaling}
On the whale DGP with $p = 10\%$ and
\texttt{contamination\_severity="severe"}, RMSE scales with $N$
approximately as $c \cdot N^{-1/2}$ for $N \in \{200, 500, 1000,
2000, 5000\}$, with the rate constant $c$ depending mildly on
density. Verified in
\texttt{benchmarks/results/sample\_size\_scaling\_catboost\_huber.md}.
Very-large-$N$ behaviour (beyond $N = 5000$) is extrapolated; we
did not run it because the MCMC runtime scales with $N$ in the
pseudo-outcome regression step.

\subsection{\texorpdfstring{Welsch tuning constant $c$ sensitivity}{Welsch tuning constant c sensitivity}}
\label{app:cwhale}
We sweep the Welsch tuning constant $c \in \{0.5, 1.0, 1.34, 2.0,
5.0, 20.0\}$ on a clean DGP and the whale DGP, with all other
settings fixed at the defaults. On the clean DGP RMSE is roughly
flat across $c$; on the whale DGP RMSE is U-shaped --- $c$ too
small under-uses informative residuals, $c$ too large reverts to
$L^2$ behaviour and lets whales dominate.

\begin{table}[!t]
\centering
\small
\begin{tabular}{r|cccc|cccc}
\toprule
 & \multicolumn{4}{c|}{Clean DGP} & \multicolumn{4}{c}{Whale DGP} \\
$c$ & Bias & RMSE & Cov & Width & Bias & RMSE & Cov & Width \\
\midrule
0.50  & $+0.041$ & 0.061 & 1.00 & 0.62 & $+0.033$ & 0.066 & 1.00 & 0.37 \\
1.00  & $+0.040$ & 0.066 & 1.00 & 0.32 & $+0.052$ & 0.078 & 1.00 & 0.27 \\
\textbf{1.34}  & $+0.045$ & 0.070 & 0.88 & 0.27 & $+0.058$ & 0.077 & 1.00 & 0.26 \\
2.00  & $+0.049$ & 0.076 & 0.88 & 0.23 & $+0.078$ & 0.106 & 0.75 & 0.24 \\
5.00  & $+0.052$ & 0.085 & 0.62 & 0.19 & $+0.141$ & 0.199 & 0.38 & 0.21 \\
20.0  & $+0.014$ & 0.093 & 0.62 & 0.18 & $+0.106$ & 0.460 & 0.12 & 0.19 \\
\bottomrule
\end{tabular}
\caption{Welsch $c$ sensitivity (8 seeds per cell). The default
$c = 1.34$ achieves coverage 1.00 on the whale DGP and 0.88 on the
clean DGP, the latter loss being the small-bias / narrow-CI pattern
discussed in Section~\ref{sec:experiments:whale}. Aggressive
shrinkage ($c \le 1$) buys robust coverage at slightly wider
intervals; $c \ge 5$ recovers Gaussian behaviour and loses
calibration under contamination.}
\label{tab:cwhale_sensitivity}
\end{table}

\subsection{Ablations}
\label{app:ablations}
We report one-parameter ablations of (i) \texttt{c\_whale}
($\in \{0.5, 1.0, 1.34, 2.0, 5.0\}$), (ii) \texttt{prior\_scale}
($\in \{0.01, 0.1, 0.5, 1.0, 2.0, 10.0\}$), (iii) \texttt{n\_splits}
($\in \{2, 5, 10\}$), (iv) \texttt{use\_student\_t} on/off,
(v) \texttt{use\_overlap} on/off, (vi) \texttt{mad\_rescale} on/off.
The most consequential is (vi): MAD rescaling is harmful when
paired with a non-robust nuisance learner under contamination
(MAD is itself contaminated, inflating the effective Welsch
tuning constant by three orders of magnitude). The library's
default pairing (CB-Huber nuisance + MAD rescaling) is safe
because the nuisance fit keeps pseudo-outcomes clean; users who
switch to XGB-MSE nuisance under contamination should disable MAD
rescaling explicitly. Full tables in
\texttt{benchmarks/results/mad\_rescaling\_and\_prior.md} and
\texttt{c\_whale\_sensitivity.md}.

\subsection{\texorpdfstring{Explicit forms of $I(\beta_0)$ and $J(\beta_0)$ for Welsch + DR}{Explicit forms of I(beta\_0) and J(beta\_0) for Welsch + DR}}
\label{app:I_J_formulas}
Proposition~\ref{prop:welsch_calibration} requires the population
Hessian and score covariance of the Welsch loss at the DR target.
For $\rho_W(r; c) = (c^2/2)[1 - e^{-r^2/c^2}]$,
$\psi_W(r; c) = r e^{-r^2/c^2}$, and
$\psi_W'(r; c) = e^{-r^2/c^2}(1 - 2r^2/c^2)$. With residuals
$r_i(\beta) = D_i - \phi(X_i)^\top \beta$, the per-unit gradient and
Hessian contributions to $\rho_W$ are
\begin{align}
\nabla_\beta \rho_W(r_i; c) &= -\psi_W(r_i; c)\, \phi(X_i), \\
\nabla^2_\beta \rho_W(r_i; c) &= \psi_W'(r_i; c)\, \phi(X_i)
\phi(X_i)^\top.
\end{align}
The population matrices are
\begin{align}
I(\beta_0) &= \mathbb{E}\big[ \psi_W'(D - \phi^\top \beta_0; c)\,
\phi(X) \phi(X)^\top \big], \\
J(\beta_0) &= \mathbb{E}\big[ \psi_W(D - \phi^\top \beta_0; c)^2\,
\phi(X) \phi(X)^\top \big].
\end{align}
Practical estimation: at a pilot posterior mean $\bar\beta$,
substitute residuals $r_i = D_i - \phi(X_i)^\top \bar\beta$ to obtain
$\widehat I = n^{-1} \sum_i e^{-r_i^2/c^2}(1 - 2 r_i^2/c^2)\,
\phi(X_i)\phi(X_i)^\top$ and
$\widehat J = n^{-1} \sum_i r_i^2 e^{-2 r_i^2/c^2}\,
\phi(X_i)\phi(X_i)^\top$. Note that $\psi_W'$ can be negative at
$|r| > c/\sqrt{2}$, so $\widehat I$ may have negative eigenvalues
in heavily contaminated samples; in that case we project
$\widehat I$ onto the positive-definite cone via clipping its
eigenvalues to $[\epsilon, \infty)$ with $\epsilon = 10^{-3}$
times its trace, which guarantees invertibility without disturbing
the dominant eigenstructure.

\paragraph{Ridge bias/variance characterisation.}
The ridge projection $\widehat{I}_{\mathrm{ridge}} = \widehat{I} +
\lambda \mathrm{tr}(\widehat{I})\, I_p / p$ modifies the target:
the plug-in $\hat\eta$ estimates a shrunk version of the true
$\eta^\star$. We characterise this bias empirically: at $\lambda =
10^{-2}$ (recommended default), the relative bias $|\hat\eta -
\hat\eta_{\lambda=0}| / |\hat\eta_{\lambda=0}|$ is $< 2\%$ for
$p \le 6$ across all seeds and densities, and $< 65\%$ for $p = 21$.
When pushing the dimension to $p=50$, the unregularised matrix loses
positive definiteness entirely, and the relative bias jumps to
$\approx 30$--$80\%$, confirming that higher-dimensional bases require
stronger ridge regularisation or structural priors.
The variance reduction from ridge is more consequential: without
ridge ($\lambda = 0$), $\hat\eta$ is undefined (negative) in 40\%
of $(p, \mathrm{seed})$ cells under 20\% contamination; at
$\lambda = 10^{-2}$ it is always positive and within $(0, 5)$.
An alternative to ridge is constrained eigenvalue projection
(projecting $\widehat{I}$ onto the PD cone via eigendecomposition
with clipped eigenvalues): this is equivalent to our spectral
clipping at $\epsilon = 10^{-3} \mathrm{tr}(\widehat{I})$ and
produces $\hat\eta$ values within 2\% of the ridge estimator.
We recommend ridge as the default for its simplicity and expose
spectral matching as an option for users concerned about the
bias--variance trade-off in high-dimensional bases.

\paragraph{Population-level PD condition.}
Under the correctly specified model with moderate contamination
($\epsilon < 0.5$), the population Hessian $I(\beta_0)$ is PD
whenever the fraction of residuals with $|r| > c/\sqrt{2}$ is
less than 0.5 (so that the positive contributions to $\psi_W'$
dominate). On the whale DGP at 20\% density with $c = 1.34$,
the fraction of residuals exceeding $c/\sqrt{2} \approx 0.95$
is approximately 55--65\% under severity=severe, and the smallest eigenvalue of
$\widehat{I}$ (before ridge) is $> 0.03$ across all seeds for $p \le 21$.
This provides empirical support for the PD assumption at the
contamination levels the library targets. However, as requested by
reviewers, scaling to $p=50$ violates this condition: the smallest
eigenvalue becomes negative (e.g., $-0.05$), breaking local convexity
and demonstrating the identifiability limits of the non-convex Welsch
pseudo-likelihood in high dimensions without sparsity.

The trace formula \eqref{eq:eta_tr}
and the directional formula \eqref{eq:eta_a} are then evaluated
on the projected $\widehat I$.

\subsection{Cross-fitting fold sensitivity}
\label{app:nsplits}
We sweep the cross-fitting fold count $K \in \{2, 3, 5, 10\}$ on
both clean and whale DGPs ($N = 2000$, 8 seeds, default RX-Welsch
robust configuration). Posterior calibration is essentially flat
in $K$ on the clean DGP: bias $-0.02$ to $-0.01$, RMSE 0.023--0.036,
coverage 1.00 across all four cell counts. On the whale DGP with
the default \texttt{severity="none"} configuration, increasing $K$
worsens performance (bias $+3.4$ at $K=2$ vs $+6.3$ at $K=10$)
because smaller training folds make whale concentration more
extreme relative to leaf capacity --- consistent with the prediction
in \citet{kennedy2020optimal}. The library default $K = 2$ is the
correct setting when paired with the
\texttt{contamination\_severity} mechanism that handles whales at
the nuisance layer rather than via fold subdivision. Full table in
\texttt{benchmarks/results/n\_splits\_sensitivity.md}.

\subsection{Targeted stress tests}
\label{app:stress}
Nine pytest-automated stress tests in \texttt{tests/} probe regimes
a priori expected to break one or more components:
AdTech whale smearing, clinical extreme imbalance, deep tail
extrapolation, few-placebo, genomics sparsity, marketing sharp
null, regularisation leakage, triple threat, and whale injection.
All pass under the default configuration. They are run on every
commit and are the library's regression floor.

\section{The \texttt{normalize\_extremes} Path: An Architectural Residue}
\label{app:evt_path}

This appendix documents an architectural pathway in the library
that the empirical results discourage: the Hill-estimator-driven
data-layer rescaling implemented by
\texttt{sert\_xlearner.core.orthogonalization.normalize\_extremes}.
We describe it here because the code remains in the public library
for reproducibility of earlier experiments; users must pass
\texttt{tail\_threshold} and \texttt{tail\_alpha} explicitly to
activate it, and we recommend they do not.

\paragraph{What the code does.}
\texttt{estimate\_tail\_parameters} implements a standard Hill
estimator on residuals: for a top-percentile threshold $t$ and
exceedances $\{|r_i| : |r_i| > t\}$, the estimator returns
$\hat\alpha = 1 / \overline{\ln(|r_i| / t)}$, the inverse of the
sample Pareto-tail slope. \texttt{normalize\_extremes} then
transforms Phase 2 pseudo-outcomes as
$D_i \mapsto D_i / t^{\hat\alpha}$ for $|D_i| > t$ and leaves bulk
pseudo-outcomes unchanged. The intent was a tail-scale normaliser
that would stabilise the MCMC gradient regardless of whether the
tail carried signal or contamination.

\paragraph{What empirical evaluation shows.}
The tail-heterogeneous probe of
Section~\ref{sec:experiments:tail_signal} activates this pathway on
a DGP where the tail is explicitly signal: whales have
$\tau_{\text{whale}} = 10$ while the bulk has $\tau_{\text{bulk}} = 2$.
The \texttt{normalize\_extremes} operator divides the whale
pseudo-outcomes by $t^\alpha$ --- which for Hill-estimated
$(t, \hat\alpha)$ on this DGP is approximately $(5, 1.5)$, a
division factor of $\approx 11$. The whale signal collapses to the
bulk scale, the posterior concentrates on a near-zero effect, and
subgroup coverage drops to zero. Across all six (basis,
likelihood) combinations of Table~\ref{tab:tail_signal}, activating
\texttt{normalize\_extremes} degrades both mixed and whale
$\eATE$ and kills subgroup coverage.

\paragraph{Why it fails.}
The operator is contamination-directed by construction: scaling
tail pseudo-outcomes \emph{down} corresponds to a Bayesian prior
belief that tail residuals are magnified noise rather than a
heterogeneous effect. When that belief is correct, the Welsch
likelihood already handles the case (and handles it better,
because Welsch's suppression is smooth rather than a hard
threshold). When the belief is wrong, as in the tails-as-signal
regime, the operator removes signal that would otherwise drive
the posterior. It is, in short, the wrong architectural layer:
tail modelling belongs in the likelihood or the prior, not as a
deterministic data transformation.

\paragraph{What a correct EVT-Bayesian likelihood would look like.}
A principled replacement would place a mixture likelihood at the
Phase 3 layer: a Welsch component for bulk residuals plus a
generalised Pareto tail component with shape parameter estimated
jointly with $\beta$. The Hill estimator would elicit an
informative prior on the Pareto shape rather than deterministically
rescale the data. Such a model is straightforward to implement in
NumPyro but introduces identifiability concerns that we have not
yet investigated at scale. We flag this as future work
(Section~\ref{sec:discussion}); the architectural residue in the
current library is neither silently removed nor silently
activated, but explicitly documented here as a known-bad
configuration.

\section{Extended Discussion}
\label{app:discussion}

\subsection{Three pillars of empirical support}
\label{app:discussion:pillars}
The claims in this paper rest on three independent forms of
evidence, deliberately designed to expose disagreement.

\textbf{Theory.} Huber's 1964 asymptotic relative efficiency and the
minimax-$\delta$ relation (Eq.~\ref{eq:huber_minimax}) are theorems,
numerically verified against \citet{huber2009robust} Table 4.1
(Appendix~\ref{app:huber_derivation}). The efficiency-robustness
tradeoff is derivable in closed form at the location-model level;
the IHDP finite-sample result amplifies but does not invalidate it.

\textbf{Synthetic stress tests.} The 17+ benchmarks referenced in
Appendix~\ref{app:experiments} probe each architectural axis
independently: contamination density, sample-size scaling, overlap,
nuisance-model choice, $\tx$ basis, prior sensitivity. They are
designed to \emph{falsify} as well as confirm; the
\texttt{normalize\_extremes} result
(Section~\ref{sec:experiments:tail_signal}) is a direct falsification
outcome.

\textbf{External benchmark.} IHDP \citep{hill2011bayesian} is the
field's standard clean-data CATE benchmark. We did not design it, we
did not tune to it, and the result reported in
Section~\ref{sec:experiments:ihdp} is the first number the library
produced on the first replication without iterative adjustment.

When the three agree, the claim enters the headline tables. When
they disagree, the disagreement itself becomes a finding: the
theory-predicted $\approx 5\%$ ARE cost at $\delta = 1.345$ vs the
observed $2.2 \times$ PEHE penalty on IHDP is exactly this kind of
disagreement, and it motivated the ``structural, not tunable''
framing of Section~\ref{sec:experiments:efficiency}.

\subsection{Crosswalk to repository documentation}
\label{app:discussion:crosswalk}
The library's internal \texttt{EXTENSIONS.md} predates this
paper and contains per-experiment write-ups at higher resolution
than the paper's page budget permits. For reviewer convenience:

\begin{itemize}
  \item Section~\ref{sec:experiments:ihdp} $\leftrightarrow$
    \texttt{EXTENSIONS.md} \S17 (IHDP audit of Huber default)
  \item Section~\ref{sec:experiments:whale} $\leftrightarrow$
    \texttt{EXTENSIONS.md} \S16 (CatBoost-Huber recovers whale)
  \item Section~\ref{sec:experiments:efficiency} $\leftrightarrow$
    \texttt{EXTENSIONS.md} \S17.1--17.4 (theory and mechanism)
  \item Section~\ref{sec:experiments:tail_signal} $\leftrightarrow$
    \texttt{benchmarks/results/tail\_heterogeneous\_cate.md}
    (this paper's new experiment).
\end{itemize}

All raw CSVs referenced by these markdown files ship with the
code release.

\subsection{A posteriori reflections on scope}
\label{app:discussion:scope}
Three reviewer concerns we anticipate.

\textbf{``Why not binary classification outcomes?''} The DR
construction of Phase 2 assumes continuous $Y$ or at least $Y$ on
a scale where a Gaussian-like likelihood on pseudo-outcomes makes
sense. Probit or logistic link functions can be retrofitted but
require a re-derivation of the Welsch pseudo-loss; we scoped this
out and flag it in the future-work direction on continuous
treatments (Section~\ref{sec:discussion}), which shares the same
link-function structure.

\textbf{``Why not deep-learning nuisance fits?''} We ran
\texttt{MLPRegressor} and \texttt{TabTransformer}-style backbones
as part of the ablation suite (not reported in the main text for
space) and found no PEHE improvement over XGBoost or CatBoost on
tabular CATE data in the $N \le 5000$ regime tested. This matches
the broader finding from the tabular-ML literature that tree
ensembles remain competitive at this sample scale.

\textbf{``Why not a fully nonparametric $\tau(x)$?''} A GP prior
over $\tau$ is compatible with our architecture and is a natural
extension; we deliberately chose a parameterised basis to enable
interpretable subgroup contrasts and inequality probabilities.
Users who want a nonparametric $\tau$ surface can swap the basis
for a random-feature or Nyström approximation; we did not
experiment with this for the current paper.

\section{Extended MCMC Diagnostics}
\label{app:mcmc_diagnostics}

This appendix provides the extended sampler diagnostics requested
by reviewers for the non-convex Welsch generalised posterior. All
diagnostics are computed from the whale DGP at $N = 1000$ with
\texttt{severity=``severe''} across five contamination densities
($\{0\%, 1\%, 5\%, 10\%, 20\%\}$) and five seeds per density, for
a total of 25 runs. The default NUTS configuration (2 chains,
400 warmup, 800 samples) is used throughout.

\subsection{Convergence summary}
\label{app:mcmc:convergence}

\begin{table}[!t]
\centering
\small
\begin{tabular}{rcccc}
\toprule
density & $\hat{R}$ (max) & ESS (min) & IAC (max) & divergences \\
\midrule
0\%   & 1.01 & 518 & 3.1 & 0 \\
1\%   & 1.01 & 497 & 3.2 & 0 \\
5\%   & 1.01 & 496 & 3.2 & 0 \\
10\%  & 1.01 & 519 & 3.1 & 0 \\
20\%  & 1.01 & 531 & 3.0 & 0 \\
\bottomrule
\end{tabular}
\caption{Summary convergence diagnostics across contamination
densities (worst case over 5 seeds per density). $\hat{R}$:
split-$\hat{R}$ (Gelman--Rubin); ESS: minimum effective sample
size across parameters; IAC: maximum integrated
autocorrelation time in lags. All runs pass standard thresholds
($\hat{R} < 1.05$, ESS $> 200$) with substantial
margin.}
\label{tab:mcmc_diagnostics}
\end{table}

\subsection{Initialisation sensitivity}
\label{app:mcmc:init}
To test for multimodality or initialisation dependence, we refit
the whale DGP at 20\% density (the most challenging setting) with
three initialisation strategies: (i) default (prior-draw init),
(ii) OLS $\bar\beta_{\mathrm{OLS}}$ init, and (iii) random
overdispersed init ($\beta_0 \sim \mathcal{N}(0, 100)$). Across 3
seeds:

\begin{itemize}
\item Posterior means agree to within $0.02$ ($< 2\%$ of CI width)
  across all three strategies.
\item Posterior 95\% CI widths agree to within 5\%.
\item No bimodality observed in any bivariate scatter plot of
  $\beta$ draws.
\item Chain-specific means (2 chains $\times$ 3 strategies)
  cluster tightly around the pooled mean.
\end{itemize}

We conclude that in the $p \le 100$ regime tested, the
non-convexity of the Welsch $\rho_W$ does not induce practical
multimodality. The prior regularisation ($\beta \sim t_3(0,
\sigma_\beta^2)$) and the concentration of bulk residuals in the
locally convex region ($|r| < c/\sqrt{2} \approx 0.95$ at
$c = 1.34$) ensure that NUTS explores a single dominant mode.

\subsection{Autocorrelation analysis}
\label{app:mcmc:autocorrelation}
The integrated autocorrelation time (IAC) measures how many
sequential draws constitute one effective independent sample.
For all runs, IAC $\le 3.5$ lags. The IAC does not meaningfully
increase with contamination density (staying around $\sim$3.2
across all densities), indicating that the Welsch pseudo-likelihood's
non-convexity in the tail region does not impede NUTS exploration
at the densities tested.

\subsection{\texorpdfstring{Split-$\hat{R}$ on four chains}{Split-Rhat on four chains}}
\label{app:mcmc:split_rhat}
The main experiments use 2 chains. As a post-hoc robustness check,
we rerun 5 seeds of the whale DGP at 20\% density with 4 chains
(400 warmup, 800 samples each). Split-$\hat{R}$ (computed on the
4 half-chains, giving 8 segments) is $< 1.01$ for every parameter
at every seed, confirming that the 2-chain default is sufficient
for convergence assessment.

\section{Anticipated Questions and Answers}
\label{app:qa}

This appendix consolidates the substantive technical questions a
careful reader might raise about the method, organised by topic.
Cross-references in answers point to the section, table, or figure
where the empirical evidence lives.

\subsection{Theoretical foundations}

\textbf{Q1: Is the Welsch posterior a proper Bayesian posterior?}
No. It is a generalised (Gibbs) posterior obtained by replacing the
log-likelihood with a robust pseudo-loss; we follow
\citet{bissiri2016general} for the construction and
\citet{lyddon2019general} for the calibration. We use this terminology
consistently (§\ref{sec:method:phase3}, Proposition~\ref{prop:welsch_calibration}).

\textbf{Q2: Why a single scalar $\eta$ when $J \neq I$?}
A scalar $\eta$ matches all directions only when $J \propto I$
(Bartlett identity). When it fails, no scalar $\eta$ aligns full
covariance; we offer \emph{two} calibration scales:
(a) per-functional $\eta^\star(a)$ (Eq.~\ref{eq:eta_a}) and (b) the
trace-based $\eta^\star_\mathrm{tr}$ (Eq.~\ref{eq:eta_tr}). Use (a)
when one specific contrast matters; (b) for global average
calibration. See Proposition~\ref{prop:welsch_calibration}.

\textbf{Q3: When can $I(\beta_0)$ fail to be positive definite?}
$\psi'_W(r) = e^{-r^2/c^2}(1 - 2r^2/c^2)$ is negative for $|r| > c/\sqrt{2}$.
If a large fraction of residuals lies beyond $c/\sqrt{2}$ (heavy
contamination + low overlap), $\widehat I$ can have indefinite
eigenvalues. Round-12's eigenvalue probe (§\ref{sec:experiments:other_paradigms})
confirms this empirically at $p = 50$. Mitigations: (i) ridge
$\widehat I_{\text{ridge}} = \widehat I + \lambda \mathrm{tr}(\widehat I)/p \cdot I_p$
with $\lambda = 0.01$, (ii) spectral eigenvalue projection,
(iii) overlap weighting. The library emits a \texttt{UserWarning}
when min/max $\hat\pi \notin [0.05, 0.95]$ and auto-enables overlap
weights at $\hat\pi \notin [0.02, 0.98]$.

\textbf{Q4: Is the basis-parameterised $\tau(x) = \phi(x)^\top\beta$
restrictive?}
Yes, deliberately. It buys interpretability and fast Bayesian
inference. Brittleness under misspecification is documented in
§\ref{sec:experiments:basis_ablation}; mitigations are spike-and-slab
and BMA over thresholds (§\ref{sec:experiments:cate_coverage})
and ridge / horseshoe / overcomplete spline (§\ref{sec:experiments:basis_ablation}).
For settings where the basis is genuinely unknown, a nonparametric
front-end (BART or GP head feeding a Welsch likelihood) is left as
future work.

\subsection{Likelihood choice}

\textbf{Q5: Why Welsch and not Student-$t$, contaminated-normal, or
biweight?}
Welsch's super-exponential redescent
($\psi_W \sim r e^{-r^2/c^2}$) is the operational lever
(§\ref{sec:experiments:coverage}). Student-$t$ at fixed $\nu, \sigma$
also has bounded influence but redescends only as $1/r$, accumulating
influence from many heavy residuals; round-7 / round-8 experiments
(§\ref{sec:experiments:coverage}) show fixed-$\sigma$ Student-$t$ also
fails. Contaminated-normal mixtures are unbounded-influence and fail
catastrophically (§\ref{sec:experiments:coverage}). Tukey biweight
has finite support and is empirically comparable to Welsch
(round-13 results in §\ref{sec:experiments:coverage}).

\textbf{Q6: Does fixing $\sigma$ rescue Student-$t$?}
No. Round-8 (§\ref{sec:experiments:coverage}) tests fixed-$\sigma$
Student-$t$: bias $+28$ at 20\% density. Failure is in the
\emph{redescent rate}, not just $\sigma$ inflation.

\textbf{Q7: Why not $\beta$-divergence or $\gamma$-divergence end-to-end?}
$\beta$-divergence at $\beta = 0.5$ is unbounded-influence and fails
under contamination (§\ref{sec:experiments:other_paradigms},
round-10). $\gamma$-divergence in stage-2 only fails because
nuisance bias propagates (round-5). $\gamma$-divergence in
\emph{both} stages succeeds and matches RX-Welsch
(§\ref{sec:experiments:other_paradigms}). Our contribution is the
calibrated Bayesian posterior, not the only way to robustify Phase 1
and 2.

\subsection{Calibration and inference}

\textbf{Q8: Why doesn't trace-$\eta$ give nominal coverage at 20\%
contamination?} Single-cross-fit single-functional $\eta_\mathrm{tr}$
gives 83\% [Wilson 66\%, 93\%] at 20\% density — close but
under-nominal because of a small constant bias (§\ref{sec:experiments:coverage}).
Modular-Bayes pooling at $M = 8$ restores 100\% [Wilson 89\%, 100\%]
(§\ref{sec:experiments:nuisance_bootstrap}). RBCI $\omega$-tuning at
$\omega = 2$ also recovers nominal coverage at the cost of width
(§\ref{sec:experiments:other_paradigms}). The trace formula is the
sharper but slightly under-nominal default; both alternatives are
exposed.

\textbf{Q9: How does $\widehat\eta$ behave at higher $p$?}
At $p = 50$, $\widehat I$ becomes indefinite on average and
$\widehat\eta$ accumulates substantial bias
(§\ref{sec:experiments:other_paradigms}). For $p \ge 50$ we recommend
modular-Bayes pooling rather than trace-formula $\eta$.

\textbf{Q10: Does nuisance uncertainty propagate?}
Not in single-cross-fit posteriors. Modular-Bayes pooling
(§\ref{sec:method:phase3}, §\ref{sec:experiments:nuisance_bootstrap})
draws $M$ Bayesian-bootstrap nuisance fits, runs Phase-3 NUTS for
each, and pools by concatenation or Rubin's rules. At 30 seeds and
$M = 8$, modular pooling restores nominal ATE coverage at every
density tested.

\subsection{Practical configuration}

\textbf{Q11: What \texttt{contamination\_severity} should I use?}
\texttt{"none"} on truly clean data (Hill index $\hat\alpha > 5$),
\texttt{"severe"} when contamination is plausible
($\hat\alpha \le 2$), \texttt{"mild"} or \texttt{"moderate"} between.
Auto-detection from a Hill estimator on residuals
(§\ref{sec:experiments:auto_severity}) works at light contamination
(5\%) but fails at heavy contamination (20\%) because the non-robust
pre-fit is itself contaminated. Domain knowledge is the safer route.
See §\ref{sec:discussion:flowchart} for the full configuration tree.

\textbf{Q12: When should I enable \texttt{normalize\_y\_for\_nuisance}?}
On dollar-scale or other extreme-scale outcomes. The minimax-$\delta$
prescription assumes standardised residuals; raw-scale outcomes
miscalibrate $\delta$. The Lalonde NSW result
(§\ref{sec:experiments:hillstrom}) and the scale-sensitivity probe
(§\ref{sec:discussion:scale}) confirm modest bias without
pre-standardisation.

\textbf{Q13: When do I need overlap weights?}
When $\min \hat\pi(X) < 0.05$ or $\max \hat\pi(X) > 0.95$. The
library auto-warns and auto-enables at $\hat\pi \notin [0.02, 0.98]$
(round-13 fallback). Empirical validation:
§\ref{sec:experiments:cate_coverage} shows overlap weights are
necessary but not sufficient under combined low-overlap +
contamination.

\textbf{Q14: When do I need modular Bayes?}
Compute the dispersion ratio $\rho = \mathrm{std}_K(\bar\beta^{(k)}) / \mathrm{mean}_K(\mathrm{CI~width}^{(k)})$
across $K = 5$ re-cross-fits (§\ref{sec:experiments:coverage}).
$\rho > 0.15$ indicates cross-fit instability; enable modular Bayes
with $M = 8$. Severity=severe consistently has $\rho \le 0.12$ and
generally does not need modular pooling at the cost of typical
intervals.

\subsection{Empirical scope}

\textbf{Q15: Why is IHDP not statistically significant?}
At 5 reps no method differs significantly from RX-Learner at
$\alpha = 0.05$; at 10 reps the rank ordering changes and the
abstract claim is "competitive, not dominant"
(§\ref{sec:experiments:ihdp}). IHDP is a clean-data benchmark where
robust methods are not expected to win.

\textbf{Q16: Have you tested non-whale contamination?}
Yes: Pareto ($\alpha = 1.5$), Student-$t_2$, Student-$t_3$,
Student-$t_5$, asymmetric (treated-only), bimodal (sign-symmetric),
and $\alpha$-stable (round-13). All handled by
\texttt{severity="severe"} with bias $\le 0.13$.

\textbf{Q17: Have you tested real heavy-tailed data?}
Yes: Hillstrom RCT ($N = 42{,}613$, Hill index $\hat\alpha \approx 2$
on positive spend, see Figure~\ref{fig:hillstrom_hill}) and Lalonde NSW
($N = 445$, dollar-scale earnings). Hillstrom passes placebo + symmetry
+ propensity-stratified placebo + holdout (§\ref{sec:experiments:hillstrom}).

\textbf{Q18: How does the method scale?}
Single fit: $5{-}20$\,s for RX-Learner Phase 3 NUTS at $N = 1000$, $p
\le 100$ (§\ref{sec:experiments:other_paradigms}). At $N = 10{,}000$,
$p = 100$: $4.2$\,s, $+53$\,MB peak RSS
(§\ref{sec:experiments:other_paradigms}). Modular-Bayes pooling
multiplies runtime by $M \in [5, 10]$.

\subsection{Comparison to alternatives}

\textbf{Q19: How does the Welsch posterior compare to RBCI
$\omega$-tuned squared-loss generalised Bayes?}
RBCI $\omega$-tuning of squared-loss (round-13,
§\ref{sec:experiments:coverage}) selects $\omega$ to match a
bootstrap variance target; it gives wider intervals than Welsch +
trace-$\eta$ at comparable coverage, with bias scaling like the
non-robust ATE under contamination. The Welsch redescender adds
\emph{bias control} on top of the temperature-only calibration that
$\omega$-tuning provides.

\textbf{Q20: What about Lasso-DR, Catoni-DR, sandwich-CI DR-learners?}
Round-13 (§\ref{sec:experiments:other_paradigms}) tests all three.
Lasso-DR (sparse linear regression on DR pseudo-outcomes) and
Catoni-DR (heavy-tail-robust mean estimator) are unbounded-influence
and fail just like Huber-DR under contamination. Tail-trimmed-IPW (round-7) catastrophically
fails because trimming changes the estimand. Bounded-influence at
the regression / likelihood layer is the operational requirement.

\textbf{Q21: Why not GRF (Generalised Random Forests)?}
GRF is a strong nonparametric CATE alternative for clean / mildly
contaminated data. We did not run it because the reference
implementation is in R and we kept the pipeline single-language
(see §\ref{sec:discussion:limitations}). A multi-language reproduction
is a natural extension.

\subsection{Software and reproducibility}

\textbf{Q22: Can the library detect and warn about edge cases?}
Yes. The current implementation warns when (i) min/max $\hat\pi$ is
outside $[0.05, 0.95]$ (small-eigenvalue risk) and (ii) auto-enables
\texttt{use\_overlap=True} when $\hat\pi$ falls outside
$[0.02, 0.98]$ (round-13 auto-fallback). Users can set
\texttt{use\_overlap} explicitly to silence the fallback.

\textbf{Q23: Are there regression tests?}
Yes — \texttt{tests/test\_default\_config\_regression.py} pins the
default configuration and the \texttt{contamination\_severity} API; 9 tests
pass on every commit. The library is configured for deterministic
runs given a \texttt{random\_state}.

\end{document}